\documentclass[11pt, headings = small]{scrartcl}
\usepackage{natbib}
\usepackage{epsfig}
\usepackage{graphicx}
\usepackage{color}
\usepackage{amsfonts}
\usepackage{setspace}
\usepackage{amsmath}
\usepackage{mathtools}
\usepackage{amssymb}
\usepackage{scrlayer-scrpage}
\usepackage{tabto}
\usepackage{url}
\usepackage{rotating}
\usepackage{booktabs} 
\usepackage{adjustbox} 
\usepackage{siunitx}
\usepackage{exsheets}
\usepackage{array}
\usepackage[utf8]{inputenc} 
\usepackage{pbox}
\usepackage{graphicx}
\usepackage{ifthen}
\usepackage{hhline}
\usepackage{lastpage}
\usepackage{wasysym}
\usepackage{subfloat}
\usepackage{makecell}
\usepackage{multirow}
\usepackage{pict2e,picture}
\usepackage{wrapfig, lipsum, booktabs}
\usepackage{ragged2e}

\usepackage{xcolor}
\newcolumntype{?}{!{\vrule width 1.5pt}}
\usepackage{tikz}
\usepackage{graphicx}
\usetikzlibrary{patterns}
\usetikzlibrary{shapes.geometric,arrows}
\RequirePackage{xcolor}
\definecolor{darkblue}{rgb}{0, 0, 0.5}
\usepackage[colorlinks=false, allcolors=darkblue]{hyperref} 

\usetikzlibrary{automata, positioning}
\usepackage{flowchart}
\usepackage{graphicx}
\usepackage{subcaption}
\usepackage{mwe}
\usetikzlibrary{backgrounds}
\usetikzlibrary{shapes,arrows}

\allowdisplaybreaks
\usepackage{paralist}
\usepackage{algorithm}
\usepackage[noend]{algpseudocode}
\usepackage{amsthm}
\makeatletter
\DeclareMathOperator*{\argmax}{arg\,max}

\def\BState{\State\hskip-\ALG@thistlm}

\makeatother
\usepackage[left=20mm,right=20mm,top=20mm,bottom=20mm]{geometry}

\usepackage{pict2e,picture}

\makeatletter
\DeclareRobustCommand{\Arrow}[1][]{%
	\check@mathfonts
	\if\relax\detokenize{#1}\relax
	\settowidth{\dimen@}{$\m@th\rightarrow$}%
	\else
	\setlength{\dimen@}{#1}%
	\fi
	\sbox\z@{\usefont{U}{lasy}{m}{n}\symbol{41}}%
	\begin{picture}(\dimen@,\ht\z@)
		\roundcap
		\put(\dimexpr\dimen@-.7\wd\z@,0){\usebox\z@}
		\put(0,\fontdimen22\textfont2){\line(1,0){\dimen@}}
	\end{picture}%
}
\makeatother
\usepackage[font=footnotesize]{caption}

\usepackage{natbib}
 \bibpunct[, ]{(}{)}{,}{a}{}{,}%
 \def\BIBand{and}%
\DeclareMathSizes{10.95}{9}{7}{5}
\title{\vspace{-2cm} From approximation error to optimality gap - Explaining the performance impact of opportunity cost approximation in integrated demand management and vehicle routing}
\author{
David Fleckenstein, Robert Klein, Vienna Klein \\ \small david.fleckenstein@wiwi.uni-augsburg.de, robert.klein@wiwi.uni-augsburg.de,\\ \small vienna.klein@uni-a.de (corresponding author)\\ \small Chair of Analytics \& Optimization, University of Augsburg\\
\\
Claudius Steinhardt \\ \small claudius.steinhardt@unibw.de\\ \small Chair of Business Analytics \& Management Science, University of the Bundeswehr Munich
}

\date{\today}
\begin{document}
	\maketitle  
	\begin{abstract}
        \noindent The widespread adoption of digital distribution channels both enables and forces more and more logistical service providers to manage booking processes actively to maintain competitiveness. As a result, their operational planning is no longer limited to solving vehicle routing problems. Instead, demand management decisions and vehicle routing decisions are optimized integratively with the aim of maximizing revenue and minimizing fulfillment cost. The resulting integrated demand management and vehicle routing problems (i-DMVRPs) can be formulated as Markov decision process models and, theoretically, can be solved via the well-known Bellman equation. Unfortunately, the Bellman equation is intractable for realistic-sized instances. Thus, in the literature, i-DMVRPs are often addressed via decomposition-based solution approaches involving an opportunity cost approximation as a key component. Despite its importance, to the best of our knowledge, there is neither a technique to systematically analyze how the accuracy of the opportunity cost approximation translates into overall solution quality nor are there general guidelines on when to apply which class of approximation approach. In this work, we address this research gap by proposing an explainability technique that quantifies and visualizes the magnitude of approximation errors, their immediate impact, and their relevance in specific regions of the state space. Exploiting reward decomposition, it further yields a characterization of different types of approximation errors. Applying the technique to a generic i-DMVRP in a full-factorial computational study and comparing the results with observations in existing literature, we show that the technique contributes to better explaining algorithmic performance and provides guidance for the algorithm selection and development process. \\[0.1cm]
	\end{abstract}
    
    \section{Introduction}\label{sec: introduction}

The proliferation of e-commerce and the progress of communication technology has led to the emergence and establishment of new business models that allow customers to book on-demand logistical services, mostly the delivery of goods (\cite{wassmuth2022demand}) or local transportation (\cite{vansteenwegen2022survey}). Prominent examples of these services are attended home delivery (AHD), same-day delivery (SDD), or mobility-on-demand (MOD). 
These business models have in common that customers expect a very high service level, e.g., in terms of the deviation from their desired service time (\cite{amorim2024customer}). Meeting these expectations makes demand consolidation challenging, which entails high fulfillment cost (\cite{ulmer2020dynamic}). To still operate profitably, operational planning for these business models has evolved: Instead of optimizing the associated vehicle routing alone, providers additionally apply demand management to achieve efficient fulfillment operations. \\[0.1cm]
The resulting integrated demand management and vehicle routing problems (i-DMVRPs) are
stochastic and dynamic with two types of integrated decisions: For each dynamically arriving customer request, the provider integratively makes a \textit{demand control decision} and a \textit{vehicle routing decision} with the overall objective of maximizing the expected profit, i.e., revenue net of operational fulfillment cost. Such an i-DMVRP can be modeled as a Markov decision process (MDP) and, theoretically, be solved by evaluating the well-known Bellman equation (\cite{puterman2014markov}). Practically, however, i-DMVRPs suffer from the curses of dimensionality ((\cite{powell2007approximate}) such that this is not tractable for realistic-sized instances. Consequently, in literature, demand control decisions for i-DMVRPs are often optimized with a decomposition-based solution approach. More precisely, two subproblems are solved sequentially for every incoming customer request (\cite{fleckenstein2021recent}, \cite{ulmer2020dynamic}, \cite{gallego2019revenue}, p. 25, \cite{klein2018model}): 
\begin{itemize}
\item[1.)] Approximating opportunity cost (OC) for each potential fulfillment option (e.g., different time windows) to measure the expected profit impact assuming the current customer chooses the respective option, given the state of the system.
\item[2.)] Solving the actual demand control problem based on the approximated OC, i.e., deciding on the pricing or availability of fulfillment options, or the acceptance/rejection of the request. 
\end{itemize}
This leads to the following conclusion: One of the main avenues for accelerating the development of practical solution approaches for i-DMVRPs is understanding and explaining the relation between the accuracy reached in (1), i.e., the \textit{accuracy of the OC approximation}, and the quality of (2), i.e., the \textit{quality of the resulting demand control decision}. However, despite the maturity of integrated demand management and vehicle routing as a research area, this relation has not been systematically explored, and therefore, largely remains a black box so far. Hence, we now close this research gap by comprehensively analyzing this relationship. To do so, we first introduce a novel explainability technique for i-DMVRPs that combines two building blocks:
\begin{quotation}
\textit{B1: Chain of influencing factors --} The first building block resembles a typical post-hoc explainability technique (\cite{arrieta2020explainable}), i.e., we define metrics to evaluate the behavior of a given policy in certain states and the respective impact on solution quality. Thereby, we aim at answering the central questions that arise along the chain of influencing factors from OC approximation error to objective value loss as depicted in Figure \ref{fig: causal_chain}. This allows deriving insights on when (in which states) and why a certain policy performs especially good or bad.\\[0.1cm]
\textit{B2: Reward decomposition --} The second building block incorporates the idea of reward decomposition (\cite{juozapaitis2019explainable}). For that, we exploit the finding that opportunity cost can be decomposed into displacement cost (DPC) and marginal cost-to-serve (MCTS) (\cite{fleckenstein2023concept}). More precisely, we propose to apply OC approximations capturing only one of the two components, with the aim of assessing the importance of the respective component for approximation accuracy. 
\end{quotation}

\noindent Afterward, in an extensive computational study, we apply the explainability technique to the generic i-DMVRP and the stylized parameter settings from \cite{fleckenstein2023concept}. Then, we complement the numerical results by an analysis of the existing literature on i-DMVRPs. Therewith, we confirm the validity of our findings and show that our identified OC approximation error types can indeed explain the observed performance of state-of-the-art solution approaches. In summary, our work has four contributions:
\begin{itemize}
\itemsep0em
    \item[1.)] To the best of our knowledge, we introduce the first explainability technique for the widely established decomposition-based solution approaches for i-DMVRPs.  
    \item[2.)] We apply our explainability technique within a comprehensive computational study and identify fundamental OC approximation error types, i.e., OC approximation errors that can occur in a broad variety of real-world i-DMVRPs. Therewith, we are the first to systematically analyze the relation between the accuracy of OC approximation and the objective value. 
    \item[3.)] We classify patterns in the occurrence of the fundamental approximation error types we identify, characterize which problem settings are prone to which error type, and propose algorithmic elements to successfully mitigate them. This yields insights that guide the selection and the design of OC approximation algorithms. 
    \item[4.)] We compile indications for the occurrence of the identified error types from existing literature and show that our findings improve explainability of the reported results. Thereby, we transform the existing implicit knowledge about specific i-DMVRPs to explicit, high-level knowledge.
\end{itemize} The remainder of this paper is structured as follows: In Section \ref{sec: literatureReview}, we review the related literature both on algorithmic explainability and i-DMVRPs. In Section \ref{sec: OCforiDMVRP}, we introduce and model the generic i-DMVRP under consideration. Then, in Section \ref{sec: methodology}, we present our novel explainability technique for i-DMVRPs in detail, and we present our computational study in Section \ref{sec: computationalStudy}. In Section \ref{sec: managerialInsights}, we derive general insights for algorithm design and summarize our work in Section \ref{sec: conclusion}.
\begin{figure}[ht]
                \centering
                \caption[Chain of influencing factors from OC approximation error to objective value loss]
                                {Chain of influencing factors from OC approximation error to objective value loss}
                                \label{fig: causal_chain}
                                \centering
                                \vspace{0.2cm}
                                \resizebox{0.8\textwidth}{!}{
                                \begin{tikzpicture}
                                \small

                                \draw[black, ultra thick, -latex] (3,1.8)--(4.5,1.8);
                                \draw[black, ultra thick, -latex] (7.5,1.8)--(9,1.8);
                                \draw[black, ultra thick, -latex] (12,1.8)--(13.5,1.8);


                                \draw[draw=black, thin, fill=white, ] (0,1) rectangle ++(3,1.5) node[align=center, pos=.5, text width=2.5cm] {\baselineskip=12pt \textbf{How bad is \\ the approxim- \\ ation error?}\par};
                                
                                \draw[draw=black, thin, fill=white, ] (4.5,1) rectangle ++(3,1.5) node[align=center, pos=.5, text width=2.5cm] {\baselineskip=12pt \textbf{How wrong is \\ the resulting \\ decision?}\par};
                                
                                \draw[draw=black, thin, fill=white] (9,1) rectangle ++(3,1.5) node[align=center, pos=.5, text width=2.5cm] {\baselineskip=12pt \textbf{How likely is \\ this decision?}\par};
                                
                                \draw[draw=black, thin, fill=white] (13.5,1) rectangle ++(3,1.5) node[align=center, pos=.5, text width=2.5cm] {\baselineskip=12pt \textbf{What is the \\ performance \\ impact?}\par};

                                \draw[black] (1.5,1)--(1.5,0.5);
                                \draw[black] (6,1)--(6,0.5);
                                \draw[black] (10.5,1)--(10.5,0.5);
                                \draw[black] (15,1)--(15,0.5);

                                \draw[draw=black, thin, fill=white] (0,-4.5) rectangle ++(3,5) node[align=center, pos=.5, text width=2.5cm] {OC approximations can underestimate or overestimate the true OC. Both can but do not necessarily cause suboptimal decisions.};
                                
                                \draw[draw=black,thin, fill=white, ] (4.5,-4.5) rectangle ++(3,5) node[align=center, pos=.5, text width=2.5cm] {Suboptimal decisions can lead to less reward than the optimal decision, transition the system to a lower-valued state, or both.};
                                
                                \draw[draw=black, thin, fill=white] (9,-4.5) rectangle ++(3,5) node[align=center, pos=.5, text width=2.5cm] {The relevance of a wrong decision depends on how likely it is that the state in which this decision is made will be encountered when following a certain policy.};
                                
                                \draw[draw=black, thin, fill=white] (13.5,-4.5) rectangle ++(3,5) node[align=center, pos=.5, text width=2.5cm] {Only those approximation errors that cause notably suboptimal decisions and that are made often result in a considerable objective value loss.};
                                
                               \end{tikzpicture}
}
\end{figure}%
    
    \section{Literature review}\label{sec: literatureReview}
Due to the cross-cutting nature of our study, the related literature spans across multiple distinct research areas. In Section \ref{sec: literature_i-DMVRP}, we review the literature on i-DMVRPs with a special focus on its origins in revenue management, dynamic pricing, and dynamic vehicle routing. In Section \ref{sec: explainability}, we then discuss algorithmic explainability techniques, particularly from explainable reinforcement learning (RL), highlighting the techniques that we adapt and apply in the work at hand. Finally, in Section \ref{sec: control behavior}, we review the descriptive analytics that authors use to explain the observed performance of their i-DMVRP solution approaches.  
\subsection{Modeling and solving i-DMVRPs}\label{sec: literature_i-DMVRP} 
In logistics, many companies dynamically collect orders for a transportation service that is fulfilled by a given fleet of vehicles. These companies face an i-DMVRP if they can both plan individual offers made in response to customer requests and plan the vehicle routes to feasibly fulfill the resulting orders. Hence, i-DMVRP research synthesizes two originally distinct research areas:
\begin{itemize}
 \item[1.)] Research in \textit{revenue management and dynamic pricing} addresses the dynamic optimization of offering decisions under the assumption that fulfillment is already pre-planned. For an extensive overview of this field, we refer the reader to the textbooks by \cite{gallego2019revenue} and \cite{talluri2004theory}  as well as the reviews by \cite{klein2020review} and \cite{strauss2018review}.
 \item[2.)]\textit{Dynamic vehicle routing} investigates the optimization of fulfillment assuming given orders that arrive dynamically. For a deeper discussion of this research area, we refer the reader to the textbook by \cite{toth2014vehicle} as well as the reviews by \cite{hildebrandt2023opportunities}, \cite{soeffker2022stochastic}, and \cite{psaraftis2016dynamic}. 
\end{itemize}
Starting with the seminal work of \cite{campbell2005decision} on an AHD system, i-DMVRPs are considered in a variety of applications such as SDD (\cite{azi2012dynamic}), MOD (\cite{atasoy2015concept}), or mobile personnel booking (\cite{avraham2021steady}). Indicative of the growing importance of this research area, there are several reviews that are either application-specific (\cite{liemerging}, \cite{wassmuth2022demand}, and \cite{snoeck2020revenue}), or aim at i-DMVRP literature in general (\cite{fleckenstein2021recent}). \\[0.1cm]
Since i-DMVRPs are dynamic and stochastic, the natural modeling approach is to formulate a Markov decision process (MDP) model (\cite{puterman2014markov}). It is important to note that MDP models not only serve as a formal problem definition. On top of that, model analysis, which can be done analytically or numerically (\cite{bravo2020mining}), yields domain knowledge that can be exploited by solution approaches. For i-DMVRPs, such model analyses can be found in \cite{fleckenstein2023concept}, \cite{lebedev2021dynamic}, and \cite{asdemir2009dynamic}. In particular, we draw on the property that OC can be decomposed into MCTS and DPC (\cite{fleckenstein2023concept}). The three works have in common that they mainly analyze models analytically. In contrast, our technique focuses on the numerical analysis of solution approaches. It also yields domain knowledge regarding a combination of a solution approach and a model of a specific i-DMVRP. \\[0.1cm]
Although i-DMVRPs can be solved to optimality by exact dynamic programming algorithms, this is impractical for realistic-sized instances.
Hence, there exists a wide variety of heuristic solution approaches for specific i-DMVRPs. According to \cite{fleckenstein2021recent}, they can be classified into two broad solution concepts: First, there are static deterministic approaches that solve auxiliary models with a rolling-horizon to iteratively derive demand management and vehicle routing decisions (e.g. \cite{klapp2020request}). Second, there are decomposition-based solution approaches that subdivide the demand management task into two sub-problems as already mentioned: OC approximation and demand management decision-making.\\[0.1cm] 
In this work, we only consider the decomposition-based solution approaches, which are adopted by the majority of authors (\cite{fleckenstein2021recent}). The applied OC approximation approaches can be sub-divided further into sampling-based (e.g. \cite{klein2021dynamic}) and learning-based (e.g. \cite{ulmer2020dynamic}) and are either targeted at approximating MCTS, DPC, or both (\cite{fleckenstein2023concept}). To evaluate the performance of these approximation approaches, authors of existing works resort to descriptive analyses. With our work, we aim at explaining the reasons for the observed results by investigating the impact of OC approximation errors.
\subsection{Explainability of algorithmic performance and behavior}\label{sec: explainability}
Explainability becomes increasingly relevant in analytics and optimization in general (see, e.g., the recent reviews by \cite{debock2023explainable} and  \cite{goerigk2023framework}). In particular, the field of explainable RL (XRL) has recently gained more attention (\cite{milani2023explainable}). Due to the close relation between RL and the OC approximation approaches observed in i-DMVPR literature, the techniques developed for XRL are also applicable to most approaches tackling i-DMVRPs.\\[0.1cm]
\cite{milani2023explainable} introduce a two-dimensional taxonomy of explainability techniques tailored to XRL. The first classification dimension proposed by the authors is borrowed from general explainable artificial intelligence: 
\begin{itemize}
\item[1.)] Explainability can be \textit{inherent} to a policy or restored \textit{post-hoc}. 
\item[2.)] We can further distinguish \textit{local} explanations that refer to individual states, and \textit{global} explanations that holistically view the behavior of the policy. 
\item[3.)] Among the post-hoc techniques, a distinction can be made regarding the \textit{degree of portability}, i.e., the range of solution approaches the technique can be readily applied to.
\end{itemize}
Regarding this classification dimension, our explainability technique is a \textit{post-hoc} explanation since it is applied to a given (decomposition-based) policy. As discussed in Section \ref{sec: literature_i-DMVRP}, this type of policy is quite common, which makes our technique \textit{portable}. Further, it features \textit{local} (state-level) metrics but also involves \textit{global} considerations since these local metrics are aggregated to explain the global behavior of the policy.\\[0.1cm] 
The second classification dimension specifically addresses XRL approaches and distinguishes explainability techniques based on the type of explanations they incorporate as follows: 
\begin{itemize}
    \item[1.)] \textit{Feature importance explanations}: explaining individual actions by providing their context, e.g., state features. Typical approaches are, e.g., surrogate policies encoded as decision trees or saliency map explanations.
    \item[2.)] \textit{Learning process and MDP explanations}: exploiting the definition of MDP model elements or training process steps to generate explanations. The aim is to identify critical drivers of the policy's individual decisions.
    \item[3.)] \textit{Policy-level explanations}: identifying recurring sequences of decisions (e.g. by clustering states) to extract patterns of the policy's overall control behavior.
\end{itemize} 
Regarding this second classification dimension, our technique can be viewed as a combination of \textit{policy level explanations} (in B1) and \textit{learning process and MDP explanations} (in B2). In the following, we briefly review the closest related literature for B1 and B2 separately.\\[0.1cm]
B1 of our explainability technique is closely related to a technique called \textit{strategy summarization} by \cite{amir2019summarizing}. They suggest identifying states of interest on the basis of importance, coverage, likelihood of encountering, and policy disagreement with the aim of aggregating these states to summarize the behavior of the policy. Applied to i-DMVRPs, measuring the OC approximation error itself can be considered equal to measuring policy disagreement with the optimal policy. To quantify state importance, we measure the impact of an approximation error in a certain state on the quality of the resulting decision. This can also be viewed as a special case of the state importance metric used by \cite{torrey2013teaching}. Further, like \cite{amir2019summarizing}, we consider the likelihood of encountering a state.\\[0.1cm] 
B2 of our explainability technique is a reward decomposition technique. It is first proposed by \cite{russell2003q} with the aim of facilitating the learning process. With the same goal, it is also applied by \cite{van2017hybrid} in the form of a hybrid reward architecture. 
However, as shown by \cite{juozapaitis2019explainable}, reward decomposition can not only be applied for designing hybrid reward architectures but also as an explainability technique. Therefore, they analyze the influence of the different reward components for explaining the behavior of a given policy. In contrast, we analyze approximation errors that result from considering only one reward component for explaining the behavior of a given policy. This idea of analyzing approximation errors in RL is first presented by \cite{mannor2007bias} with the aim of computing confidence intervals. \\[0.1cm]
Regarding the application of our explainability technique, a distinguishing feature compared to most existing works in XRL is that we consider a large number of small problem instances and solve them to optimality. Thereby we derive generic domain knowledge, in the form of fundamental OC approximation error types, rather than analyzing heuristic policies for large instances. In this regard, we only found one similar approach by \cite{bravo2020mining}. They use machine learning to analyze optimal policies for small problem instances of, amongst others, traditional revenue management problems.\\[0.1cm]
In summary, our methodology combines a variety of existing RL explainability techniques in a novel way: Besides adapting them to the problem structure of i-DMVRPs, we introduce the new idea of combining strategy summarization and reward decomposition and applying both to derive characterizations of fundamental OC approximation errors.
\subsection{Performance metrics in i-DMVRP literature}
\label{sec: control behavior}
In contrast to ``pure" revenue management and dynamic pricing, where explainability has already received some attention (e.g., \cite{biggs2021model}, \cite{bravo2020mining}), we find no systematic application of techniques from XRL in the literature on i-DMVRPs. Instead, most authors evaluate the performance of their solution approaches by incorporating descriptive analytics, as we summarize in the following.\\[0.1cm]
\textit{Aggregate metrics --} Apart from the arithmetic mean of profit, which is the objective in most of the considered i-DMVRPs, many authors additionally report the following aggregate metrics describing the performance of policies: Among the most widely reported metrics are average or overall revenue, cost, and number of orders (\cite{campbell2005decision}). Further, some authors also report revenue per order (\cite{klein2018model}), cost per order (\cite{yang2016choice}), average number of fulfillment options offered to each customer (\cite{mackert2019choice}), pooling rate (\cite{anzenhofer2023analyzing}), or fleet utilization (\cite{klein2021dynamic}). In addition to the arithmetic mean, the standard deviation (\cite{yang2016choice}) or the coefficient of variation (\cite{anzenhofer2023analyzing}) are reported in a few studies.\\[0.1cm]
\textit{Decision-making --} For a more detailed analysis of a policies' performance, authors analyze how the resulting decision-making differs over time, i.e., over the course of the booking horizon, or for different types of requests: For any i-DMVRP, the acceptance rate or conversion rate (\cite{mackert2019choice}) or the cumulative revenue over time (\cite{lang2021anticipative}) can be reported. If customers can choose from a set of fulfillment options, the number (\cite{abdollahi2023demand}) or composition of offered fulfillment options (\cite{klein2021dynamic}), or the chosen fulfillment options (\cite{anzenhofer2023analyzing}) can be analyzed. If dynamic pricing is applied, average prices of offered (\cite{klein2018model}) or chosen fulfillment options (\cite{yang2016choice}) are reported.\\[0.1cm]
\textit{Opportunity cost --} If a parametric OC approximation is used, its parameter values (\cite{lang2021anticipative}) or the function values for certain parameter values (\cite{avraham2021steady}) can be investigated. Only very rarely, authors directly consider approximated OC values for different groups of similar requests (\cite{yang2017approximate}) or over time (\cite{koch2020route}).\\[0.1cm]
In general, we identify three central problems that limit the explanatory power of the existing descriptive analyses: First, observations of the performance and the behavior of a policy do not provide direct evidence of whether or how exactly an OC approximation error influences the observed performance. Due to a lack of conclusive explanations, the reasoning is often limited to formulating hypotheses. Second, the metrics are only analyzed in an aggregate form, which does not allow distinguishing different types of errors that originate in certain regions of the state space. Third, since typically, a specific solution approach for a specific i-DMVRP is considered, the results are hardly attributable to certain characteristics of the problem structure, the instance structure, or the solution approach. This again limits conclusiveness and transferability. \\[0.1cm]
Overall, there is a clear research gap regarding the development of explainability techniques for i-DMVRPs and the formulation of generalizable explanations for policy performance. 
    \section{Problem definition and modeling}\label{sec: OCforiDMVRP}
In this section, we formally characterize i-DMVRPs with a particular focus on the generic MDP model for i-DMVRPs by \cite{fleckenstein2023concept}.\\[0.1cm]
Typically, an i-DMVRP is structured as follows: During a \textit{booking horizon} customers log-in to the business platform and place a service request by entering service parameters like pick-up/drop-off locations, desired fulfillment times, or vehicle types. In response, the provider either presents a set of suitable \textit{fulfillment options} with different prices to choose from, or accepts/rejects the request. Then, a successfully placed customer request turns into a confirmed \textit{customer order}. All customer orders are eventually served by the provider within the \textit{service horizon}, which can either be \textit{disjoint} or \textit{overlapping} with the booking horizon. The former is typical for AHD, where customer and provider agree on a delivery time window for a certain day in advance. The latter is typical for SDD or MOD, where the customer expects to receive a service on short notice.\\[0.1cm] 
In the following, we consider the generic i-DMVRP model as in \cite{fleckenstein2023concept} but adapt it specifically for the case of disjoint booking horizons and service horizons. Further, the underlying demand control subproblem features an accept/reject demand control. However, the generalization to multi-option demand control is straightforward (see \cite{fleckenstein2023concept}).
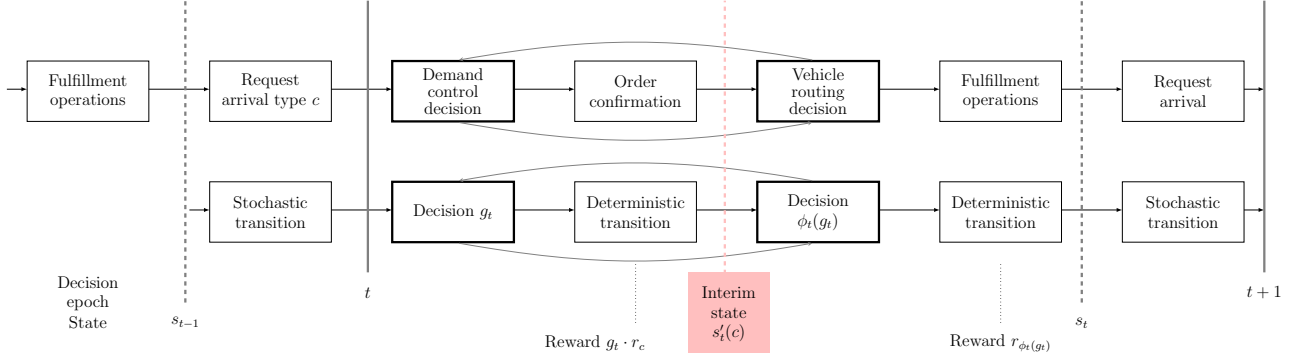
\begin{figure}[ht]
                \centering
                \caption[Overview of the MDP model of the i-DMVRP booking and fulfillment process including the interim state, copied from \cite{fleckenstein2023concept}]
                                {Overview of the MDP model of the i-DMVRP booking and fulfillment process including the interim state (\cite{fleckenstein2023concept})}
                                \label{fig: overview_iDMVRPs_MDP_interim}
                                \centering
                                \vspace{0.2cm}
                                \large
                                \resizebox{\textwidth}{!}{
                                \begin{tikzpicture}
                                
                                \draw[draw=black, thin, fill=white] (-4.5, 3) rectangle ++(3,1.5) node[align=center, pos=.5] { Fulfillment\\ operations};
                                
                                \draw[draw=black, thin, fill=white] (0,3) rectangle ++(3,1.5) node[align=center, pos=.5] { Request\\ arrival type $c$};
                                
                                \draw[draw=black, ultra thick, fill=white, ] (4.5,3) rectangle ++(3,1.5) node[align=center, pos=.5, text width=2.3cm] {\baselineskip=12pt Demand \\ control \\ decision\par};
                                
                                \draw[draw=black, thin, fill=white] (9,3) rectangle ++(3,1.5) node[align=center, pos=.5] { Order\\ confirmation};
                                
                                \draw[draw=black, ultra thick, fill=white] (13.5,3) rectangle ++(3,1.5) node[align=center, pos=.5, text width=2.3cm] { \baselineskip=12pt Vehicle\\ routing \\ decision\par};
                                
                                \draw[draw=black, thin, fill=white] (18, 3) rectangle ++(3,1.5) node[align=center, pos=.5] { Fulfillment\\ operations};
                                
                                \draw[draw=black, thin, fill=white] (22.5,3) rectangle ++(3,1.5) node[align=center, pos=.5] { Request\\ arrival};

        \draw[black, -latex] (-5,3.8)--(-4.5,3.8);
                                \draw[black, -latex] (-1.5,3.8)--(0,3.8);
                                \draw[black, -latex] (3,3.8)--(4.5,3.8);
                                \draw[black, -latex] (7.5,3.8)--(9,3.8);
                                \draw[black, -latex] (12,3.8)--(13.5,3.8);
                                \draw[black, -latex] (16.5,3.8)--(18,3.8);
                                \draw[black, -latex] (21,3.8)--(22.5,3.8);
                                \draw[black, -latex] (25.5,3.8)--(26,3.8);
                                

                                \draw[draw=black, thin, fill=white] (0,0) rectangle ++(3,1.5) node[align=center, pos=.5] {Stochastic \\ transition};
                                
                                \draw[draw=black, ultra thick, fill=white, ] (4.5,0) rectangle ++(3,1.5) node[align=center, pos=.5, text width=2.3cm] {Decision $g_t$ };
                                
                                \draw[draw=black, thin, fill=white] (9,0) rectangle ++(3,1.5) node[align=center, pos=.5] { Deterministic\\ transition};
                                
                                \draw[draw=black, ultra thick, fill=white] (13.5,0) rectangle ++(3,1.5) node[align=center, pos=.5, text width=2.3cm] { Decision $\phi_{t}(g_t)$};
                                
                                \draw[draw=black, thin, fill=white] (18, 0) rectangle ++(3,1.5) node[align=center, pos=.5] { Deterministic\\ transition};
                                
                                \draw[draw=black, thin, fill=white] (22.5, 0) rectangle ++(3,1.5) node[align=center, pos=.5] { Stochastic\\ transition};
                                
                                \draw[black, -latex] (-0.5,0.8)--(0,0.8);
                                \draw[black, -latex] (3,0.8)--(4.5,0.8);
                                \draw[black, -latex] (7.5,0.8)--(9,0.8);
                                \draw[black, -latex] (12,0.8)--(13.5,0.8);
                                \draw[black, -latex] (16.5,0.8)--(18,0.8);
                                \draw[black, -latex] (21,0.8)--(22.5,0.8);
                                \draw[black, -latex] (25.5,0.8)--(26,0.8);
                               
                                \node at (-3, -1) {\textcolor{black}{Decision}};
                                \node at (-3, -1.5) {\textcolor{black}{epoch}};
                                
                                \node at (-3, -2) {\textcolor{black}{State}};
                                
                                \node at (9.5, -2.5) {\textcolor{black}{Reward $g_{t} \cdot r_{c}$}};
                                \node at (19.5, -2.5) {\textcolor{black}{Reward $r_{\phi_{t}(g_t)}$}};
                                
                                \draw[black, dotted] (10.5, -2) to (10.5, -0.5);
                                \draw[black, dotted] (19.5, -2) to (19.5, -0.5);
                                
                                \draw[gray, ultra thick] (3.9, -0.75) to (3.9, 6);
                                \draw[gray, ultra thick] (26, -0.75) to (26,6);
                               
                                \node at (3.9, -1.25) {\textcolor{black}{$t$}};
                               \node at (26, -1.25) {\textcolor{black}{$t+1$}};
                               
                                \draw[gray, dashed, ultra thick] (-0.6, -1.5) to (-0.6, 6);
                                \draw[gray, dashed, ultra thick] (21.5, -1.5) to (21.5,6);
                               
                                \node at (-0.6, -2) {\textcolor{black}{$s_{t-1}$}};
                               \node at (21.5, -2) {\textcolor{black}{$s_{t}$}};

                                \node (1) at (6,4.5) {};
                                \node (2) at (15,4.5) {};
                                \node (3) at (6,3) {};
                                \node (4) at (15,3) {};
        
        \path[draw=gray, -stealth]    
        (2) to [bend right=10] (1);
        
        \path[draw=gray, stealth-]  
        (4) to [bend left=10] (3);

        \node (5) at (6,1.5) {};
                                \node (6) at (15,1.5) {};
                                \node (7) at (6,0) {};
                                \node (8) at (15,0) {};
        
        \path[draw=gray, -stealth]    
        (6) to [bend right=10] (5);
        
        \path[draw=gray, stealth-]  
        (8) to [bend left=10] (7);

        \fill [pink] (11.8, -2.75) rectangle ++(2,2) 
        ;
        
        \node at (12.8, -1.25) {\textcolor{black}{Interim}};
                                \node at (12.8, -1.75) {\textcolor{black}{state}};
                                \node at (12.8, -2.25) {\textcolor{black}{$s'_t(c)$}};
                                
                                \draw[pink, ultra thick, dashed] (12.7,-1)--(12.7,6);              
                                
                               \end{tikzpicture}
}
\end{figure}\\[0.1cm]
\normalsize
\noindent\textit{Decision epoch --} A decision epoch marks the start of the MDP model's stages. In the considered problem, such stages correspond to (constant) time steps $t=1,...,T$. A \textit{customer request of type} $c \in C$ can arrive in stage $t$ with a certain \textit{arrival rate} $\lambda_c^t$. With each customer request of type $c$, the provider also receives data on the associated \textit{location(s)} $l_c$ and \textit{revenue} $r_c$. Individual customer requests are then uniquely identified by combining this information with their \textit{request time} $\tau$. Arrival rates are assumed to be small enough that at most one customer request arrives per stage. \\[0.1cm]
\textit{State --}  The system state $s_t = (\mathcal{C}_t,\phi_t)$ comprises two sets. The first set $\mathcal{C}_t$ consists of tuples $(c, \tau, o)$, which store customer orders for which fulfillment has not yet started. The second set $\phi_t$ stores the tour plan. Since we assume disjoint booking and service horizons, in our case, $\phi_t$ is either preliminary or empty for all $t<T$. Please note that $s_t$ defines a post-decision state. The state space of a decision epoch $t$ is denoted as $\mathcal{S}_t$ and comprises all potential realizations of customer orders $\mathcal{C}_t$ and tour plans $\phi_t$. Thus, $\forall t \in {1,..,T}: s_t \in \mathcal{S}_t $.\\[0.1cm]
\textit{Action --} An action in response to an arriving customer request of type $c$ integrates an accept/reject decision for demand control $g_{t}\in \mathcal{G}(s_{t-1}, c) \subseteq \{0,1\}$, and a tour planning decision $\phi_{t}(g_t)\in\Phi(s_{t-1}, c, g_t)$. Again, $\phi_t(g_t)$ is either preliminary or empty for all $t<T$ due to the disjoint horizons. The action space for the tour planning, denoted as $\Phi(s_{t-1}, c, g_t)$, is defined by the routing constraints of the problem and depends on the preceding state $s_{t-1}$, the type $c$ of the arriving request, and the demand control decision $g_t$. The action space for the demand control, denoted as $\mathcal{G}(s_{t-1}, c)$, in turn, depends on $\Phi(s_{t-1}, c, g_t)$ since $g_t=1$ is only feasible if $\Phi(s_{t-1}, c, g_t) \neq \emptyset$. Thus, $\mathcal{A}_t(s_{t-1},c) = \{(g_{t},\phi_t(g_{t})): g_{t} \in \mathcal{G}(s_{t-1}, c), \phi_t(g_t) \in \Phi(s_{t-1}, c, g_t)\}$.\\[0.1cm] 
\textit{Rewards --} As a consequence of an acceptance decision $g_t=1$, a revenue $r_{c}$ is received. A rejection yields no reward. A routing decision $\phi_{t}(g_t)$ entails a reward $r_{\phi_{t}(g_t)}$. It equals the newly arising fulfillment cost, which, given the triangle inequality holds, is non-positive. Again, since we assume disjoint booking and service horizons, $\forall t< T: r_{\phi_{t}(g_t)} = 0$. \\[0.1cm]
\textit{Transition --} When transitioning to state $s_{t}$, $\phi_{t}$ is set to $\phi_{t}(g_{t})$. The first state component $\mathcal{C}_{t-1}$, also changes. More precisely, if the newly arriving request of type $c_t$ is accepted, the resulting customer order is added. \\[0.1cm]
\noindent\textit{Objective --} The provider aims at maximizing profit after fulfillment. Therefore, it is required to determine a policy $\pi$ that returns the optimal decision for each state that can potentially be reached. These decisions of a policy $\pi$ can be denoted as $a_t^\pi(s_{t-1},c_{t})=(g_{t}^\pi(s_{t-1},c_{t}),\phi_{t}^\pi(g_{t}^\pi(s_{t-1},c_{t})))$ at decision epoch $t$. Then, the objective function is: 
\begin{equation}\label{eq: objectiveOPs}
     \smash{\displaystyle\max_{\pi}} ~ \mathbb{E} \biggl( \sum_{t=1}^T (r_{c_t} \cdot g_{t}^\pi(s_{t-1},c_{t}) + r_{\phi_{t}^\pi(g_{t}^\pi(s_{t-1},c_{t}))}) \mid s_{0} \biggr).
\end{equation}

\noindent\textit{Bellman equation --}
The objective function (\ref{eq: objectiveOPs}) can be expressed in the form of a Bellman equation, which defines a value $V_{t}(s_{t})$ for each state $s_{t}$. Solving this equation yields the optimal policy $\pi^{*}$. 
\begin{equation}\label{eq: Bellman_OPs_2max}
\begin{split}
	V_{t-1}(s_{t-1}) = & \sum_{c \in C} \lambda_{c}^{t} \cdot \smash{\displaystyle\max_{g_{t} \in \mathcal{G}(s_{t-1}, c)}} \biggl( g_{t} \cdot r_{c} + \smash{\displaystyle\max_{\phi_{t}(g_{t}) \in \Phi(s_{t-1}, c, g_{t})}} \bigl(r_{\phi_{t}(g_{t})} + V_{t}(s_{t} \mid s_{t-1},\phi_{t}(g_{t})) \bigr) \biggr) \\
	&+ \bigl(1-\sum_{c \in C} \lambda_{c}^{t}\bigr) \cdot \smash{\displaystyle\max_{\phi_{t}(0) \in \Phi(s_{t-1},0,0)}} \biggl(r_{\phi_{t}(0)} + V_{t}(s_{t} \mid s_{t-1},\phi_{t}(0))\biggr),
\end{split}
\end{equation}
with boundary condition:
\vspace{-0.7cm}
\begin{equation}\label{eq: boundaryCondition}
   V_T(s_{T}) = 0.
\end{equation}
In Equation (\ref{eq: Bellman_OPs_2max}), both types of decisions are represented in an integrated form. Thus, an interim state $s'_t \mid s_{t-1}, c, g_t$ can be defined to isolate the impact of the demand control decision from the impact of the vehicle routing decision as also depicted in Figure \ref{fig: overview_iDMVRPs_MDP_interim}. Further, substituting the OC of accepting a request of type $c$, i.e., $\Delta V_{t}(s_{t-1},c)$, we obtain the following reformulation. Note that we denote interim states $s'_t \mid s_{t-1}, c, 1$ by $s'_t(c)$ and interim states $s'_t \mid s_{t-1}, c, 0$, or $s'_t \mid s_{t-1}, 0, 0$, by $s'_t(0)$.
\begin{alignat}{5}\label{eq: Bellman_iDMVRPs_reform_OC}
\begin{split}
	V_{t-1}(s_{t-1}) = & \sum_{c \in C} \lambda_{c}^{t} \cdot \vphantom{\biggl(} \smash{\displaystyle \max_{g_{t} \in \mathcal{G}(s_{t-1}, c)}} \biggl(g_{t} \cdot \bigl(r_{c} - \Delta V_{t}(s_{t-1}, c)\bigr)\biggr) + V'_{t}(s'_{t}(0)),
\end{split}
\end{alignat}
with 
\begin{alignat}{5}\label{eq: interim_state_definition}
\begin{split}
    V'_t(s'_t \mid s_{t-1}, c, g_t) = \smash{\displaystyle\max_{\phi_{t}(g_{t}) \in \Phi(s_{t-1}, c, g_{t})}} \bigl(r_{\phi_{t}(g_{t})} + V_{t}(s_{t} \mid s_{t-1},\phi_{t}(g_{t})) \bigr) \\ 
    = r_{\phi^{*}_{t}(g_t)} + V_t(s_{t}\mid s_{t-1}, \phi^{*}_{t}(g_t)),
    \end{split}
\end{alignat}
and
\begin{alignat}{5}\label{eq: generalOC_DJPs}
        \Delta V_t(s_{t-1}, c) &= V'_t(s'_t(0)) - V'_t(s'_t(c)) \geq 0.
\end{alignat}
Further, $\phi^*_t(g_t)$ denotes the optimal routing decision for a given demand management decision $g_t$.

    \section{Explainability technique}\label{sec: methodology}
In this section, we present our novel explainability technique for i-DMVRPs, which comprises two separate building blocks. Both are later applied for the comprehensive analysis of the relation between OC approximation error and the quality of the resulting demand management decisions. 
\subsection{Building block 1}\label{sec: B1}
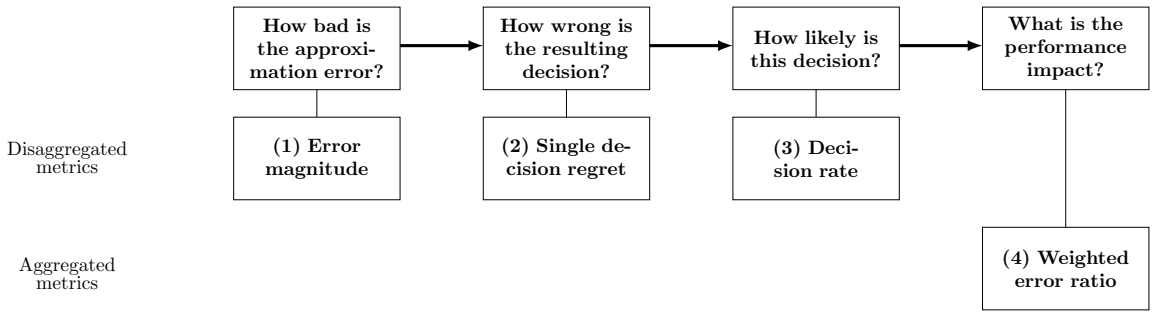
\begin{figure}[H]
                \centering
                \caption[Metrics among the chain of influencing factors from OC approximation error to objective value loss]
                                {Metrics among the chain of influencing factors from OC approximation error to objective value loss}
                                \label{fig: metrics_causal_chain}
                                \centering
                                \vspace{0.2cm}
                                \resizebox{0.9\textwidth}{!}{
                                \begin{tikzpicture}
                                \small
                                \draw[draw=black, thin, fill=white, ] (0,3) rectangle ++(3,1.5) node[align=center, pos=.5, text width=2.5cm] {\baselineskip=12pt \textbf{How bad is the approximation error?}\par};
                                
                                \draw[draw=black, thin, fill=white, ] (4.5,3) rectangle ++(3,1.5) node[align=center, pos=.5, text width=2.5cm] {\baselineskip=12pt \textbf{How wrong is the resulting decision?}\par};
                                
                                \draw[draw=black, thin, fill=white] (9,3) rectangle ++(3,1.5) node[align=center, pos=.5, text width=2.5cm] {\baselineskip=12pt \textbf{How likely is this decision?}\par};
                                
                                \draw[draw=black, thin, fill=white] (13.5,3) rectangle ++(3,1.5) node[align=center, pos=.5, text width=2.5cm] {\baselineskip=12pt \textbf{What is the performance impact?}\par};

                                \draw[black, ultra thick, -latex] (3,3.8)--(4.5,3.8);
                                \draw[black, ultra thick, -latex] (7.5,3.8)--(9,3.8);
                                \draw[black, ultra thick, -latex] (12,3.8)--(13.5,3.8);

                                \draw[black] (1.5,3)--(1.5,2.5);
                                \draw[black] (6,3)--(6,2.5);
                                \draw[black] (10.5,3)--(10.5,2.5);
                                \draw[black] (15,3)--(15,0.5);
   

                                \draw[draw=black, thin, fill=white, ] (0,1) rectangle ++(3,1.5) node[align=center, pos=.5, text width=2.5cm] {\baselineskip=12pt \textbf{(1) Error magnitude}\par};
                                
                                \draw[draw=black, thin, fill=white, ] (4.5,1) rectangle ++(3,1.5) node[align=center, pos=.5, text width=2.5cm] {\baselineskip=12pt \textbf{(2) Single decision regret}\par};
                                
                                \draw[draw=black, thin, fill=white] (9,1) rectangle ++(3,1.5) node[align=center, pos=.5, text width=2.5cm] {\baselineskip=12pt \textbf{(3) Decision rate}\par};
                                
                                \draw[draw=black, thin, fill=white] (13.5,-1) rectangle ++(3,1.5) node[align=center, pos=.5, text width=2.5cm] {\baselineskip=12pt \textbf{(4) Weighted error ratio}\par};


                                \node at (-3, 1.9) {\textcolor{black}{Disaggregated}};
                                \node at (-3, 1.6) {\textcolor{black}{metrics}};

                                \node at (-3, -0.2) {\textcolor{black}{Aggregated}};
                                \node at (-3, -0.5) {\textcolor{black}{metrics}};
                                
                               \end{tikzpicture}
}
\end{figure}

The basic idea of B1 is to define metrics for each step in
the chain of influencing factors (\ref{fig: causal_chain}) behind the losses in objective value observed when following a certain policy. By this, we aim at identifying the regions of the state space that are especially relevant regarding the respective overall objective value loss. Therefore, we analyze the occurrence, the sign, and the magnitude of OC approximation errors in the respective states. Then, by suitable visualizations of the metrics, we compare them over various settings of problem parameter values (in the following referred to as \textit{settings}) resembling different real-world i-DMVRPs. Based on that, we classify fundamental types of approximation errors, i.e., OC approximation errors that a broad variety of real-world i-DMVRPs are prone to. Therewith, we can eventually explain the performance of the considered policy. Both the metrics and the respective visualizations are generally valid, i.e., can be applied to any policies derived from different OC approximation approaches. In the following, we first describe the chain of influencing factors between OC approximation error and objective value loss. Afterwards, we describe the metrics we use to quantify each step in this chain of influencing factors and, finally, we propose visualizations of these metrics. \\[0.1cm] 
\textbf{Chain of influencing factors}\\[0.1cm]
At the beginning of the chain of influencing factors, there is an approximation error in a certain state $s_{t-1}$, that could either be an underestimation or an overestimation of the true OC. Depending on the actual magnitude of such an approximation error, the magnitude of the true OC, i.e., $\Delta V_t(s_{t-1}, c)$, and the immediate reward $r_c$, this error can but not necessarily must result in a suboptimal decision.\\[0.1cm] 
Generally, a suboptimal decision in a certain state $s_{t-1}$ can either yield less immediate reward than the optimal decision, transition the system to a lower-valued state than the optimal decision, or both. However, the respective negative effect on the objective value itself can vary from barely notable to considerable.\\[0.1cm] 
Whether the chain of influencing factors continues further, depends on the likelihood that $s_{t-1}$ is encountered and the respective suboptimal decision is made when following the policy under consideration.\\[0.1cm]
We now define disaggregated metrics to quantify ``how bad" an OC approximation error is, ``how wrong" the resulting decision is, and also, ``how likely" this decision is. Additionally, we define a fourth metric that captures the aggregated overall impact of OC underestimations or OC overestimations on the objective value.\\[0.1cm]
\noindent\textbf{Metrics}\\[0.1cm]
Figure \ref{fig: metrics_causal_chain} shows which of the metrics presented in the following corresponds to which step in the previously described chain of influencing factors between OC approximation error and objective value loss. Please note, we assume that we examine an OC approximation relative to the true OC, i.e., a suboptimal policy relative to the optimal policy. 

\begin{figure}[H]
    \begin{center}
          \caption{Weighted error ratio - exemplary for four settings}\label{fig: errorTypeRatio_example}  
          \vspace{0.1cm}
          \includegraphics[width=0.5\textwidth]{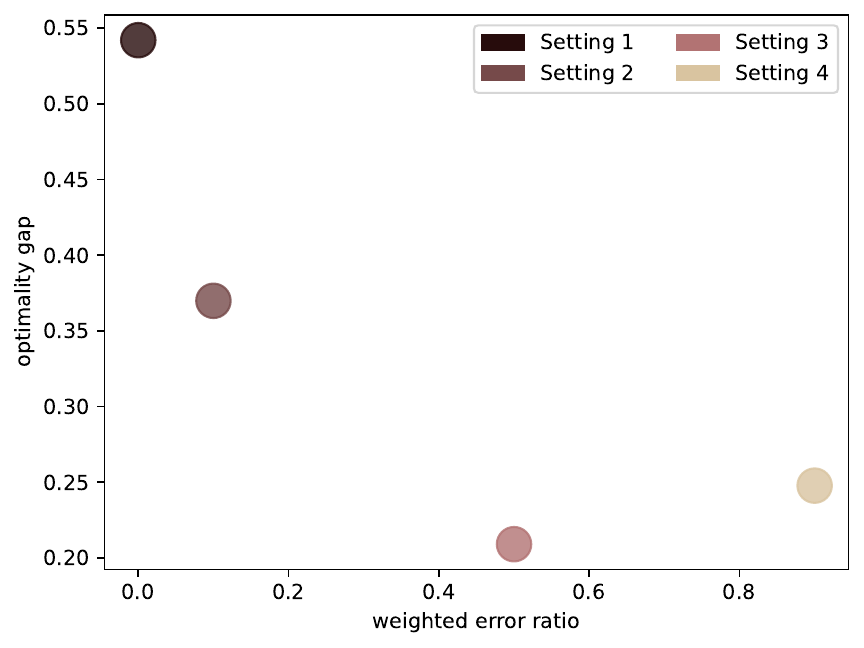}
  \end{center}
\end{figure}

\noindent \textit{(1) Error magnitude --} In general, an OC approximation can be inaccurate in both directions. By separately evaluating \textit{overestimation} and \textit{underestimation}, we can analyze whether the consequences are different. Hence, we calculate the magnitude of overestimation errors $e^{o}(s_{t-1},c)$ and underestimation errors $e^{u}(s_{t-1},c)$, separately, for all individual states $s_{t-1}$ and request arrivals $c$ as follows:
\begin{equation}\label{eq: error_over)}
    e^{o}(s_{t-1},c)=\Delta \tilde{V}_{t}(s_{t-1}, c)-\Delta V_{t}(s_{t-1}, c),
\end{equation}
\begin{equation}\label{eq: error_under)}
    e^{u}(s_{t-1},c)=\Delta V_{t}(s_{t-1}, c)-\Delta \tilde{V}_{t}(s_{t-1}, c).
\end{equation}
This metric allows us to identify regions of the state space, where an approximation systematically overestimates ($e^{o}(s_{t-1},c)>0$) or underestimates ($e^{u}(s_{t-1},c)>0$) the true OC. Hence, it provides information about where approximation errors originate and how strongly the chain of influencing factors is triggered. \\[0.1cm]
\noindent
\textit{(2) Single decision regret --} As is well-known in revenue management (e.g., \cite{talluri2004theory}), an OC approximation error in itself is not problematic because the resulting decision may still be fairly accurate or even optimal. As previously described, the chain of influencing factors only continues if there is a suboptimal decision. Hence, to quantify the ``suboptimality" of a single decision when a request of type $c$ arrives in state $s_{t-1}$, we introduce the metric \textit{single decision regret}, denoted as $\delta(s_{t-1},c)$. It computes the overall reward difference between a single decision based on a, potentially wrong, OC approximation and the optimal decision. In less technical terms, to isolate the regret of one single decision, all future decisions from decision epoch $t+1$ onward are assumed to be made based on the optimal policy in both cases. 
Then, again depending on the observed underlying error magnitude $e^{o}(s_{t-1},c)$ and $e^{u}(s_{t-1},c)$, we can distinguish between overestimation regret and underestimation regret, even though both are calculated based on the same expression:
\begin{equation}\label{eq: regret_definition)}
    \delta(s_{t-1},c)=g_t^{*}(s_{t-1}, c) \cdot \bigl(r_c-\Delta V_{t}(s_{t-1}, c)\bigr)-\tilde{g}_t(s_{t-1}, c) \cdot \bigl(r_c-\Delta V_{t}(s_{t-1}, c)\bigr),
\end{equation}
with $g^*_t(s_{t-1}, c)$ denoting the optimal demand control decision and $\tilde{g}_t(s_{t-1}, c)$ denoting the demand control decision when following the policy under consideration. Then, we define the overestimation regret as:
\begin{equation}\label{eq: regret_over)}
    \delta^{o}(s_{t-1},c)= \left\{
    \begin{array}{ll}
    \delta(s_{t-1},c), & e^{o}(s_{t-1},c)>0 \\
    0, & \, \textrm{otherwise,} \\
    \end{array}
    \right.
\end{equation}
and the underestimation regret as:
\begin{equation}\label{eq: regret_under)}
    \delta^{u}(s_{t-1},c)= \left\{
    \begin{array}{ll}
    \delta(s_{t-1},c), & e^{u}(s_{t-1},c)>0 \\
    0, & \, \textrm{otherwise.} \\
    \end{array}
    \right.
\end{equation}
With this metric, we can assess whether an overestimation error or underestimation error leads to a suboptimal decision, and by which amount it causes the objective value to deteriorate assuming optimal decisions over the remaining booking process. Please note, for ease of readability, in the following, we refer to the single decision regret as \textit{regret}.\\[0.1cm]
\noindent
\textit{(3) Decision rate --} As a third step in the chain of influencing factors, the relevance of a suboptimal decision must be considered. It depends on how likely it is to visit the state in which the decision is made. We measure the likelihood in the form of the decision rate $P(s_{t-1},c)$, which denotes the probability that a policy $\Tilde{\pi}$ visits state $s_{t-1}$ and decides on the acceptance/rejection of a customer request of type $c$ at decision epoch $t$. To calculate it, we simulate decision-making based on the considered OC approximation for a sufficiently high number of drawn sample paths. This metric provides information about the relevant areas of the state space, and thus, to what extent the regret in a certain state impacts the objective value.\\[0.1cm]
\textit{(4) Weighted error ratio --} Additionally to the metrics involved in our chain of influencing factors, we propose an aggregate metric to approximate the share of the loss in objective value that is caused by overestimation and underestimation, respectively. We refer to this metric as \textit{weighted error ratio} and define it for a certain setting as follows:
\begin{equation}\label{eq: error type ratio)}
    E = \frac{\sum_{s \in \mathcal{S}} \sum_{c \in C} \delta^{o}(s,c) \cdot P(s,c)}{\sum_{s \in \mathcal{S}} \sum_{c \in C} (\delta^{o}(s,c)+\delta^{u}(s,c)) \cdot P(s,c)}
\end{equation}
Thus, a weighted error ratio of $1$ means that the full observed objective value loss of a certain policy is caused by overestimation errors, and a weighted error ratio of $0$ means that only underestimation causes objective value loss.\\[0.1cm]
\textbf{Visualizations}\\[0.1cm]
\begin{figure}[h]
\centering
\footnotesize
\caption{Technique of error characterization -- exemplary for two policies (A and B) applied to the same setting}\label{fig: intro_heatmaps}
	\centering
	\captionsetup{justification=centering}
    \vskip\baselineskip
	\begin{subfigure}{0.8\textwidth}
    \renewcommand\thesubfigure{a}
			\centering
    		\includegraphics[width=\textwidth]{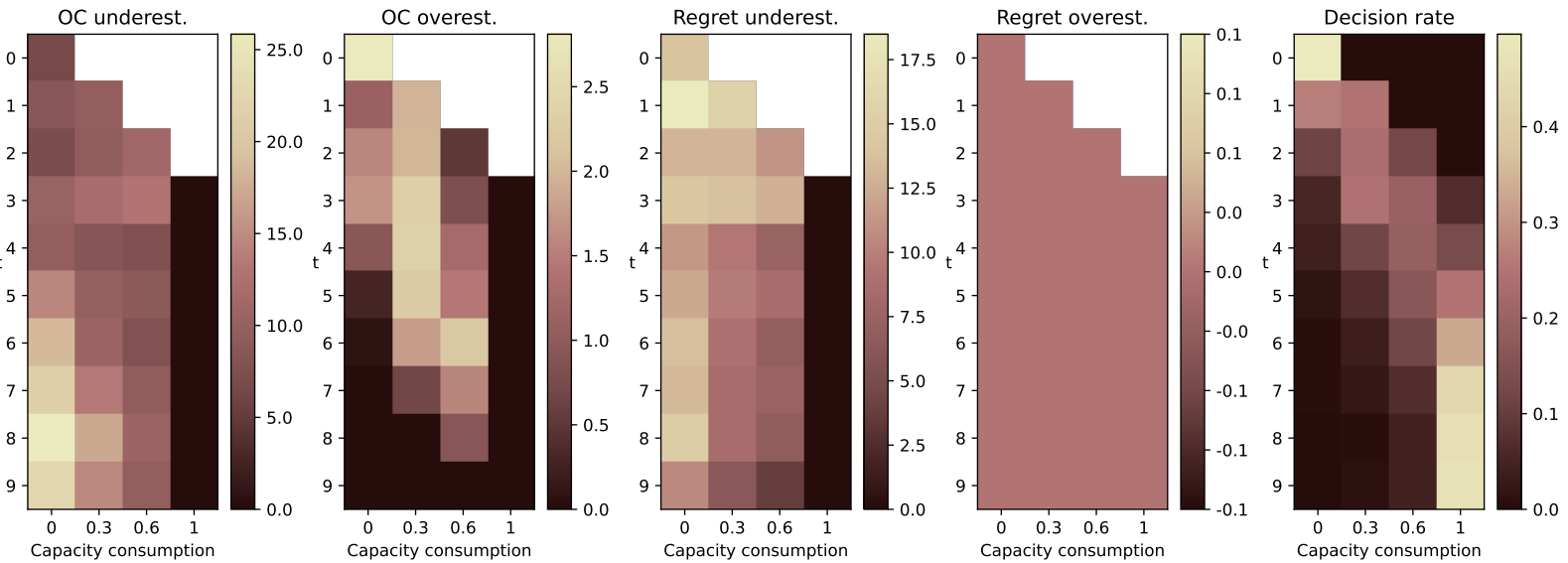}
     \caption[]{Policy A: Average objective value reached = -8.334}
	\end{subfigure}
    \vskip\baselineskip
    \begin{subfigure}{0.8\textwidth}
    \renewcommand\thesubfigure{b}
			\centering
    		\includegraphics[width=\textwidth]{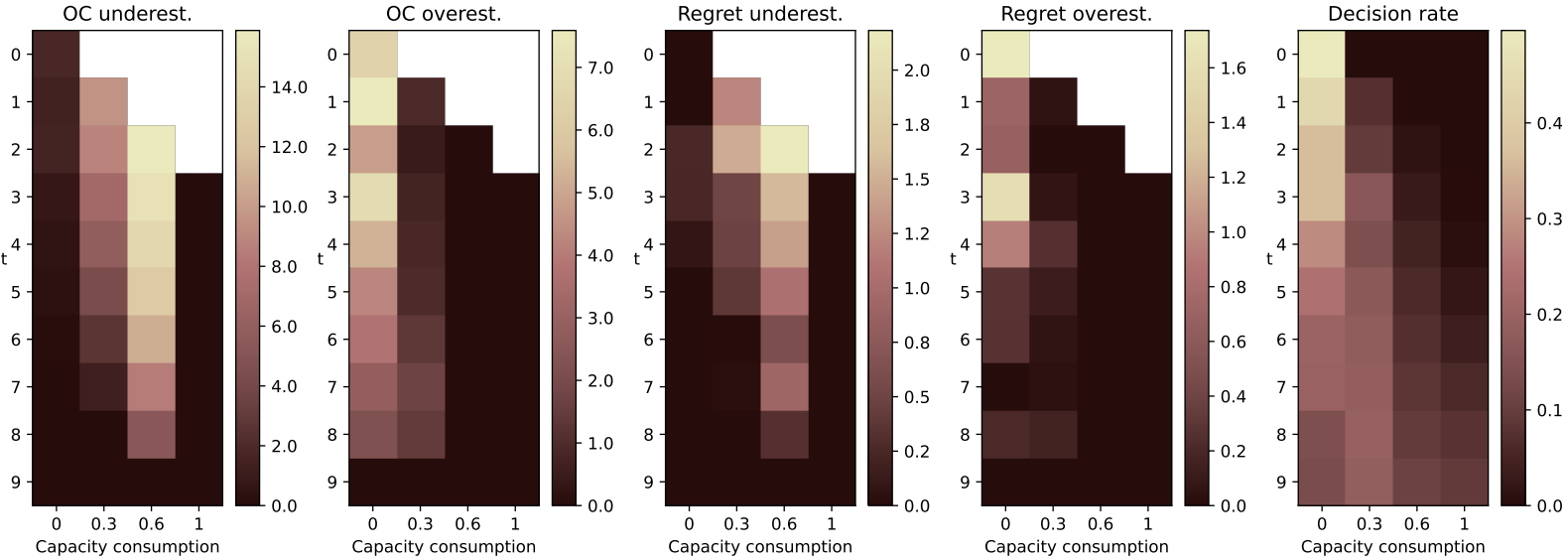}
     \caption[]{Policy B: Average objective value reached = 12.447}
     \end{subfigure}
	\label{fig: intro_heatmaps}
\end{figure}
\noindent \textit{Disaggregated metrics (1)-(3) --} Due to the curses of dimensionality, analyzing the values of the metrics (1)-(3) for individual states as introduced above is impractical. Hence, we propose a state space aggregation for their visualization and the subsequent analysis. For this, we draw on two-dimensional lookup tables, which are well-known from reinforcement learning (e.g., \cite{powell2022reinforcement}). We aggregate states according to the dimensions \textit{decision epoch} $t$ ($y$-axis) and \textit{capacity consumption} ($x$-axis), where we measure the latter in percentage of the available capacity. Since we consider multiple instances per setting, we further aggregate the results of all instances per setting in one look-up table by averaging the respective numbers. In sum, these aggregations allow us to examine the average magnitude of overestimation and underestimation, the associated regret, and the decision rate in the different regions of the state space. Additionally, all instance-specific effects are averaged out. Further, this two-dimensional aggregation allows us to represent the resulting look-up tables in heatmaps, as exemplarily depicted in Figure \ref{fig: intro_heatmaps}.\\[0.1cm] 
Now, it is possible to qualitatively analyze and compare the results for different policies as exemplarily demonstrated in the following: In the example presented in Figure \ref{fig: intro_heatmaps}, we analyze two suboptimal policies A and B that, over the same set of instances, yield entirely different average objective values. However, by analyzing the heatmaps of our metrics (calculated based on the optimal policy), we can now explain these drastic results: Policy A suffers from severe underestimation errors that cause high regret in particularly relevant areas of the state space. Overestimation, in turn, also occurs but does not cause any regret. Contrary, policy B only exhibits mild underestimation with substantially lower regret in rather irrelevant states. Further, for policy B, overestimation also causes regret that is even slightly more relevant. Overall, the aggressive acceptance of (early) customers due to underestimation errors by Policy A causes a severe objective value loss. In comparison, the (slightly too) conservative behavior of Policy B leads to a much better performance. Hence, if we were to develop a policy for an i-DMVRP with this setting structure, we conclude that Policy A is missing crucial information. Hence, in the development process, we can now, e.g., integrate algorithmic elements to tackle the systematic underestimation by Policy A, or draw the conclusion to focus more development effort on Policy B due to its structural advantages.\\[0.1cm]
\noindent\textit{Aggregated metric (4) -- } To visualize and interpret the fourth metric, i.e., the weighted error ratio, we propose a scatter plot, where the result of applying a policy to a certain setting is plotted as a point according to the weighted error ratio on the $x$-axis and the relative optimality gap 
on the $y$-axis (see Figure \ref{fig: errorTypeRatio_example}). This visualization enables us to analyze in which settings a policy is prone to either underestimation or overestimation in combination with the resulting performance impact. \noindent The example given in Figure \ref{fig: errorTypeRatio_example} shows a policy that is mainly affected by underestimation errors in Setting 1 and Setting 2, with a more severe impact in Setting 1. Setting 3 is equally affected by overestimation and underestimation, whereas Setting 4 is mainly affected by overestimation.
\normalsize

\subsection{Building block 2}\label{sec: B2}
To apply the previously presented technique for gaining general insights on i-DMVRPs within our computational study, we now rely on a second explainability technique, namely on reward decomposition (\cite{juozapaitis2019explainable}). More precisely, although B1 can be applied to \textit{any} OC approximation in comparison to the optimal policy and already yields valuable insights, we apply it to \textit{specific} OC approximations that base on the idea of reward decomposition. This comes with two advantages: First, the resulting policies resemble typical approximation approaches from literature that (predominantly) capture one of the reward components. Second, and more importantly, they reveal which reward component is more relevant in a certain setting of a given i-DMVRP and, therewith, hold further explainability potential.\\[0.1cm]
Thus, to define the policies we analyze in Section \ref{sec: computationalStudy}, we draw on the results of \cite{fleckenstein2023concept}. They show that the OC of an i-DMVRP, $\Delta V_t(s_{t-1}, c)$, can be decomposed into two components: DPC, formally denoted as $\Delta R_t(s_{t-1}, c)$, and MCTS, formally denoted as $\Delta F_t (s_{t-1}, c)$. While the former captures the loss of future revenue, i.e., a decision's impact on the positive rewards, the latter measures the respective increase of fulfillment cost, i.e., the impact on the negative rewards. For the formal definition, we first define the expected future revenue of a given interim state $s'_{t-1}$ at decision epoch $t-1$ as $R'_{t-1}(s'_{t-1})$ and the expected future fulfillment cost of a given interim state $s'_{t-1}$ at decision epoch $t-1$ as $F'_{t-1}(s'_{t-1})$. Then, we can define DPC and MCTS as follows: 
\begin{equation}
    \Delta R_{t}(s_{t-1}, c) = R'_{t} (s'_t(0)) - R'_{t}(s'_t(c))
\end{equation}
and
\begin{equation}
    \Delta F_t (s_{t-1}, c) = F'_t (s'_t(0)) - F'_t (s'_t(c)),
\end{equation}
and $\Delta V_t(s_{t-1}, c) = \Delta R_t(s_{t-1}, c) + \Delta F_t(s_{t-1}, c)$ holds.\\[0.1cm]
Based on this finding, we derive and analyze OC approximations that only capture DPC or MCTS, i.e., completely neglect the other component. By considering these most extreme cases, we make sure that any systematic errors resulting from inadequately approximating one component occur as clearly as possible. Further, to reduce random errors related to the varying performance of a heuristic solution algorithm, we formulate the Bellman equations corresponding to our policies and solve them in an exact way by backwards recursion. The DPC-based approximation $\Delta \tilde{R}_{t}(s_{t-1}, c)$ results from: 
\begin{alignat}{5}\label{eq: Bellman_DPC_based}
	\tilde{R}_{t-1}(s_{t-1}) = \sum_{c \in C} \lambda_{c}^{t} \cdot  \smash{\displaystyle \max_{g_{t} \in \mathcal{G}(s_{t-1}, c)}} \biggl(g_{t} \cdot \bigl(r_{c} - \Delta \tilde{R}_{t}(s_{t-1}, c)\bigr)\biggr) + \tilde{R}_{t}^{'}(s'_{t}(0))
\end{alignat}
Since the DPC-based approximation neglects fulfillment cost, i.e., $r_{\phi_{t}(s^{'}_t)} = 0 \quad \forall t=1,...,T$,  the revenue is the only type of reward, and the future cost impact of an acceptance decision is ignored. Solving Equation (\ref{eq: Bellman_DPC_based}), we obtain the DPC policy $\pi^R$, the DPC-based decision  $g^{R}_t(s_{t-1},c)=\smash{\displaystyle \argmax_{g_{t} \in \mathcal{G}(s_{t-1}, c)}} \biggl(g_{t} \cdot \bigl(r_{c} - \Delta \tilde{R}_{t}(s_{t-1}, c)\bigr)\biggr)$ being in state $s_{t-1}$ and observing the arrival of a request of type $c$, and the DPC-based objective value $J^{R}=\tilde{R}_{0}(s_0)$.\\[0.1cm]
The MCTS-based approximation $\Delta \tilde{F}_{t}(s_{t-1}, c)$ results from: 
\begin{equation}\label{eq: Bellman_MCTS_based}
 	\tilde{F}_{t-1}(s_{t-1}) =  \sum_{c \in C} \lambda_{c}^{t} \cdot \smash{\displaystyle \argmax_{g_{t} \in \mathcal{G}(s_{t-1}, c)}} \biggl(g_{t} \cdot \bigl(r_{c} - \Delta \tilde{F}_{t}(s_{t-1}, c)\bigr)\biggr) \cdot \bigl(- \Delta \tilde{F}_{t}(s_{t-1}, c)\bigr)
     + \tilde{F}_{t}^{'}(s'_{t}(0)). 
\end{equation}
Since the MCTS-based approximation neglects displacement cost, formulating the Bellman equation of this approximation requires an $argmax(\cdot)$ operator. This prevents the revenue from being included in the state value while still comparing it to the future cost impact for deciding on each single request's acceptance. Then, the result of the $argmax(\cdot)$ operator, which encodes the binary demand control decision, is multiplied with the future cost impact, i.e., the MCTS. Solving Equation (\ref{eq: Bellman_MCTS_based}), we obtain the MCTS policy $\pi^F$, the MCTS-based decision  $g^{F}_t(s_{t-1},c)=\smash{\displaystyle \argmax_{g_{t} \in \mathcal{G}(s_{t-1}, c)}} \biggl(g_{t} \cdot \bigl(r_{c} - \Delta \tilde{F}_{t}(s_{t-1}, c)\bigr)\biggr)$ being in state $s_{t-1}$ and observing the arrival of a request of type $c$, and the MCTS-based objective value $J^{F}=\tilde{F}_{0}
(s_0)$.
    \section{Computational study}\label{sec: computationalStudy}
We now apply the explainability technique consisting of B1 and B2 to a generic i-DMVRP as introduced in Section \ref{sec: OCforiDMVRP}. The aim of this numerical analysis is to identify and characterize fundamental types of approximation errors that can occur in any real-world i-DMVRP. In Section \ref{sec: experimentalDesign}, we first present the experimental design, the tested parameter settings, and the benchmarks that we consider next to the already introduced DPC-based and MCTS-based approximations. Then, in Section \ref{sec: errors}, we characterize the observed types of approximation errors. Finally, we investigate the impact of these types of errors on the objective value in Section \ref{sec: performanceImpact} to further refine the characterization.
\subsection{Experimental design}\label{sec: experimentalDesign}
Overall, we draw on the full-factorial study design proposed in \cite{fleckenstein2023concept}, which consists of $66$ different settings of an i-DMVRP with disjoint booking horizon and service horizon and pure accept/reject decisions. The integrated VRP is a distance-constrained, capacitated VRP in all settings. Further, all settings have in common that we assume a single fulfillment vehicle and a booking horizon of $T=10$ potential decision epochs with $10$ potentially arriving customer requests such that at most one customer request arrives per decision epoch. \\[0.1cm]
The settings differ in the following parameters:  \textit{location distribution}, \textit{revenue distribution}, general \textit{profitability} of a setting, and binding \textit{capacity constraints}. The former two parameters are customer-related, the latter two are provider-related. In detail, the following parameter values are possible:\\[0.1cm]
\noindent\textit{Location distribution --} Regarding the customers' location distribution, we consider two realizations: A stream of customer request locations $l_c$ with $c = 1,...,10$ are either drawn from (1) a uniform distribution over a line segment with length $50$ in the interval $[-25, 25]$ representing a single urban area, or (2) in random order from two truncated normal distributions with means $-10$ and $20$ and the same standard deviation of $2.5$ to generate two equal-sized clusters representing, e.g., two villages in a rural area. The first location distribution is referred to as \textit{unif} for uniform distribution. The second location distribution is referred to as \textit{clust} since there is two clusters of customers.\\[0.1cm]
\textit{Revenue distribution --} Regarding the customers' revenue distribution, we consider four realizations: Besides (1) homogeneous customer streams with all revenues equal to $15$ monetary units referred to as \textit{homog}, (2) heterogeneous customer streams are generated by randomly assigning a revenue of $25$ monetary units to $30\%$ of the customers in a stream of customer requests. The other $70\%$ of the customers are assigned revenues equal to $15$ monetary units. If heterogeneous customers are considered in a setting, the high-revenue customers can either (2.1) strictly arrive in the beginning (referred to as \textit{h-b-l} for high-before-low), (2.2) randomly (referred to as (\textit{rand})), or (2.3) strictly in the end (referred to as \textit{l-b-h} for low-before-high) of the booking horizon. Further, for each of the realizations (2.1)-(2.3), an additional clustered setting is considered in which all high-revenue customers are located in the distant cluster. We refer to these settings as \textit{clust\_sort}. Also, if required (as in h-b-l and l-b-h settings), we sort the drawn customer streams.\\[0.1cm]
These customer-related parameter realizations yield $11$ meaningful combinations (referred to as \textit{customer settings}), as depicted as leaf nodes in Figure \ref{fig: settings} in Appendix A. Each of these settings is then considered six times, according to six combinations of the two provider-related parameters' realizations that we explain in the following.\\[{0.1cm}]
\textit{Profitability --} To vary the general profitability of the settings, we modify the routing cost factor. More precisely, each customer setting is considered three times with different routing cost factors of $0.2$, $0.6$, or $1$ monetary units per distance unit. We refer to these settings by addressing the profitability as \textit{high}, \textit{med}, or \textit{low}.\\[0.1cm]
\textit{Capacity constraints --} Regarding the logistical capacity constraints, we assume two different realizations. We either limit the route length to $50$ length units and refer to these settings as \textit{distance-constrained} (\textit{dist}), or we limit the physical capacity of the fulfillment vehicle to $3$ units. In the latter case, we assume unit demand for all customers and refer to the respective settings as \textit{load-constrained} (\textit{load}).\\[0.1cm]
For each of the resulting $66$ settings, we consider $50$ different instances in our computational study. Each instance is defined by an individual customer stream of $10$ customers, sampled according to the above-mentioned customer-related parameter values in advance. Of these customer streams, each customer then places a request with probability $\lambda_c^t=0.5$ if $c = t$ and $\lambda_c^t=0$, otherwise. This is the only source of stochasticity once an instance is fully specified. We apply the policies presented in Section \ref{sec: B2} to each instance, calculate the metrics introduced in Section \ref{sec: B1}, and average the respective results over all $50$ instances per setting to derive the setting-specific results we report.\\[0.1cm]
Note, the choice of such a basic problem is deliberate, and typically done in literature, to obtain findings that are valid for a broad variety of real-world problems (e.g., \cite{ulmer2020meso}).\\[0.1cm] 
\textit{Benchmark policies -- }As a reference for evaluating the performance of the previously introduced DPC- and MCTS-based approximations, we mainly consider two benchmark approaches, which we also apply to all instances as previously described. First, we solve Equation (\ref{eq: Bellman_iDMVRPs_reform_OC}) to compute the true OC $\Delta V_{t}(s_t)$. Thereby, we also obtain the optimal policy $\pi^{*}$ and the optimal objective value $J^{*}=V_{0}(s_{0})$ as well as the optimal decision $g_{t}^{*}(s_{t-1},c)=\smash{\displaystyle \argmax_{g_{t} \in \mathcal{G}(s_{t-1}, c)}} \biggl(g_{t} \cdot \bigl(r_{c} - \Delta V_{t}(s_{t-1}, c)\bigr)\biggr)$. \\[0.1cm]
Second, we consider a myopic OC approximation due to its high relevance as a benchmark policy in the i-DMVRP literature (e.g., \cite{arian2022joint}, \cite{klein2021dynamic}, or \cite{yang2016choice}). Instead of a state value difference, the respective OC approximation is defined as the insertion cost into the current myopic route plan, i.e., a route plan based on only confirmed customer orders.\\[0.1cm]
The objective values of all policies are depicted in Figure \ref{fig: profilLines} in Appendix B.
\subsection{Identification of fundamental approximation errors}\label{sec: errors}
In the following, we identify and characterize the fundamental types of approximation errors that we observe for the DPC policy and the MCTS policy. This analysis is based on the heatmaps introduced in Section \ref{sec: B1}. Since it is not possible to include the full set of heatmaps for all settings in the paper at hand, we provide two types of supplementary material for the interested reader. First, we present carefully selected heatmaps in Appendix C. These are intended to be used as "textbook" examples that show a certain error and its characteristics particularly clearly. Second, to allow for full reproducibility, the complete set of heatmaps can be viewed or downloaded at \href{https://zenodo.org/records/14197957?token=eyJhbGciOiJIUzUxMiJ9.eyJpZCI6IjViYzRkY2UxLTk0M2YtNDlhOS05MzkwLWM3ZjI5OTM3NzRhOSIsImRhdGEiOnt9LCJyYW5kb20iOiJmMGE0NGFjNDA3ZGU4NDM5NWY1NjU5OGEzZjY4M2YxZCJ9.XeOM-YyBFnBIRi_0sLDyNR2ybKNk4T5picbMQxzvgkErE6t6uoG-yJgZ-ADmR6-flZOF7Qk9UdjGS1aYwkqjww}{zenodo.org/full-set-of-heatmaps}.\\[0.1cm] 
When discussing an error type's characteristics in the following, especially regarding the influence of setting parameters, we only mention those error types that are reasonably pronounced and occur over many different settings. Thereby, we ensure generalizability beyond the basic settings we consider. Then, we explain the prevalence of these error types in the different settings as well as their interplay. \\[0.1cm] 
\textbf{DPC policy}\\[0.1cm] 
The DPC policy lacks information on how well a request can be consolidated with other orders to a profitable fulfillment tour. It can only observe this information indirectly if it is reflected in the request's capacity consumption, and thus, affects the displacement of future revenue. Apart from that, the policy assumes perfect consolidation of orders, i.e., MCTS of zero for each request. In the following, we introduce the three distinctive types of approximation errors resulting from this DPC policy. More precisely, we discuss two types of underestimation errors and one type of overestimation error, that occur when applying the DPC policy.\\[0.1cm]
\textit{\underline{Underestimation error type 1:} neglecting better consolidation of future requests}\\[0.1cm]
This error occurs when the acceptance of the current request causes the displacement of future requests with a similar revenue that can be consolidated better with other future requests or orders already received. This type of underestimation error becomes smaller over time as there are less opportunities for collecting orders that allow better consolidation (see Figure \ref{fig: HM0a}). In the case that future requests have a higher revenue than the current request, as especially visible in our \textit{l-b-h} settings, the policy protects a certain amount of capacity. This reduces the error's regret since the DPC are sufficiently high to correctly cause a rejection (compare Figure \ref{fig: HM0a} and Figure \ref{fig: HM0b}). The regret also becomes smaller with more correlation between consolidation and capacity consumption, as observable in our \textit{dist} constrained settings (compare Figure \ref{fig: HM0b} and Figure \ref{fig: HM0c}). In general, the error occurs in a broad region of the state space with a relatively high decision rate (see Figure \ref{fig: HM0}).\\[0.1cm]
\textit{\underline{Underestimation error type 2:} neglecting inherent request unprofitability}\\[0.1cm]
This error mainly occurs when the current request is neither profitable based on consolidation with the orders already received nor is it expected to become profitable taking future consolidation opportunities into account. Hence, the effect becomes stronger over time with shrinking future consolidation opportunities and tends to occur in states with low capacity consumption where not much consolidation is already established (see Figure \ref{fig: HM0a}). The associated regret is only dampened when high-revenue requests arrive in the affected states as especially observable in our \textit{l-b-h} settings (see Figure \ref{fig: HM0b}). Compared to underestimation error type 1, considering the decision rate reveals that this error type has much less impact because the region of late states with low capacity consumption is hardly visited in any of our settings (see Figure \ref{fig: HM0a}-\ref{fig: HM0c}). However, if revenues reflect the distance from the depot as in our \textit{clust\_sort} settings, the error can occur earlier, i.e., in states with a higher decision rate (see Figure \ref{fig: HM0d}).\\[0.1cm] 
In extreme cases, the underestimation due to these two types of errors can become large enough that an acceptance decision is made in every state.
\\[0.1cm]
\textit{\underline{Overestimation error:} protecting capacity for high-revenue demand with poor consolidation}\\[0.1cm] 
The missing information on consolidation opportunities can also lead to an OC overestimation. This error occurs when the policy anticipates demand that has indeed higher revenue but cannot be consolidated as well as the current request. In such a state, it is better to accept the current customer request despite its lower revenue than to reserve capacity for the higher-revenue demand that is less profitable due to poor consolidation. Consequently, a heterogeneous revenue distribution that is not strictly \textit{h-b-l} is prone to this error (compare Figure \ref{fig: HM1a} and Figure \ref{fig: HM1b}). We then can locate the error in all states early in the booking horizon but it only leads to regret in states in which displacement effects are sufficiently strong, i.e., states with high capacity consumption in case of \textit{load}-constrained settings or states with low capacity consumption in case of \textit{dist}-constrained settings (compare Figure \ref{fig: HM1a} and Figure \ref{fig: HM1c}). It remains constant over time until the high-revenue demand is expected to realize. Its magnitude is proportional to the mismatch between revenue and consolidation opportunities offered by the demand, or in other words, between revenue displacement observed by the policy and the actual profit displacement. A particularly strong occurrence can be observed in \textit{low}, \textit{clust\_sort} settings in which the high-revenue requests also cause the highest routing cost (see Figure \ref{fig: HM1d}). In the region of the state space, in which this error type occurs, the decision rate is comparatively high (see Figure \ref{fig: HM1}).\\[0.1cm]
\textbf{MCTS policy}\\[0.1cm] 
The MCTS policy has information about all parameters of expected future requests including their revenue. However, it only uses the information about the revenue when making a decision on an individual request, as revenues generally do not enter its value function. Thus, the policy can only anticipate which future requests it will accept in case of any realization of future demand and the resulting fulfillment cost but not the associated cumulative revenue. For this policy, we find three distinctive estimation errors.\\[0.1cm]
\textit{\underline{Underestimation error type 1:} neglecting future high-revenue demand}\\[0.1cm]
Since the MCTS policy cannot observe heterogeneity in revenue, it fails to reserve capacity if there is more profitable, high-revenue demand arriving at a later phase of the booking process as especially observable in our \textit{l-b-h} settings. The resulting regret due to earning less revenue per order is roughly proportional to the magnitude of the underestimation error. The error becomes stronger the more capacity is consumed and the earlier a state is (see Figure \ref{fig: HM2a}). The error magnitude is insensitive toward the profitability (compare Figure \ref{fig: HM2a} and Figure \ref{fig: HM2b}). 
\\[0.1cm]
\textit{\underline{Underestimation error type 2:} missing information on the volume of future demand}\\[0.1cm]
Even if the revenues are homogeneous, underestimation occurs when applying the MCTS policy since its OC approximations do not fully reflect the volume of demand to come. Only in the end of the booking horizon, the low volume of demand to come impacts the approximation via the decreasing number of consolidation opportunities. In earlier states, this information can only be derived from the expected revenue to come, and thus, the displacement cost. The consequence of this underestimation is that the policy does not protect capacity for requests offering better consolidation, i.e., are less costly to serve. This can either lead to less efficient routing, which reduces profitability on the cost side, or it may also reduce the number of accepted orders due to the higher capacity consumption, which affects the revenue side. The error and the associated regret occur predominantly in early states (see Figure \ref{fig: HM2c}). As for underestimation error type 1, the error magnitude does not change with profitability (compare Figure \ref{fig: HM2c} and Figure \ref{fig: HM2d}).\\[0.1cm]
Both types of MCTS underestimation errors are usually superimposed on each other. Similar to the type 1 underestimation error of the DPC policy, they occur in a fairly large region of the state space with a high decision rate. In the worst case, the combination of both errors leads to solutions with less accepted orders, less revenue per order, and an inefficient routing. \\[0.1cm]
\noindent \textit{\underline{Overestimation error}: wrong cost attribution}\\[0.1cm]
The missing information on future revenues can also lead to overestimation errors by the MCTS policy. This is caused by the following mechanism: If none or only very few orders are confirmed already, accepting a request momentarily entails a high increase in routing cost compared to rejecting it. If the instance is sufficiently profitable, anticipating optimal future decision-making would reveal that the final cost difference between accepting the current request and rejecting it would not be as high because of future consolidation. The MCTS policy, however, anticipates that in the reject-case, no or only few additional requests would be accepted, again because of their momentarily high increase in routing cost compared to a rejection. In summary, the policy wrongly attributes high routing cost to early arriving requests. \\[0.1cm]
The error and the resulting regret become more severe with lower profitability (compare Figure \ref{fig: HM3a} and Figure \ref{fig: HM3b} or Figure \ref{fig: HM3c} and Figure \ref{fig: HM3d}), and its recursive character makes this error particularly harmful. At worst, it can lead to the policy stalling in states with no or few orders, which becomes apparent from the plots showing the decision rate (see Figure \ref{fig: HM3b} and Figure \ref{fig: HM3d}). The reason is that suboptimal rejections due to overestimation lead to successor states in which the overestimation error is likely to occur again. Thus, the error occurs mainly in states with no or a very short tentative route and becomes more severe with decreasing profitability. By contrast, error and regret are reduced by heterogeneous revenues. In addition, the region of high decision rates shifts away from the states with low capacity consumption, where the error occurs, which indicates that the vicious cycle is broken early in the booking process. Responsible for this are high-revenue requests that are accepted despite the overestimation (compare Figure \ref{fig: HM4a} and Figure \ref{fig: HM4b}). These orders then serve as seed customers to establish consolidation independent from their distribution over time. Similarly, a uniform distribution of locations (\textit{unif}) can reduce the error due to orders located close to the depot serving as seed customers (compare Figure \ref{fig: HM4a} and Figure \ref{fig: HM4c})
\subsection{Resulting performance impact}\label{sec: performanceImpact}
\begin{figure}[ht]
\centering
\footnotesize
\caption{Weighted error ratio and average optimality gap}\label{fig: errorTypeRatio}
	\centering
	\captionsetup{justification=centering}
    \vskip\baselineskip
	\begin{subfigure}{\textwidth}
    \renewcommand\thesubfigure{a}
        \begin{subfigure}{0.5\textwidth}
    		\renewcommand\thesubfigure{a1}
			\centering
    		\includegraphics[width=\textwidth]{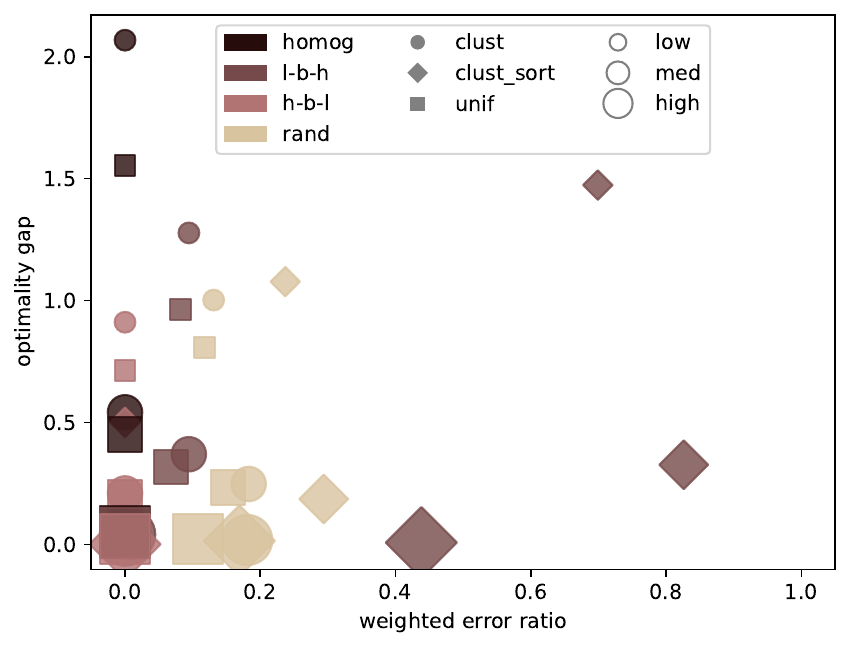}
            \caption{$load$}
        \end{subfigure}
        \begin{subfigure}{0.5\textwidth}
    		\renewcommand\thesubfigure{a2}
			\centering
    		\includegraphics[width=\textwidth]{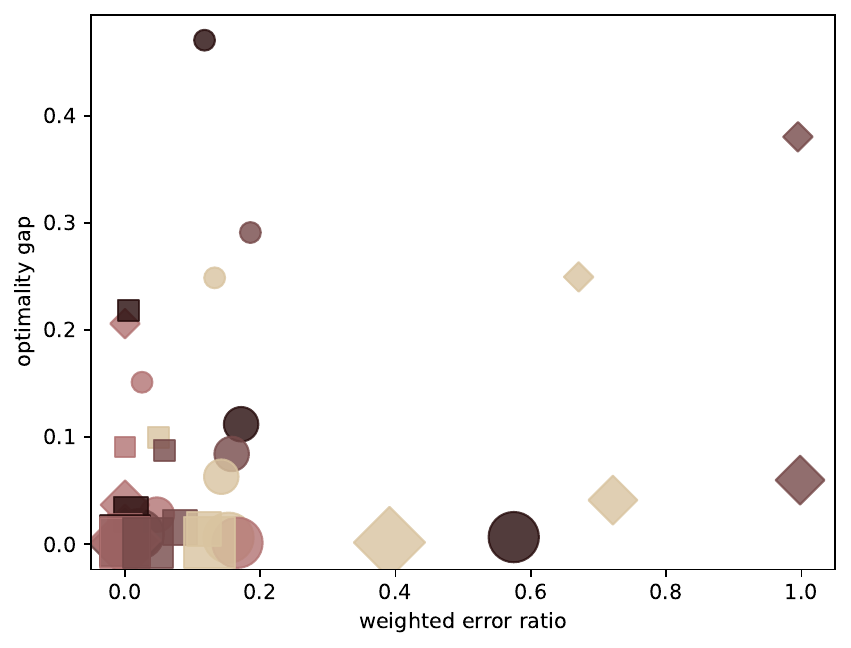}
            \caption{$dist$}
        \end{subfigure}
     \caption[]{DPC policy}
	\end{subfigure}
    \vskip\baselineskip
	\begin{subfigure}{\textwidth}
    \renewcommand\thesubfigure{b}
        \begin{subfigure}{0.5\textwidth}
    		\renewcommand\thesubfigure{b1}
			\centering
    		\includegraphics[width=\textwidth]{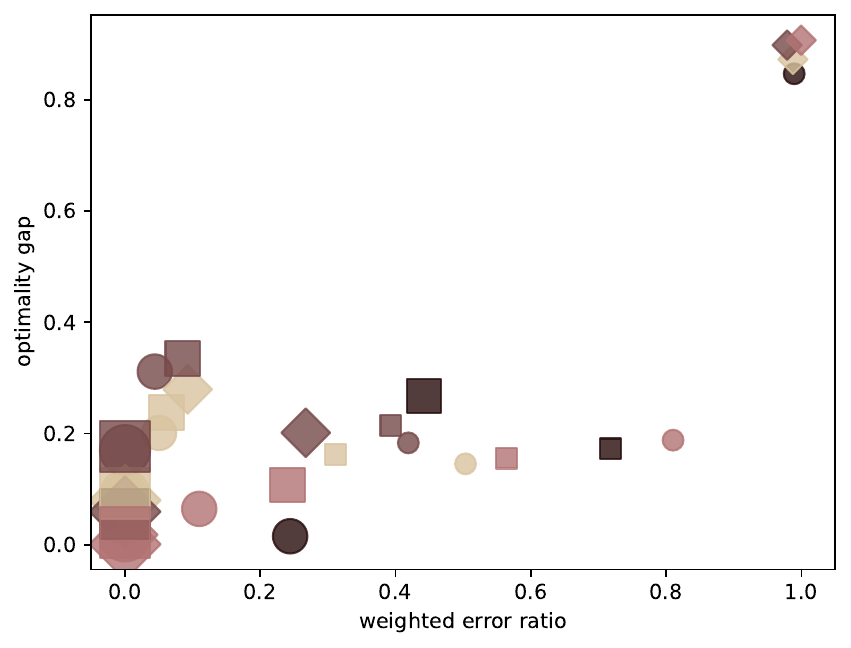}
            \caption{$load$}
        \end{subfigure}
        \begin{subfigure}{0.5\textwidth}
    		\renewcommand\thesubfigure{b2}
			\centering
    		\includegraphics[width=\textwidth]{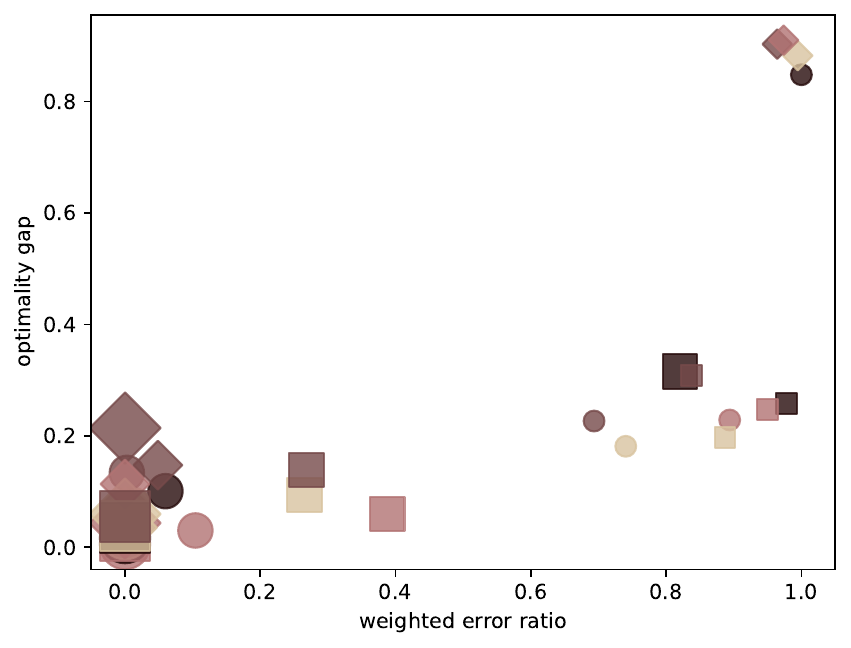}
            \caption{$dist$}
        \end{subfigure}
     \caption[]{MCTS policy}
	\end{subfigure}
	\label{fig: bubbles}
\end{figure}

Using the weighted error ratio and its visualization introduced in Section \ref{sec: B1}, we now analyze the contribution of underestimation errors and overestimation errors to the losses in objective value relative to the optimal solution. Again, we consider the DPC policy and the MCTS policy in the $66$ different settings. The results are depicted in Figure \ref{fig: errorTypeRatio}. As discussed in the following, there are some clearly observable patterns between the optimality gap, the weighted error ratio, and certain setting parameter values. From these patterns, we can derive insights on how location distribution, revenue distribution, and profitability cause underestimation errors or overestimation errors when applying a given policy and how this affects the optimality gap. However, depending on the policy applied, there may hardly be any patterns, as we observe for our MCTS policy. This shows that there can be complex interactions between underestimation and overestimation errors, e.g., both can offset. 

\noindent\textbf{DPC policy}\\[0.1cm]
When applying the DPC policy, the average optimality gap generally decreases with increasing profitability, but the weighted error ratio stays almost constant with only a slight shift toward underestimation. The type of the binding capacity constraints also has an influence on performance since the average optimality gap is smaller for \textit{dist}-constrained settings, and there is more observations of settings in which the optimality gap results mainly from overestimation. Both scenario parameters combined, the observations reach from an average optimality gap below $1\%$ in \textit{high}-profitability, \textit{dist}-constrained settings, to even negative objective values in some \textit{low}-profitability, \textit{load}-constrained settings. \\[0.1cm]
Generally, in \textit{clust} settings, the average optimality gap tends to be larger than in \textit{unif} settings. Since we only observe a slight shift toward overestimation, this is mainly caused by more severe regret from underestimation errors. If the requests' revenues are proportional to their distance from the depot, i.e., in \textit{clust\_sort} settings, the share of overestimation increases but without a clear impact on the average optimality gap. For \textit{med}-profitability settings, the gap becomes smaller but it increases for \textit{low}-profitability settings. \\[0.1cm]
Regarding the revenue distribution, we find the highest average optimality gap for our \textit{homog} settings and the smallest for \textit{h-b-l} settings. In both types of settings, there is no overestimation in case of \textit{load}-constrained settings and only a small impact of overestimation in \textit{dist}-constrained settings. In case of heterogeneous revenues, i.e., in our \textit{h-b-l}, \textit{l-b-h}, or \textit{rand} settings, the average optimality gap and the share of overestimation decrease the earlier the high revenue arrives, as especially observable in our \textit{h-b-l} settings.\\[0.1cm]
\noindent\textbf{MCTS policy}\\[0.1cm]
For the MCTS policy, the patterns are less clear. In \textit{high}-profitability settings, underestimation is by far the dominant error. Due to the dominating underestimation error, the average optimality gap is higher the more high-revenue demand arrives late, i.e., in \textit{l-b-h} settings, as well as in \textit{load}-constrained settings, where less consolidation is possible. Despite the error ratio shifting toward overestimation with the setting becoming less profitable, this has no coherent impact on the average optimality gap because of the complex interactions of underestimation errors and overestimation errors. E.g., we observe average optimality gaps increasing consistently with decreasing profitability in \textit{h-b-l} settings that are \textit{dist}-constrained with \textit{unif} or \textit{clust} locations, and \textit{load}-constrained, \textit{clust\_sort} settings. In \textit{load}-constrained settings with \textit{unif} or \textit{clust} location distribution, however, the average optimality gap decreases between \textit{med}-profitability and \textit{low}-profitability settings due to offsetting errors for \textit{rand} and \textit{l-b-h} settings. Another example for these inconsistencies are settings with \textit{homog} revenues. Here, the average optimality gap is much higher for \textit{unif} settings compared to \textit{clust} settings, when considering \textit{med} profitability. In \textit{low}-profitability settings, we observe an inverse relation. \\[0.1cm]
Regarding the location distribution, the general findings are that \textit{unif} settings tend to have a higher average optimality gap compared to \textit{clust} settings and that \textit{low}-profitability, \textit{clust\_sort} settings are prone to overestimation.\\[0.1cm]
Considering the distribution of revenues, we find that \textit{homog} or \textit{l-b-h} settings tend to have a higher average optimality gap.

    \section{Insights for algorithm selection and algorithm design}\label{sec: managerialInsights}
In this section, we formulate five actionable insights based on our findings that support developers and users of OC approximation approaches in selecting and designing algorithmic components for specific i-DMVRPs. By closely incorporating the key computational results of the existing literature, we now show that our findings explain much of the performance differences observed in literature. Therewith, we also compile evidence for that our results derived from the numerical analysis of a generic i-DMVRP are fundamental to the entire family of i-DMVRPs.\\[0.1cm]
\textbf{Opportunity cost approximation errors can be grouped into a few distinct types}:
Neglecting one of the OC components causes systematic errors that can be grouped into a small number of fundamental types consistently observable across a variety of settings. Furthermore, we find many patterns regarding their occurrence in certain regions of the state space and in different settings. While some of those patterns were already suspected in existing literature based on the decision-making behavior of the respective solution approaches, we are the first to provide direct, numerical evidence to characterize these patterns and are also able to identify entirely novel ones. E.g., the existence of underestimation errors when applying an MCTS policy has been discussed but not conclusively proven by \cite{mackert2019choice} and \cite{yang2017approximate}. Overestimation, however, has only been briefly mentioned by one author (\cite{mackert2019choice}). Overall, our main finding is that neglecting either component can lead to both overestimation errors and underestimation errors. As we show in the work at hand, both are systematic in the sense that they occur even if an exact algorithm, i.e., applying backwards recursion to the (modified) Bellman equations (\ref{eq: Bellman_iDMVRPs_reform_OC}), (\ref{eq: Bellman_DPC_based}), and (\ref{eq: Bellman_MCTS_based}), is used to compute the respective OC estimate. \\[0.1cm]
\textbf{Underestimation is the dominant error when neglecting DPC or MCTS}:
The fundamental issue with neglecting DPC or MCTS is that the resulting approximation does not reach the correct absolute level of the true OC since the contribution of the other component is missing. Hence, underestimation is by far more common than overestimation (see Figure \ref{fig: errorTypeRatio}): For the DPC policy, underestimation (weighted error ratio $<0.5$) is dominant in $89.4\%$ of all settings. For the MCTS policy, this occurs in $69.7\%$ of the settings. This dominance of underestimation leads to greedy decision-making overall. The fact that many authors report such greedy behavior in studies with realistic-sized instances shows the general validity of this result. This accounts for both availability control (e.g., \cite{arian2022joint},  \cite{lang2021anticipative}, \cite{mackert2019choice}, and \cite{campbell2005decision}), and dynamic pricing (e.g., \cite{abdollahi2023demand}, \cite{klein2021dynamic}, \cite{klein2018model}, \cite{yang2017approximate}, and \cite{yang2016choice}). In the former case, the policy accepts too many customers or offers too many, or even all of the feasible fulfillment options to each requesting customer resulting in a first-come-first-served decision-making behavior. In the latter case, the average price level is too low and can drop markedly during the booking horizon. In MOD applications, this observation is also known as the "wild goose chase" (\cite{castillo2022matching}). In accordance with our findings, which show the potential severity of the underestimation errors, some works find that the resulting greedy behavior can lead to a worse performance than static pricing (\cite{abdollahi2023demand}, \cite{yang2017approximate}).\\[0.1cm]
Due to the absence of research on DPC policies, there are no direct indications for respective underestimation errors in the literature. However, there is one indirect observation, which can now be better explained in the light of our findings: \cite{koch2020route} observe that average OC increases quite strongly in the end of the booking horizon. This can be attributed to only unpopular fulfillment options remaining available, for which only few orders are collected already. Thus, collecting an additional order for such an option incurs high MCTS. If MCTS is not adequately captured by the OC approximation, this results in severe underestimation errors. \\[0.1cm]
Another important observation from Figure \ref{fig: errorTypeRatio} is that neglecting either MCTS or DPC tends to cause a higher performance loss in \textit{load}-constrained settings than in \textit{dist}-constrained settings. This points toward a link of the two components by the consumption of logistical capacity associated with an order. In settings in which a high capacity consumption causes both strong displacement effects and a high marginal increase of fulfillment cost, DPC and MCTS will both be high, and vice versa. This means that an approximation based on one of the components can still correctly determine the relative ordering of the OC across different requests or different fulfillment options for one request. However, if the absolute value of the OC approximations is too low due to the high relevance of the missing component, this information gain does not come into effect. In \cite{abdollahi2023demand}, we find further evidence for this mechanism. They report that their MCTS policy performs well in terms of the number of collected orders and the cost per order, which indicates that the policy correctly captures the ``right" orders to achieve high consolidation. However, their policy still fails to improve profit due to the general price level being too low. \\[0.1cm]
Regarding the design of solution algorithms, the explicit consideration of both components is the safest way of avoiding structural underestimation. However, this may be associated with a much higher computational effort. Instead, the described findings related to underestimation suggest two low-threshold algorithmic strategies for mitigating its performance loss: 
\begin{itemize}
\item[1.) ] The OC approximation can be raised to the correct level by adding a rough estimate of the other component. An example for this can be found in \cite{yang2016choice}. They introduce a cost penalty as an additional rough DPC estimate if an order cannot be feasibly inserted into the sampled route plan (indicating strong displacement effects). The main challenge with such an approach is the correct adjustment of the rough DPC estimate. \cite{klein2018model} benchmark their own approach, which features a more sophisticated DPC approximation, against the approach by \cite{yang2016choice}, and the results show that its performance does not decline monotonically with a tighter capacity restriction. Based on our results, the reason for this is that the performance depends on how well the penalty is adjusted to result in the correct level of DPC rather than on the general magnitude of displacement effects.
\item[2.) ] The demand control approach can be made more robust against the consequences of underestimation rather than tackling the underestimation error itself. Hence, the goal is to reduce the regret resulting from a given underestimation error. \cite{ulmer2020dynamic} present an example for such an approach. They introduce a basis price that is charged even if the OC approximation would suggest an even lower price. This curtails greedy control behavior resulting from underestimation.
\end{itemize}
\textbf{Neglecting DPC or MCTS can cause severe overestimation errors}:
In contrast to underestimation, which is presumed to cause performance differences in several existing works, there is only one publication discussing possible overestimation errors. \cite{mackert2019choice} observes that using a less accurate approximation of routing cost leads to the policy making less offers on average, which points toward an increase of average OC, i.e., toward a potential overestimation. \\[0.1cm]
With our computational study, we provide definitive proof that overestimation errors as a consequence of neglecting DPC or MCTS exist. In most settings, overestimation contributes less than underestimation to the optimality gap. However, if overestimation is the dominant error, the optimality gap is particularly large in most cases. This is especially true for the MCTS policy, which can completely stall due to the recursive nature of the type of overestimation error it exhibits. At the same time, this recursiveness is what makes the error manageable by providing some form of anticipatory information on future consolidation opportunities. This explains the success of skeletal or sampled route planning that is applied as a well-performing algorithmic component in several existing publications (e.g., \cite{anzenhofer2023analyzing}, \cite{koch2020route}, and \cite{yang2016choice}). However, none of these publications contains a conclusive explanation of why it is beneficial.\\[0.1cm] 
\textbf{The early phase of the booking process is critical}:
Computational experiments in the existing literature suggest that demand control in the early phase of the booking process has a critical influence on solution quality (\cite{anzenhofer2023analyzing}, \cite{campbell2006incentive}). Our computational results support this finding as we observe that the regret of a suboptimal demand control decision decreases over time for most types of overestimation and underestimation errors. Since the effect is observable independent of the revenue distribution, we can derive that the underlying main reason is not the loss in immediate reward but the transition to a state that has a much lower value than the successor state resulting from the optimal decision.\\[0.1cm]
\textbf{Anticipation is not an end in itself}: 
The existing literature studying specific i-DMVRPs consistently emphasizes the importance of anticipation for the performance of solution approaches (\cite{fleckenstein2021recent}). However, our computational study shows that myopic policies can be hard to beat in certain settings (see Figure \ref{fig: profilLines} in Appendix \ref{app: numResults}, e.g., \textit{med}-profitability or \textit{low}-profitability, \textit{load}-constrained settings). This result becomes even more relevant given that, in the existing literature, anticipatory policies only achieve maximal improvements of a around $10-15\%$ over myopic benchmarks for some specific i-DMVRPs (e.g., \cite{azi2012dynamic}, \cite{heitmann2023combining}, \cite{koch2020route}), and that these numbers may even be subject to a negative results bias (\cite{fanelli2012negative}). From our computational study, we can derive some of the underlying reasons:
\begin{itemize}
\item[1.) ] In some settings, an i-DMVRP can be "easy" to solve in general, e.g., \textit{high}-profitability, \textit{h-b-l} settings (see Figure \ref{fig: profilLines}). Highly profitable demand arriving early in the booking process hardly impacts estimation errors but substantially reduces the resulting regret. Therefore, a more accurate OC approximation by an anticipatory policy may not necessarily translate into a reduced regret. 
\item[2.)] Myopic policies benefit from the fact that estimation errors can offset. Therefore, higher overestimation errors compared to an anticipatory policy can, counter-intuitively, be beneficial because the underestimation errors are then compensated better (e.g., see Figure \ref{fig: HM5}). 
\item[3.)] In settings with \textit{low} profitability or low consolidation, overestimation, which myopic policies are prone to, is generally less harmful than underestimation because it leads to a more conservative acceptance behavior. 
\end{itemize}
In summary, we conclude that investing a substantial amount of the scarce computation time at each decision epoch into an anticipatory OC approximation may not be the most efficient approach for certain i-DMVRPs. Instead, a greater performance improvement might be possible by investing it into solving the other complex subproblems of i-DMVRPs, e.g., the demand control subproblem, or the dynamic vehicle routing subproblem.
\section{Conclusion and future research opportunities}\label{sec: conclusion}
With this work, we present a novel post-hoc explainability technique for i-DMVRPs as well as the results of its first systematic application. The proposed explainability technique allows quantifying and visualizing the extent of OC approximation errors, the regret associated with suboptimal decisions, and the rate at which such decisions are made. Applying this technique to a generic i-DMVRP and considering many settings with a broad variety of parameter values and their combinations, we derive a comprehensive identification and analysis of fundamental types of approximation errors. Finally, we show that this knowledge can be used to explain performance differences observed in the existing literature and to guide algorithm selection and algorithm design. In the following, we conclude by discussing how developers and users can benefit from our findings:
\begin{itemize}
\item[1.)] Our computational results provide a guideline for how important an accurate consideration of DPC and MCTS is in certain problem settings, and how likely the different fundamental types of errors are to occur. This provides indications for the selection of a basic solution concept.
\item[2.)] The technique presented in Section \ref{sec: methodology} can be directly applied to small instances of the specific i-DMVRP under consideration to compare variants of the pre-selected solution concept with the optimal policy. Thereby, developers and users can gain detailed insights into which types of estimation errors are most relevant for their problem and make targeted adaptions to their algorithms to tackle them. 
\item[3.)] Providing training for users based on our findings can enable them to better diagnose anomalies in the behavior of policies and react accordingly.
\end{itemize}
Since algorithmic explainability is still an emerging concept that is of high relevance both from a theoretical and a practical perspective, we believe that there is great potential for future research connected to our work:
\begin{itemize}
\item[1.)] It would be interesting to apply post-hoc explainability techniques similar to ours to related problem classes, such as network revenue management problems arising, e.g., in the airline industry (\cite{klein2020review}). In particular, also given the promising results of related research (\cite{bravo2020mining}), we believe that optimal policies for small instances are a valuable source of domain knowledge for problems that cannot be solved to optimality when considering realistic-sized instances.
\item[2.)] Complementary to post-hoc techniques based on OC, we see a similar potential for the development of inherently explainable solution approaches for i-DMVRPs, such as inherently explainable policy function approximations.
\item[3.)] Finally, our numerical results suggest that myopic policies deserve more attention. While they are currently mainly considered for pure benchmarking purposes in comparison to anticipatory approaches, their potential as carefully designed, full-fledged solution approaches should also be investigated more thoroughly. 
\end{itemize}

    \appendix
    \clearpage
\section{Settings}\label{app: settings}
\vspace{-10 pt}
\begin{figure}[ht]
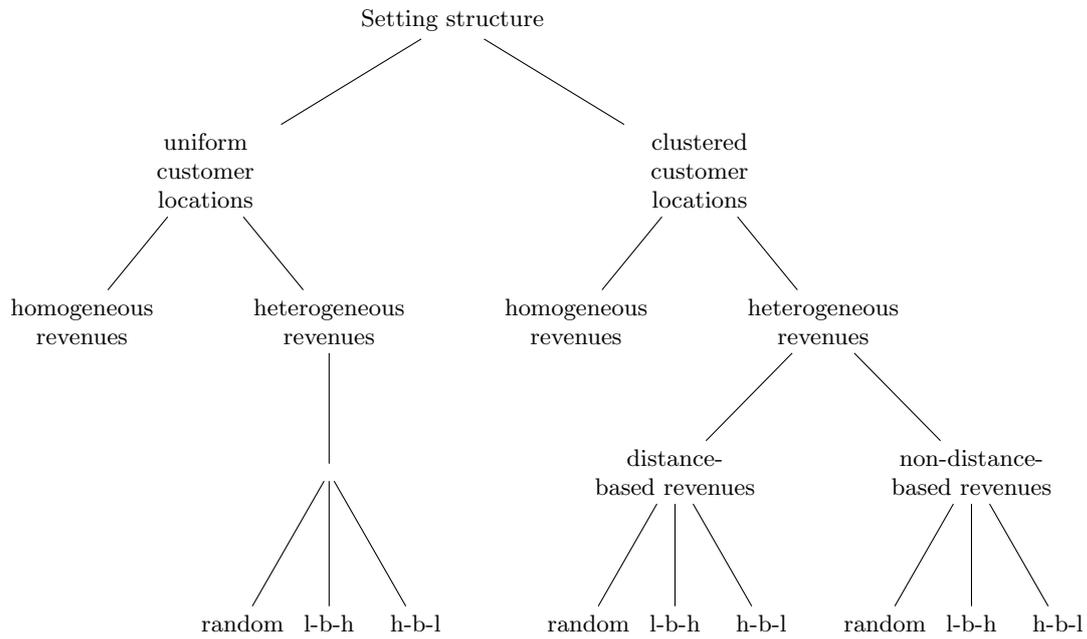

\centering
\footnotesize
\caption{11 customer settings which are solved six times each, for all combinations of routing cost factors $0.2$, $0.6$, $1$ with either binding load constraints or binding distance constraints.}\label{fig: settings}
\vspace{5pt}
\usetikzlibrary {trees}
\tikz [font=\footnotesize,
grow=down, 
level 1/.style={sibling distance=20em},
level 2/.style={sibling distance=10em}, 
level 3/.style={sibling distance=12em},
level 4/.style={sibling distance=3.5em},
level distance=2cm]
\node {Setting structure} 
child {  node[align=center, pos=.5, text width=2.5cm] {uniform \\ customer \\locations}
child { node[align=center, pos=.5, text width=2.5cm] {homogeneous revenues} }
child { node[align=center, pos=.5, text width=2.5cm] {heterogeneous revenues} 
child { node {} child {  node[align=center, pos=.5, text width=2.5cm] {random}} child{ node[align=center, pos=.5, text width=2.5cm] {l-b-h} } child{ node[align=center, pos=.5, text width=2.5cm] {h-b-l} } }
}}
child {  node[align=center, pos=.5, text width=2.5cm] {clustered customer locations}
child {  node[align=center, pos=.5, text width=2.5cm] {homogeneous revenues} }
child {  node[align=center, pos=.5, text width=2.5cm] {heterogeneous revenues} 
child {  node[align=center, pos=.5, text width=2.5cm] {distance-based revenues}child {  node[align=center, pos=.5, text width=2.5cm] {random}} child{ node[align=center, pos=.5, text width=2.5cm] {l-b-h} } child{ node[align=center, pos=.5, text width=2.5cm] {h-b-l} }} child { node[align=center, pos=.5, text width=2.5cm] {non-distance-based revenues} child {  node[align=center, pos=.5, text width=2.5cm] {random}} child{ node[align=center, pos=.5, text width=2.5cm] {l-b-h} } child{ node[align=center, pos=.5, text width=2.5cm] {h-b-l} }}
}};
\end{figure}
\section{Numerical results}\label{app: numResults}
\begin{figure}[H]
\centering
\footnotesize
\caption{Objective values resulting from the different opportunity cost approximations -- Averaged across 50 instances per setting}\label{fig: profilLines}
    \centering
    \vskip\baselineskip
	\captionsetup{justification=centering}
    \begin{subfigure}{0.75\textwidth}
    \renewcommand\thesubfigure{a}
			\centering
    		\includegraphics[width=\textwidth]{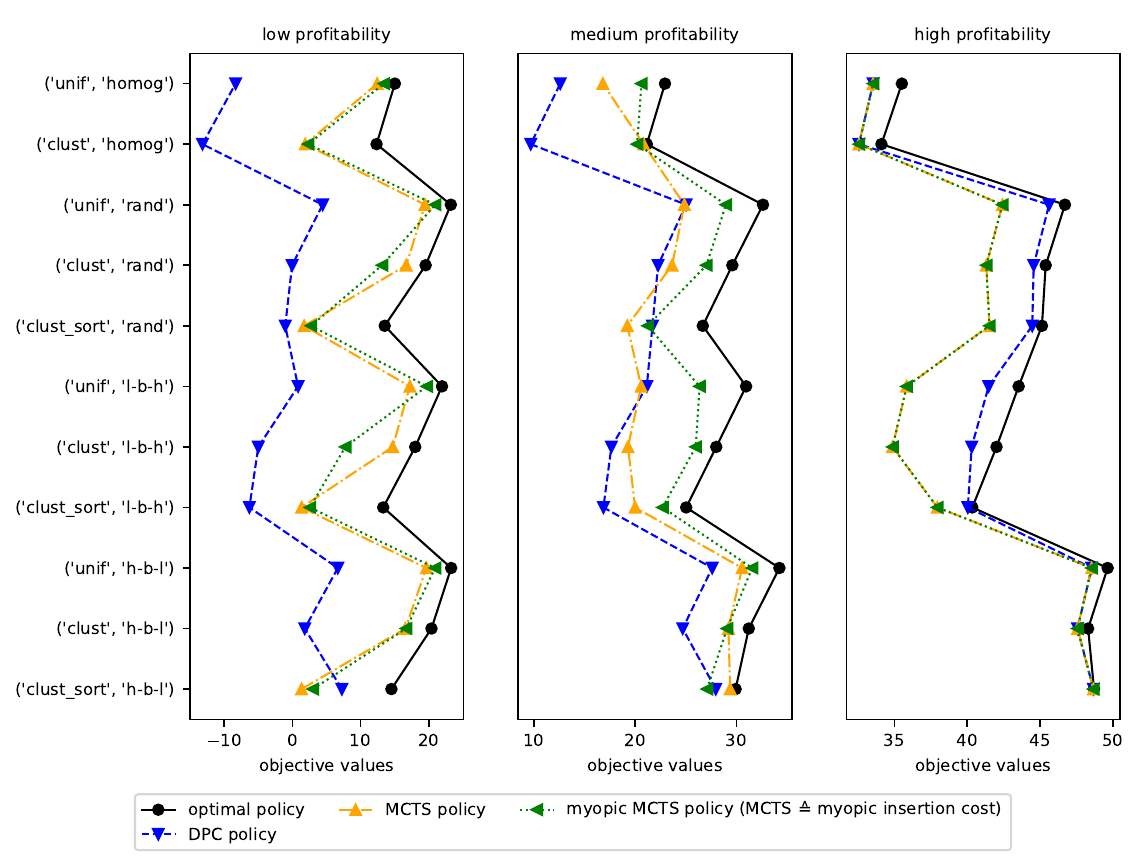}
     \caption[]{Load-constrained settings}
	\end{subfigure}
    \vskip\baselineskip

    \begin{subfigure}{0.75\textwidth}
    \renewcommand\thesubfigure{b}
			\centering
    		\includegraphics[width=\textwidth]{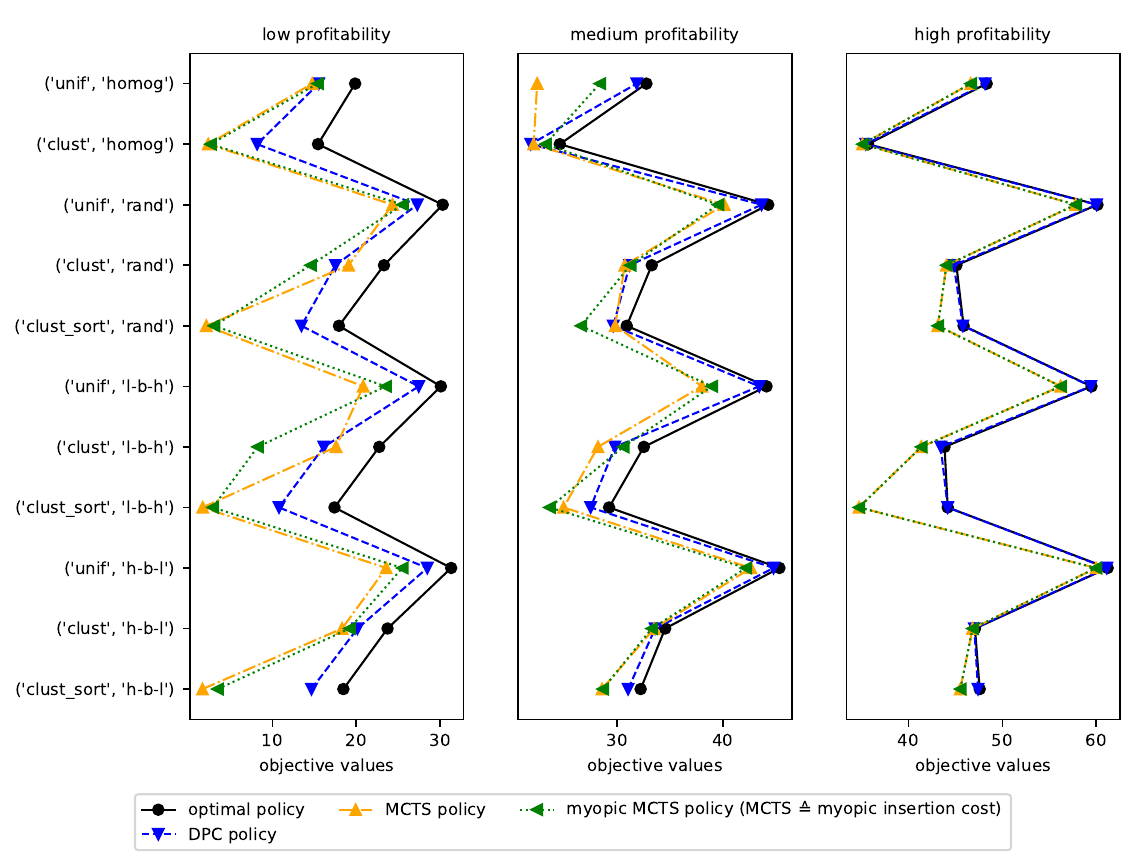}
     \caption[]{Distance-constrained settings}
     \end{subfigure}
	\label{fig: profilLines}
\end{figure}
\section{Heatmaps}\label{app: heatmaps}
\vspace{-1cm}
\begin{figure}[H]
\centering
\footnotesize
\caption{DPC policy - Underestimation errors}
\label{fig: HM0}
    \vskip\baselineskip
	\centering
	\captionsetup{justification=centering}
    \begin{subfigure}{0.6\textwidth}
    \renewcommand\thesubfigure{a}
        \begin{subfigure}{\textwidth}
    		\renewcommand\thesubfigure{a}
            \begin{subfigure}[]{0.315\textwidth}
    			\centering
    			\includegraphics[width=\textwidth]{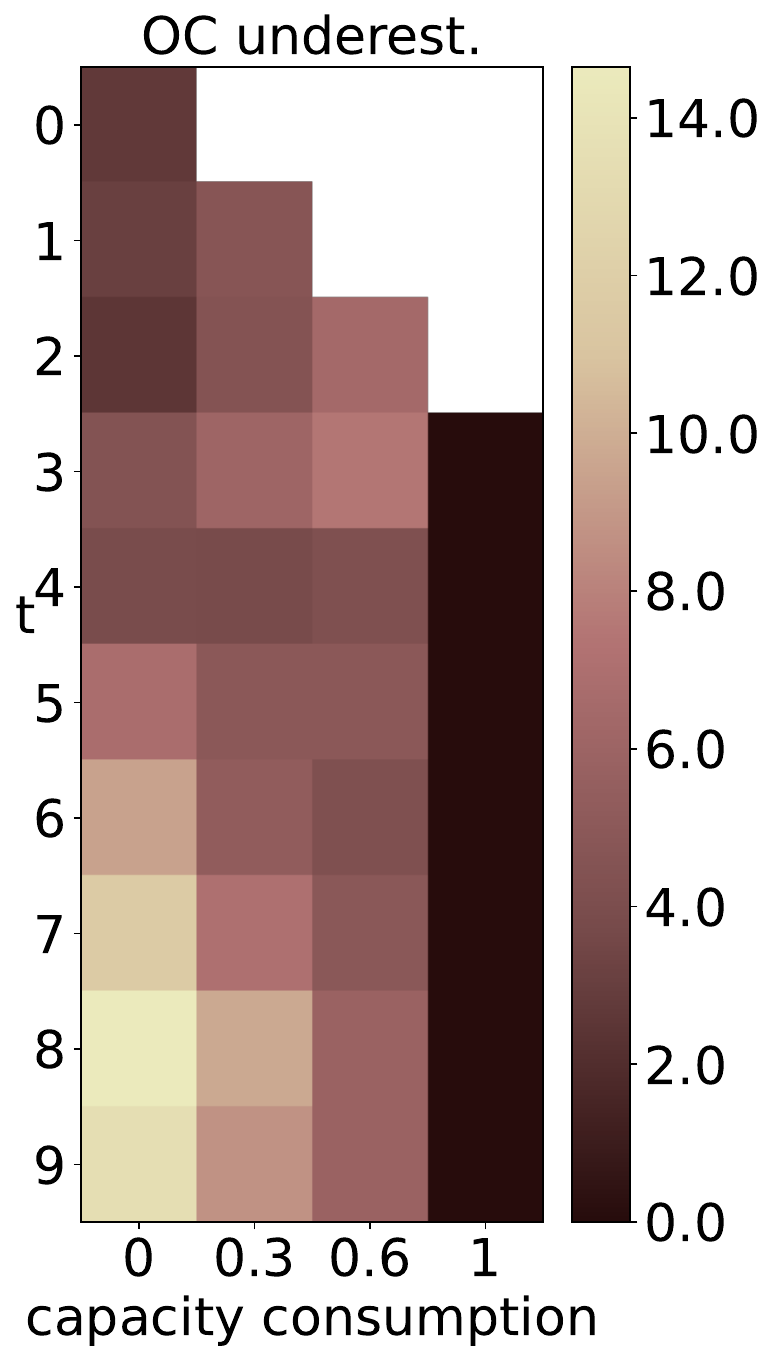}
    		\end{subfigure}
    		\begin{subfigure}[]{0.3\textwidth}
    			\centering
    			\includegraphics[width=\textwidth]{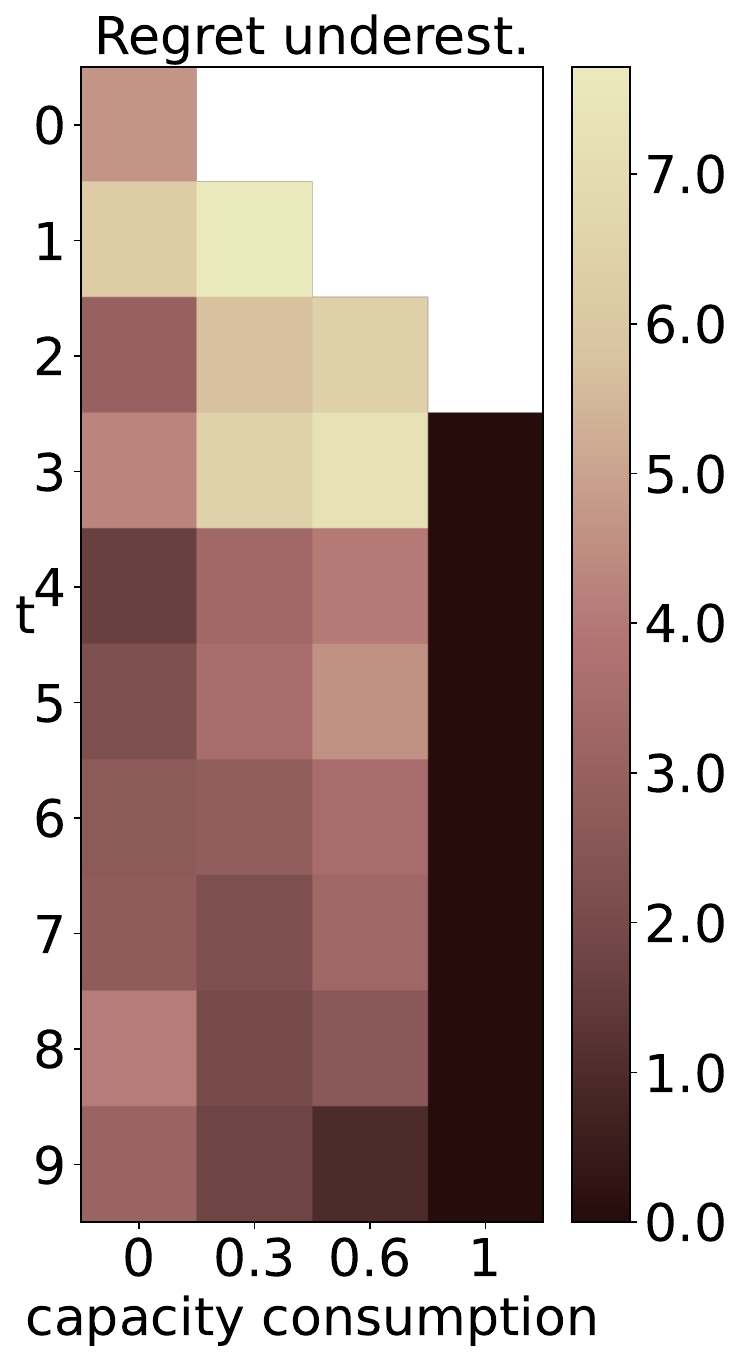}
    		\end{subfigure}
            \begin{subfigure}[]{0.3\textwidth}
    			\centering
    			\includegraphics[width=\textwidth]{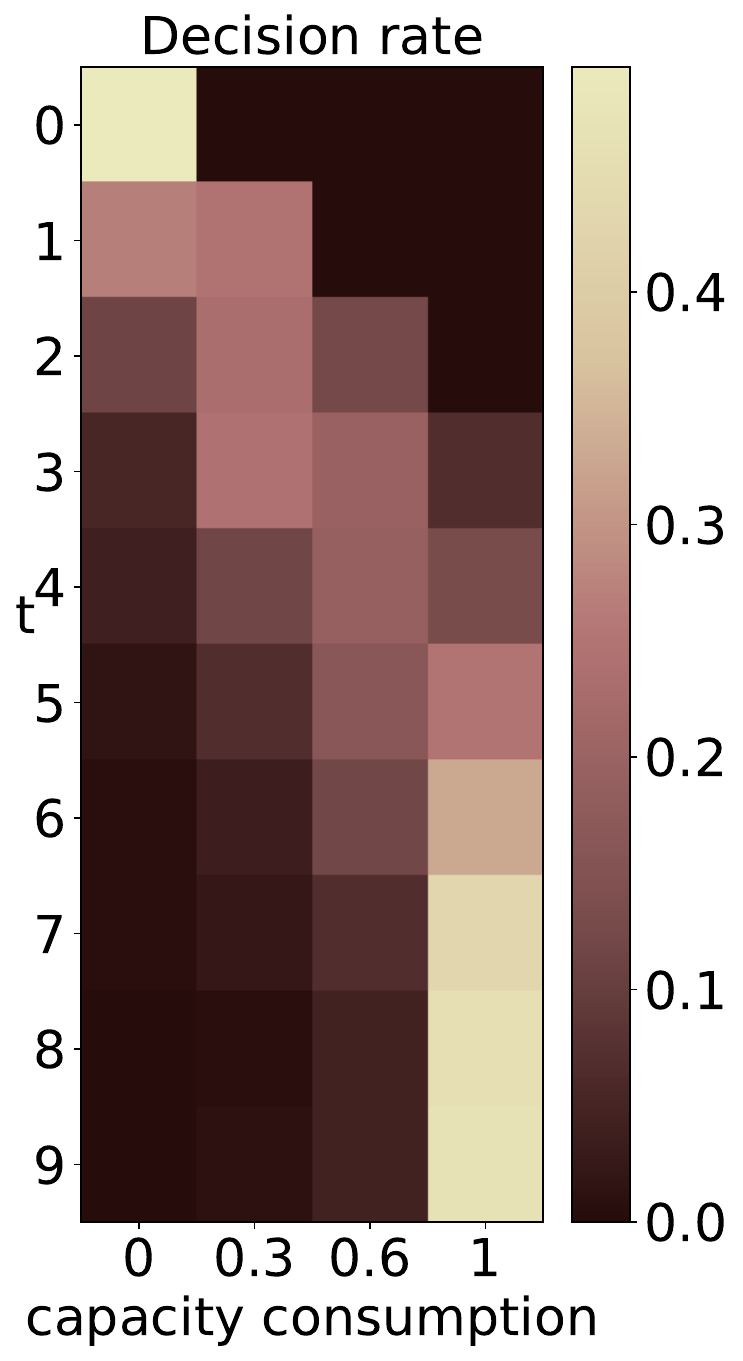}
    		\end{subfigure}
        \caption{med $\mid$ load $\mid$ unif $\mid$ homog}%
        \label{fig: HM0a}
    \end{subfigure}
        \vskip\baselineskip

        \begin{subfigure}{\textwidth}
		\renewcommand\thesubfigure{b}
      	\begin{subfigure}[]{0.315\textwidth}
    			\centering
    			\includegraphics[width=\textwidth]{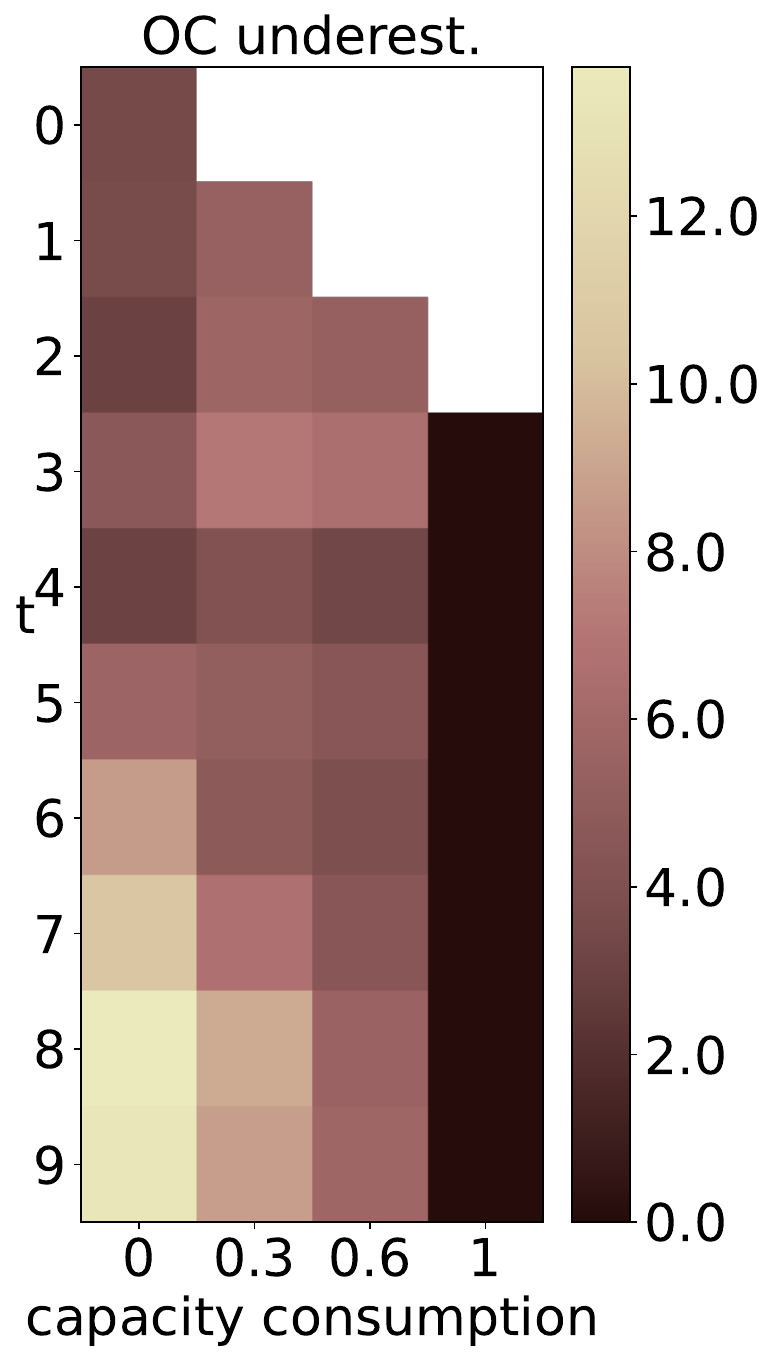}
    		\end{subfigure}
    		\begin{subfigure}[]{0.3\textwidth}
    			\centering
    			\includegraphics[width=\textwidth]{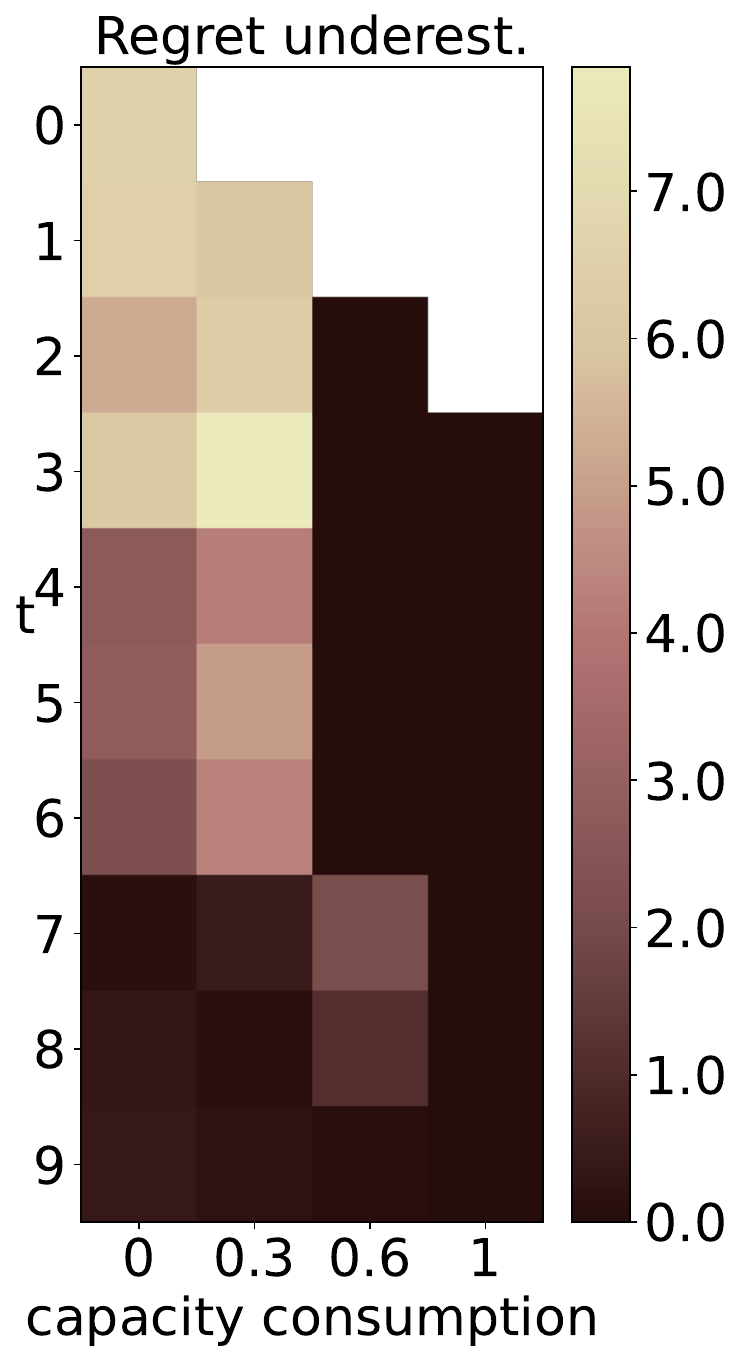}
    		\end{subfigure}
            \begin{subfigure}[]{0.3\textwidth}
    			\centering
    			\includegraphics[width=\textwidth]{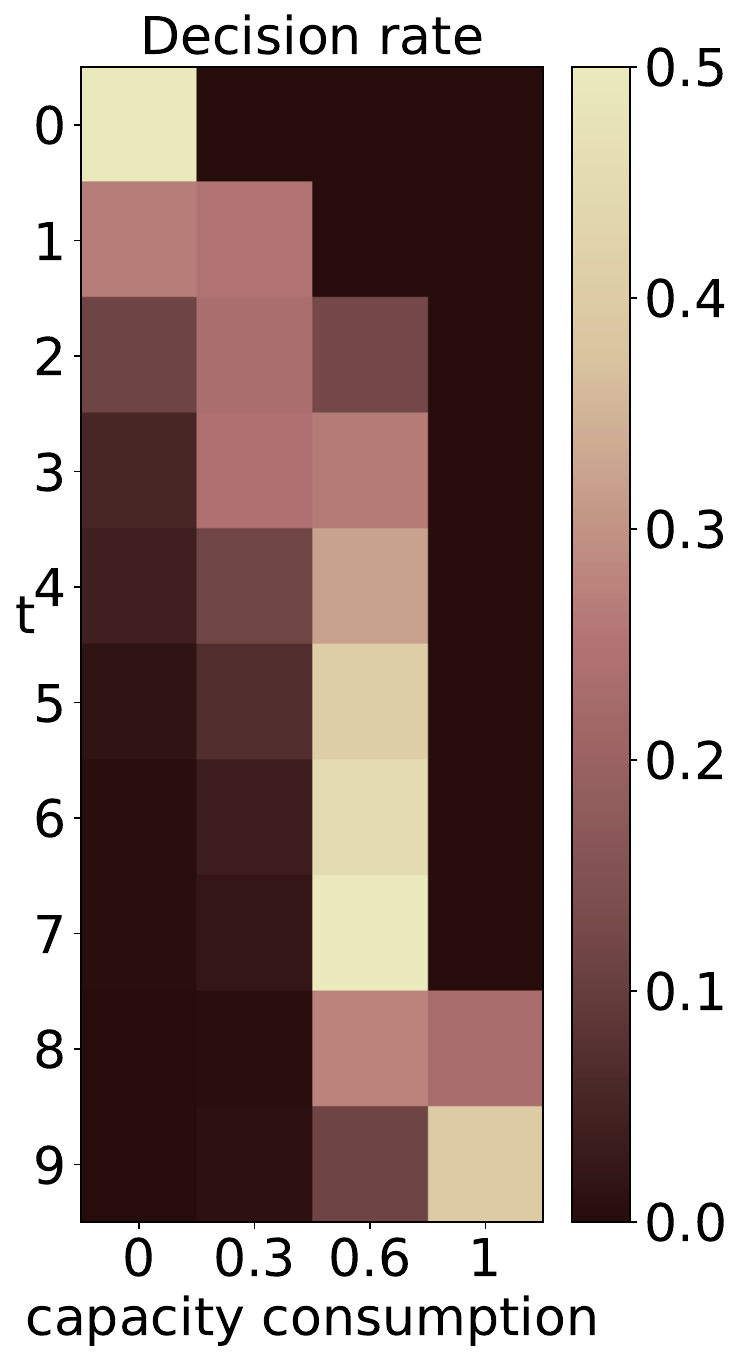}
    		\end{subfigure}
        \caption{med $\mid$ load $\mid$ unif $\mid$ l-b-h }
        \label{fig: HM0b}
        \end{subfigure}
	\end{subfigure}
        \vskip\baselineskip

    \begin{subfigure}{0.8\textwidth}
    \renewcommand\thesubfigure{b}
        \begin{subfigure}{\textwidth}
    		\renewcommand\thesubfigure{c}
            \begin{subfigure}[]{0.3\textwidth}
    			\centering
    			\includegraphics[width=\textwidth]{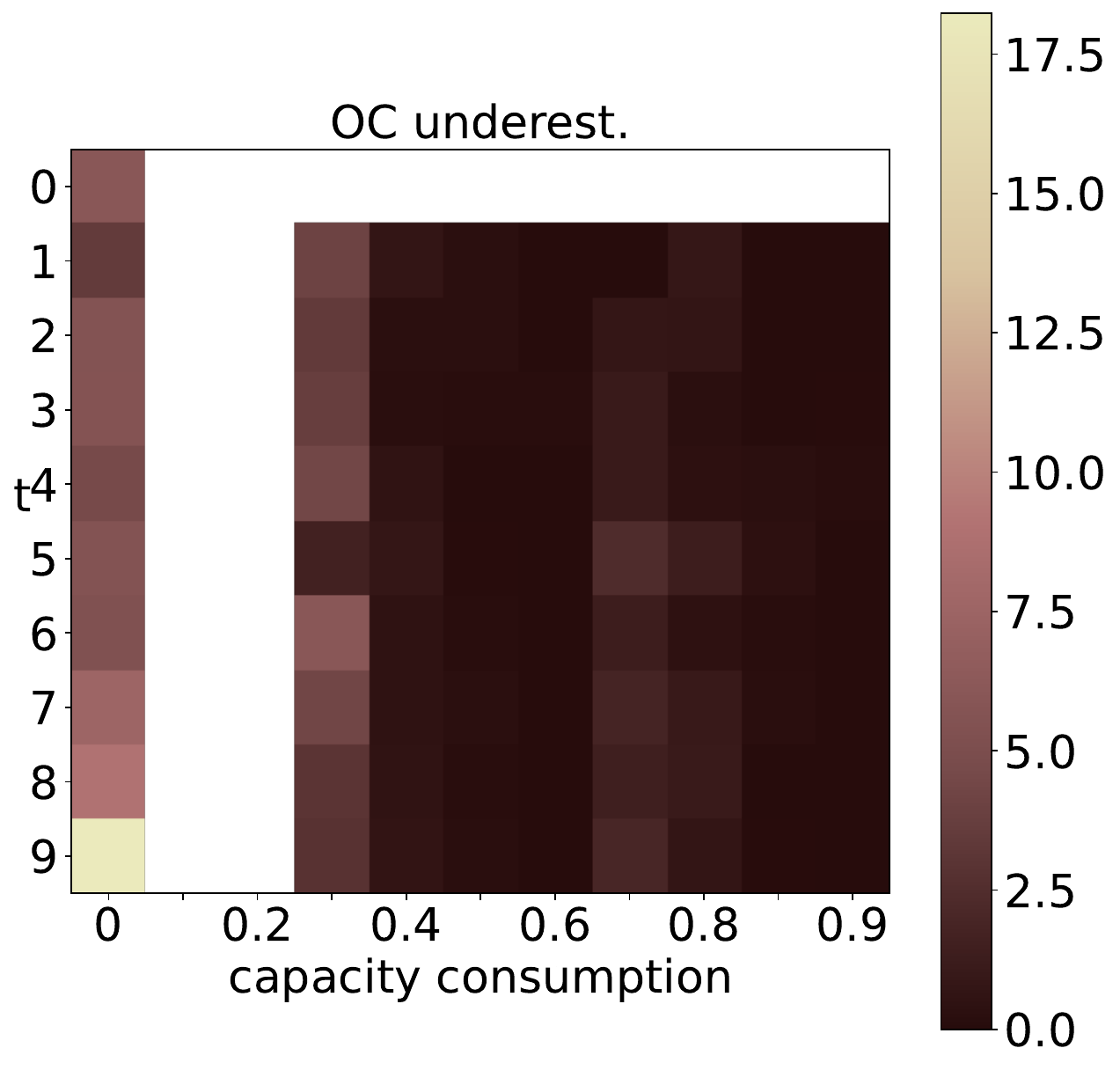}
    		\end{subfigure}
    		\begin{subfigure}[]{0.3\textwidth}
    			\centering
    			\includegraphics[width=\textwidth]{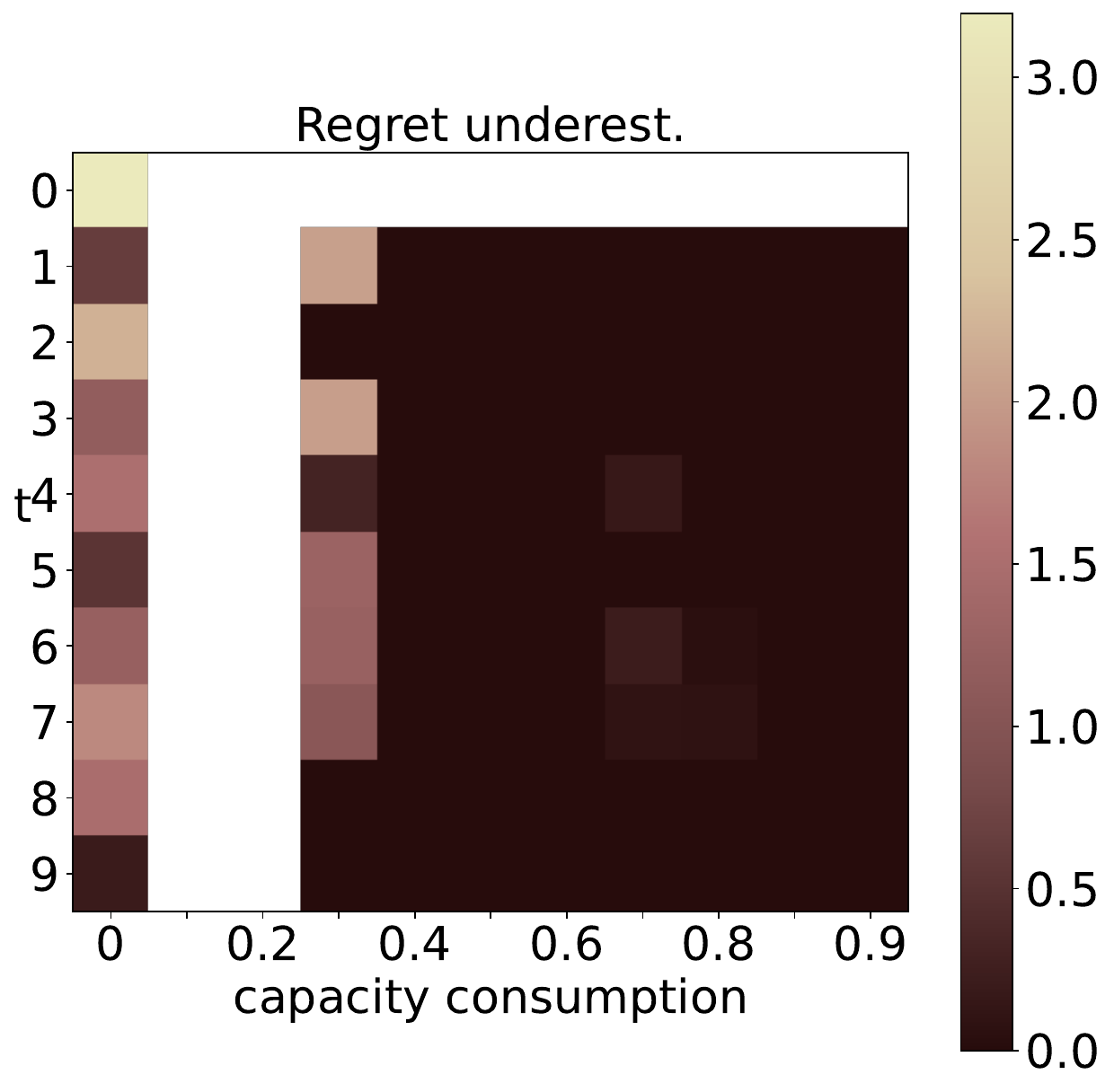}
    		\end{subfigure}
          \begin{subfigure}[]{0.3\textwidth}
    			\centering
    			\includegraphics[width=\textwidth]{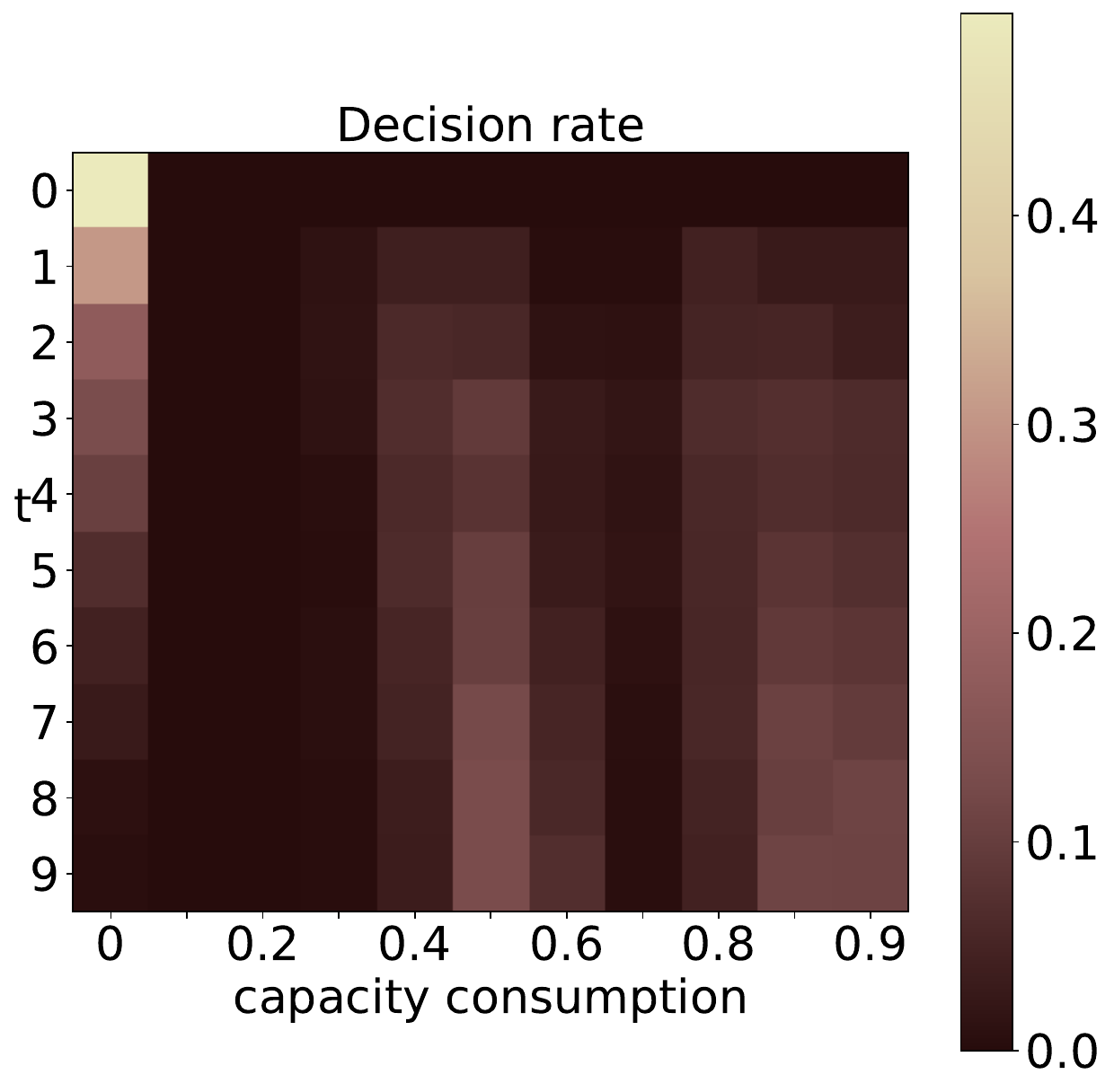}
    		\end{subfigure}
        \caption{med $\mid$ dist $\mid$ clust $\mid$ l-b-h }
        \label{fig: HM0c}
        \end{subfigure}
        \end{subfigure}
    \vskip\baselineskip

        \centering
        \footnotesize
        \begin{subfigure}{0.6\textwidth} 
		\renewcommand\thesubfigure{d}
      	\begin{subfigure}[]{0.3\textwidth}
    			\centering
    			\includegraphics[width=\textwidth]{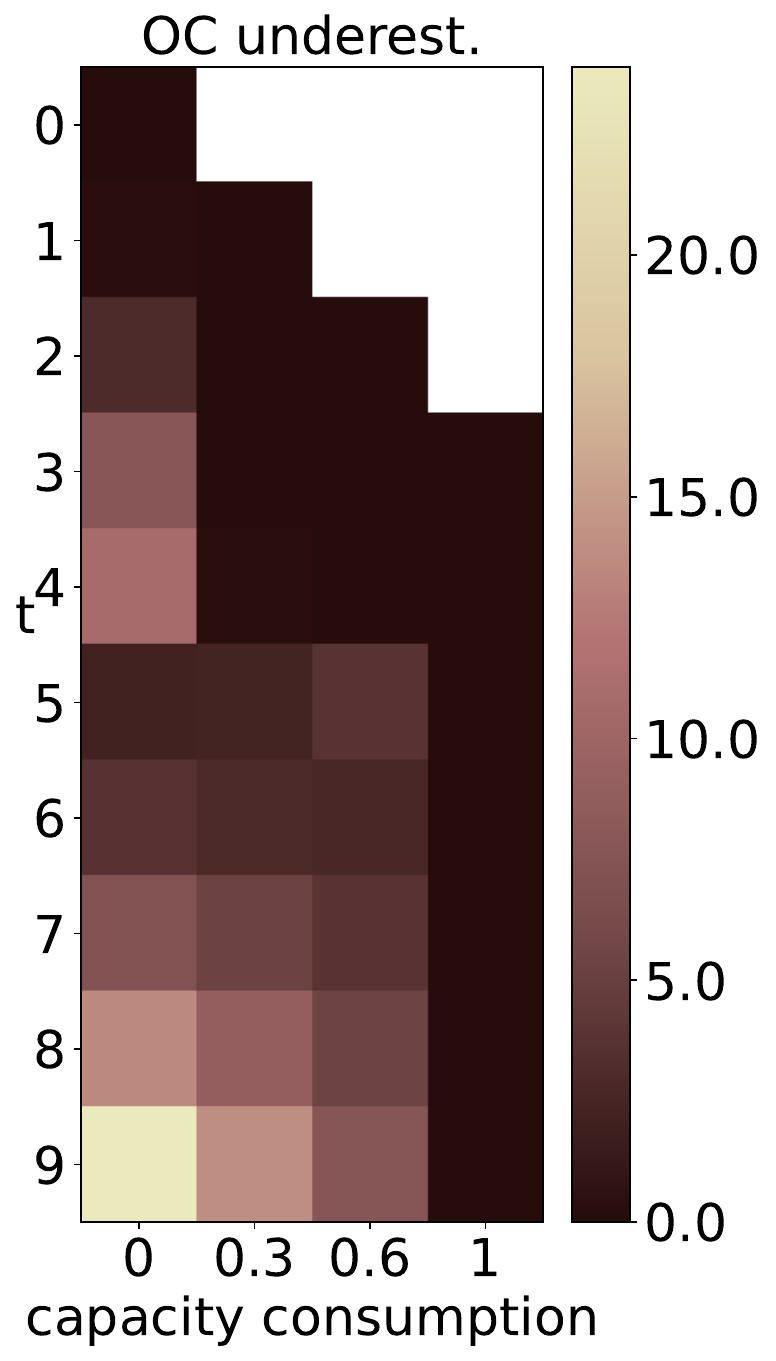}
    		\end{subfigure}
    		\begin{subfigure}[]{0.3\textwidth}
    			\centering
    			\includegraphics[width=\textwidth]{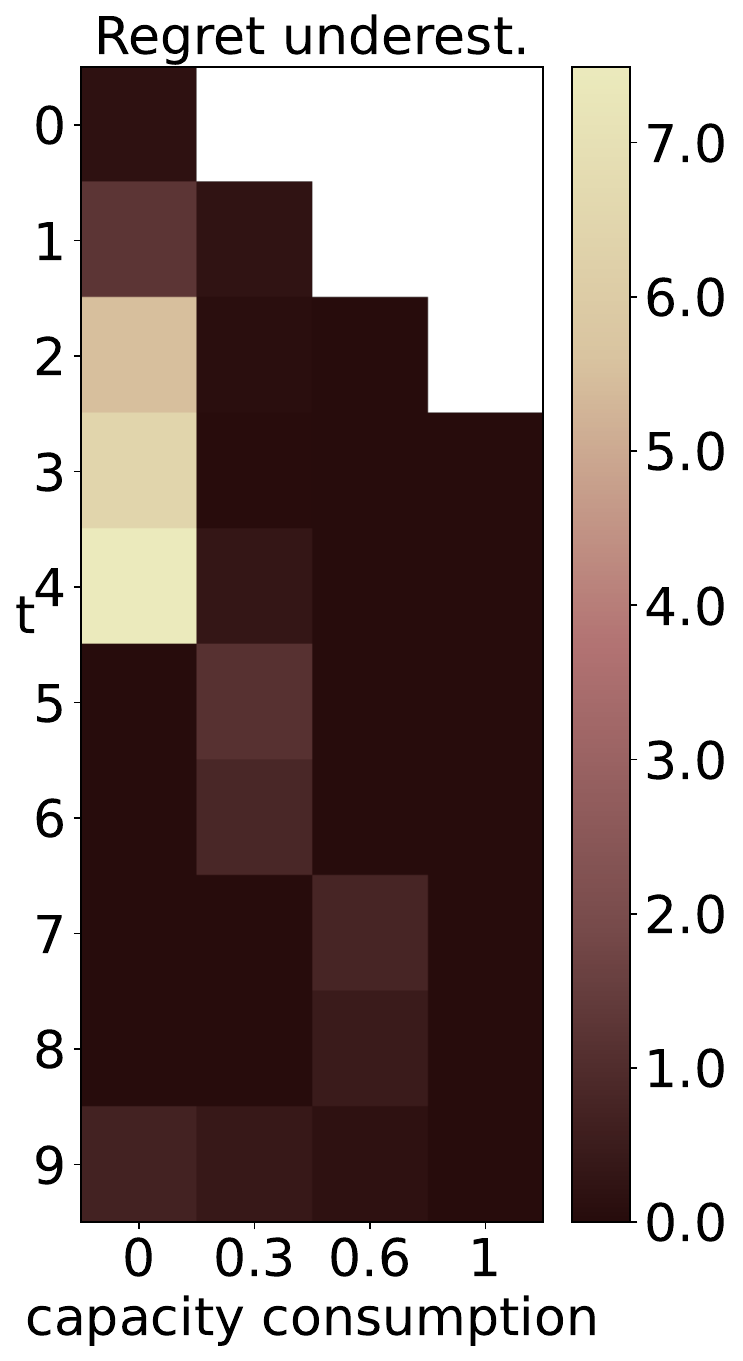}
    		\end{subfigure}
            \begin{subfigure}[]{0.3\textwidth}
    			\centering
    			\includegraphics[width=\textwidth]{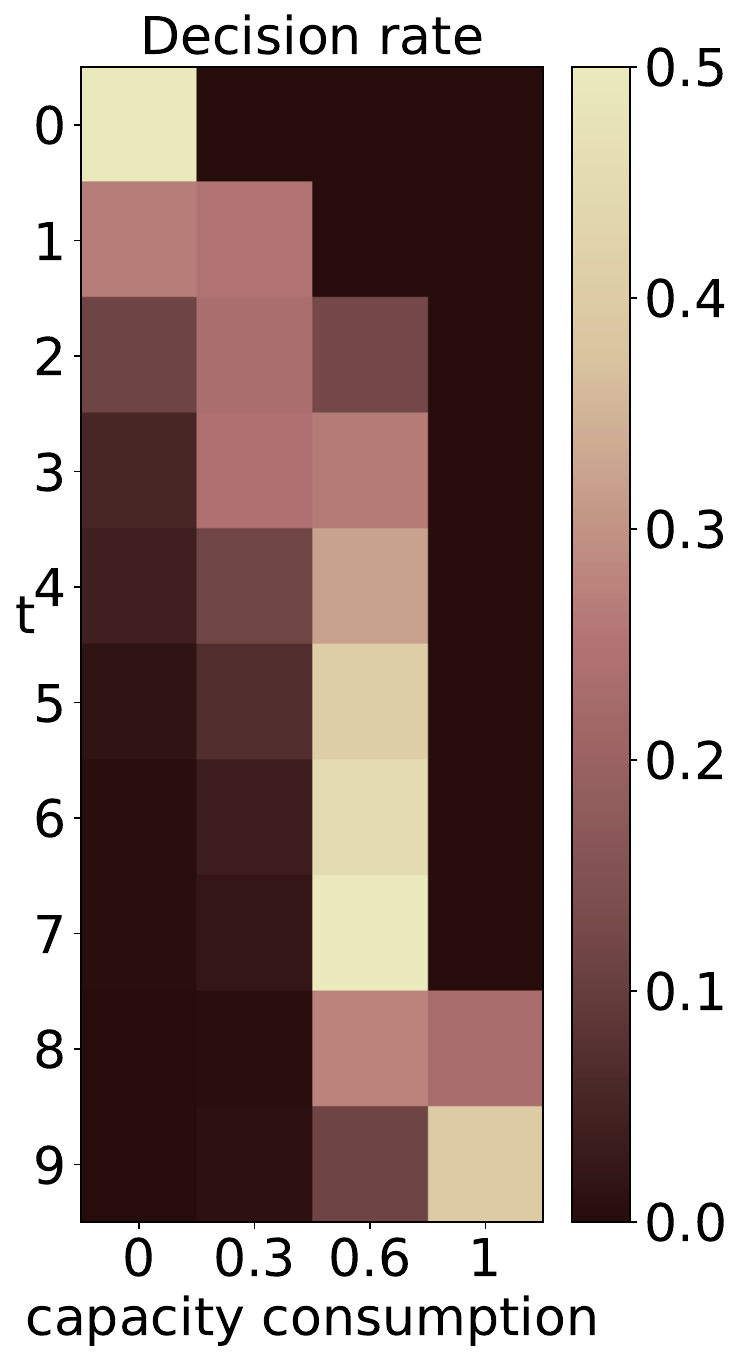}
    		\end{subfigure}
        \caption{med $\mid$ load $\mid$ clust\_sort $\mid$ l-b-h}
        \label{fig: HM0d}
     \end{subfigure}
	\end{figure}
 \vspace{-1cm}
\begin{figure}[H]
\centering
\footnotesize
\caption{DPC policy - overestimation error}
\label{fig: HM1}
\vskip\baselineskip
\centering
	\captionsetup{justification=centering}
	\begin{subfigure}{0.6\textwidth}
        \begin{subfigure}{\textwidth}
    		\renewcommand\thesubfigure{a}
            \begin{subfigure}[]{0.3\textwidth}
    			\centering
    			\includegraphics[width=\textwidth]{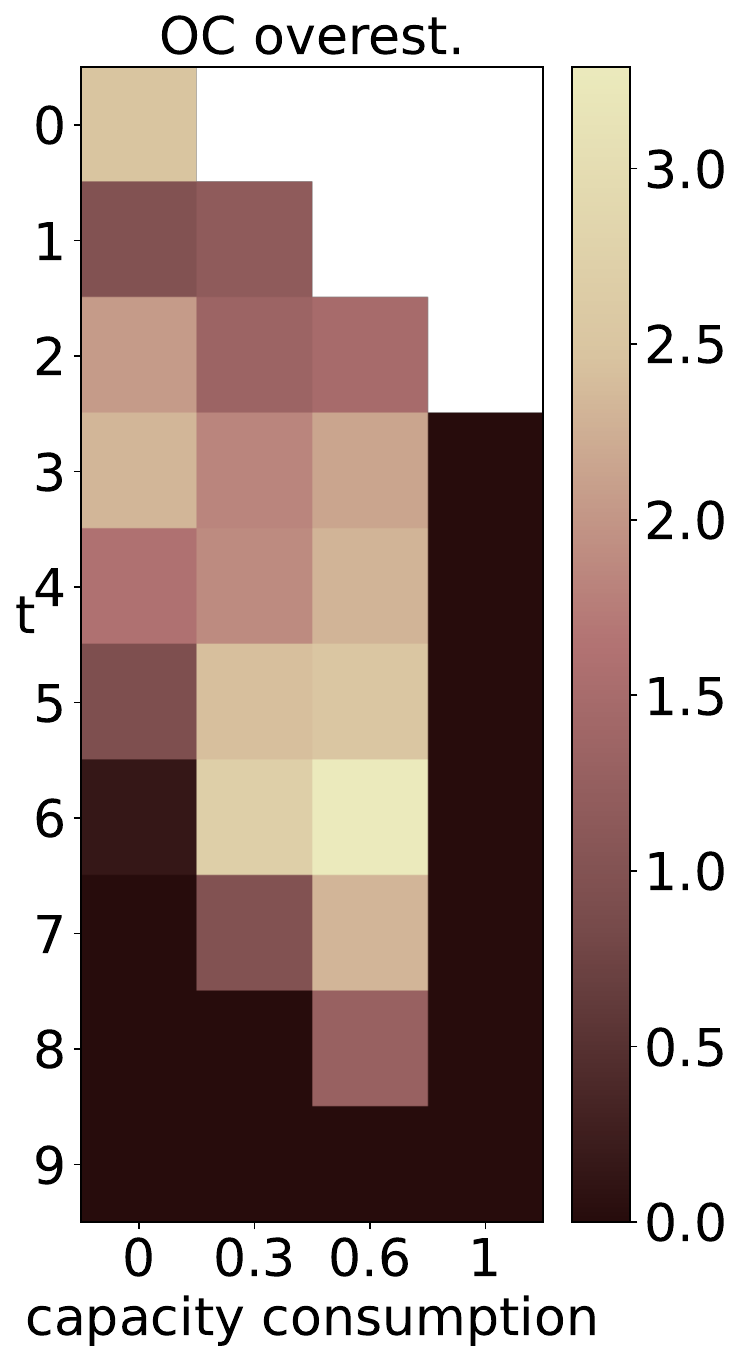}
    		\end{subfigure}
    		\begin{subfigure}[]{0.3\textwidth}
    			\centering
    			\includegraphics[width=\textwidth]{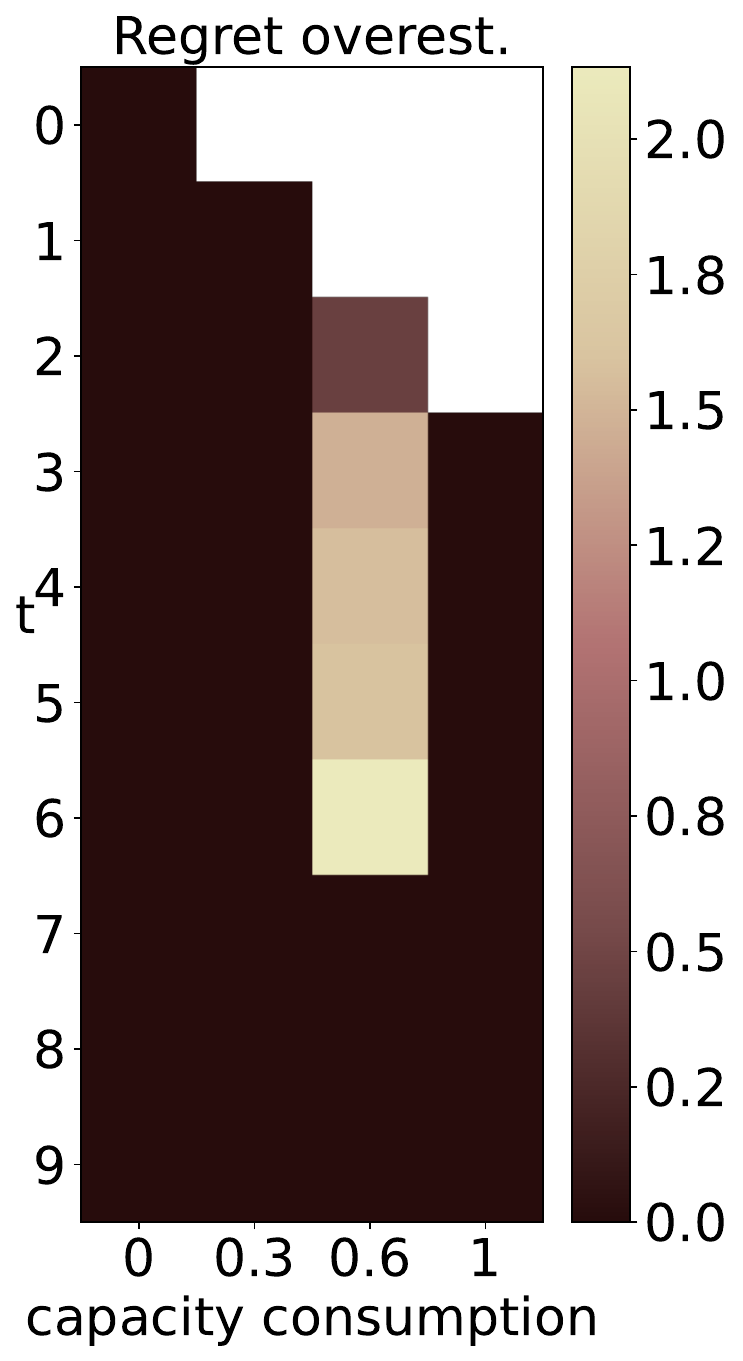}
    		\end{subfigure}
            \begin{subfigure}[]{0.3\textwidth}
    			\centering
    			\includegraphics[width=\textwidth]{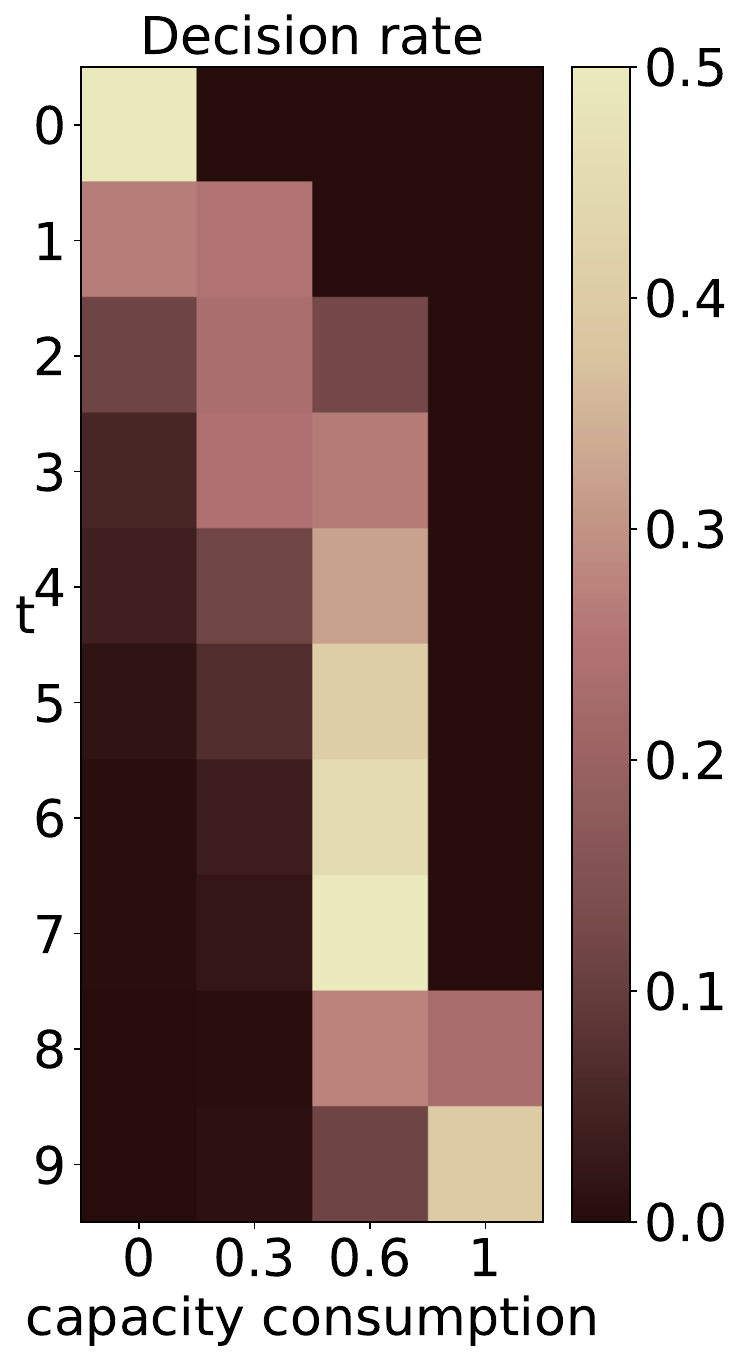}
    		\end{subfigure}
        \caption{low $\mid$ load $\mid$ unif $\mid$ l-b-h}%
        \label{fig: HM1a}
        \end{subfigure}
            \vskip\baselineskip
        \begin{subfigure}{\textwidth}
		\renewcommand\thesubfigure{b}
      	\begin{subfigure}[]{0.3\textwidth}
    			\centering
    			\includegraphics[width=\textwidth]{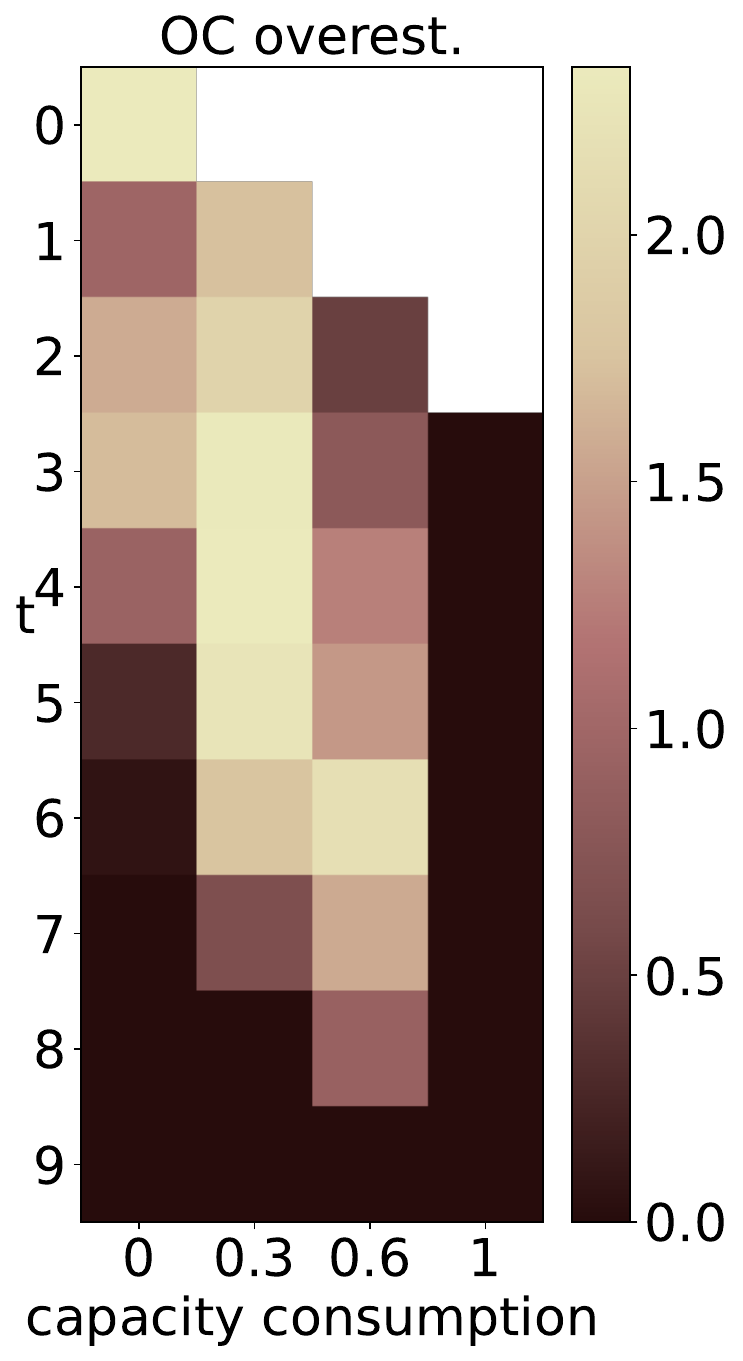}
    		\end{subfigure}
    		\begin{subfigure}[]{0.3\textwidth}
    			\centering
    			\includegraphics[width=\textwidth]{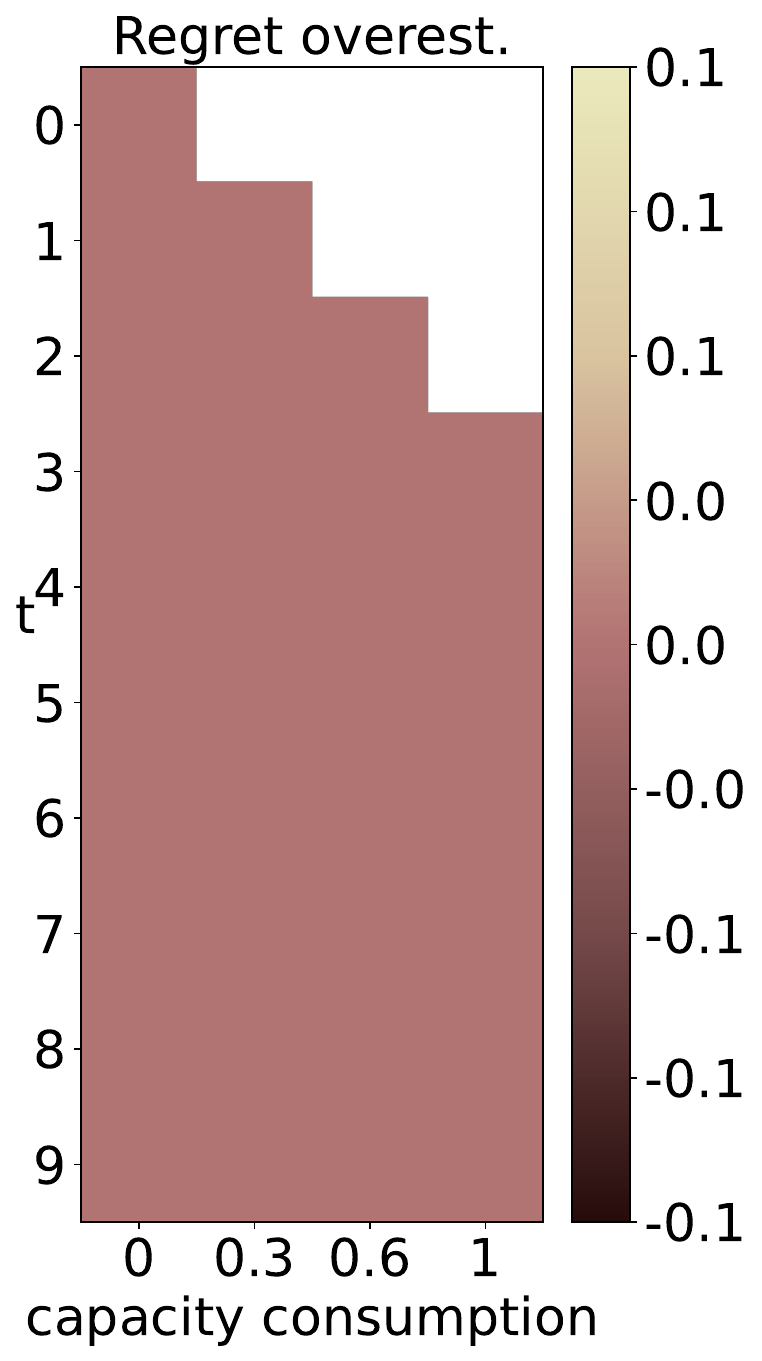}
    		\end{subfigure}
           \begin{subfigure}[]{0.3\textwidth}
    			\centering
    			\includegraphics[width=\textwidth]{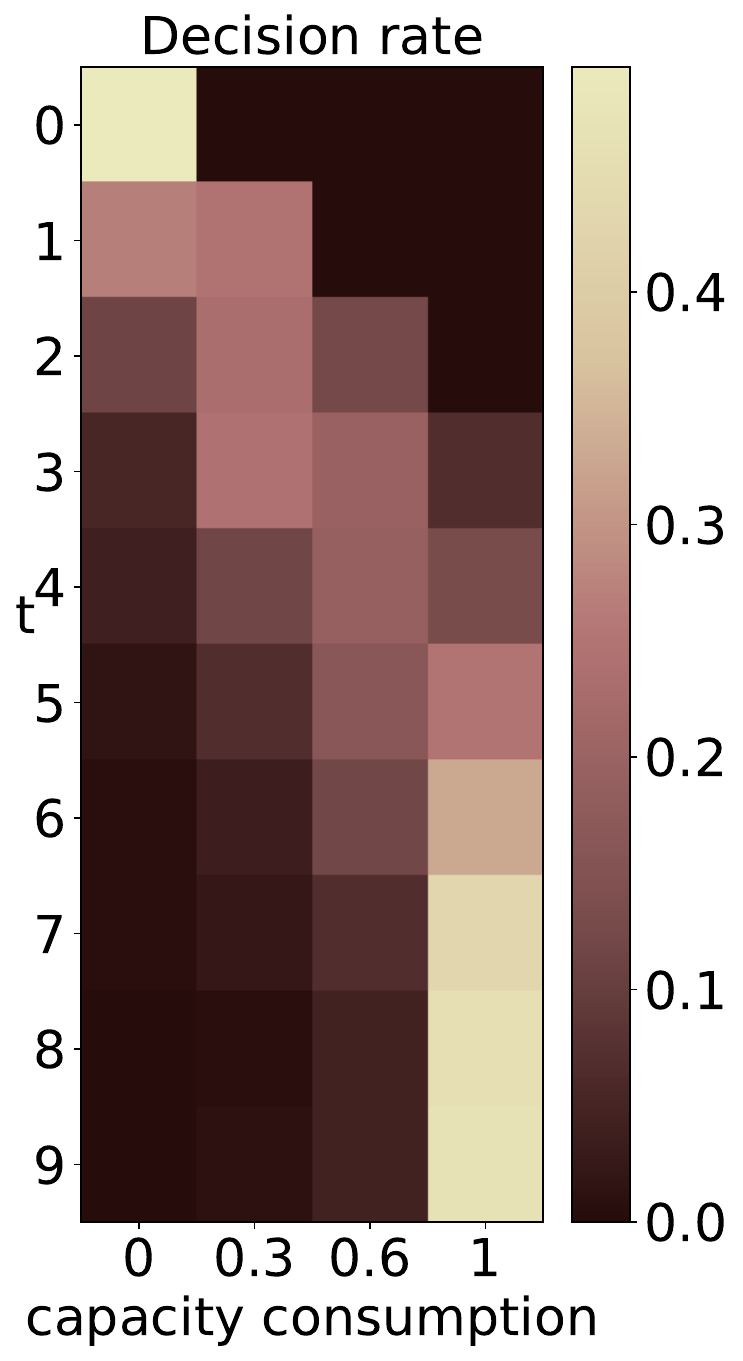}
    		\end{subfigure}
        \caption{low $\mid$ load $\mid$ unif $\mid$ h-b-l}%
        \label{fig: HM1b}
        \end{subfigure}
	\end{subfigure}
    \vskip\baselineskip

        \begin{subfigure}{0.8\textwidth}
    		\renewcommand\thesubfigure{c}
            \begin{subfigure}[]{0.3\textwidth}
    			\centering
    			\includegraphics[width=\textwidth]{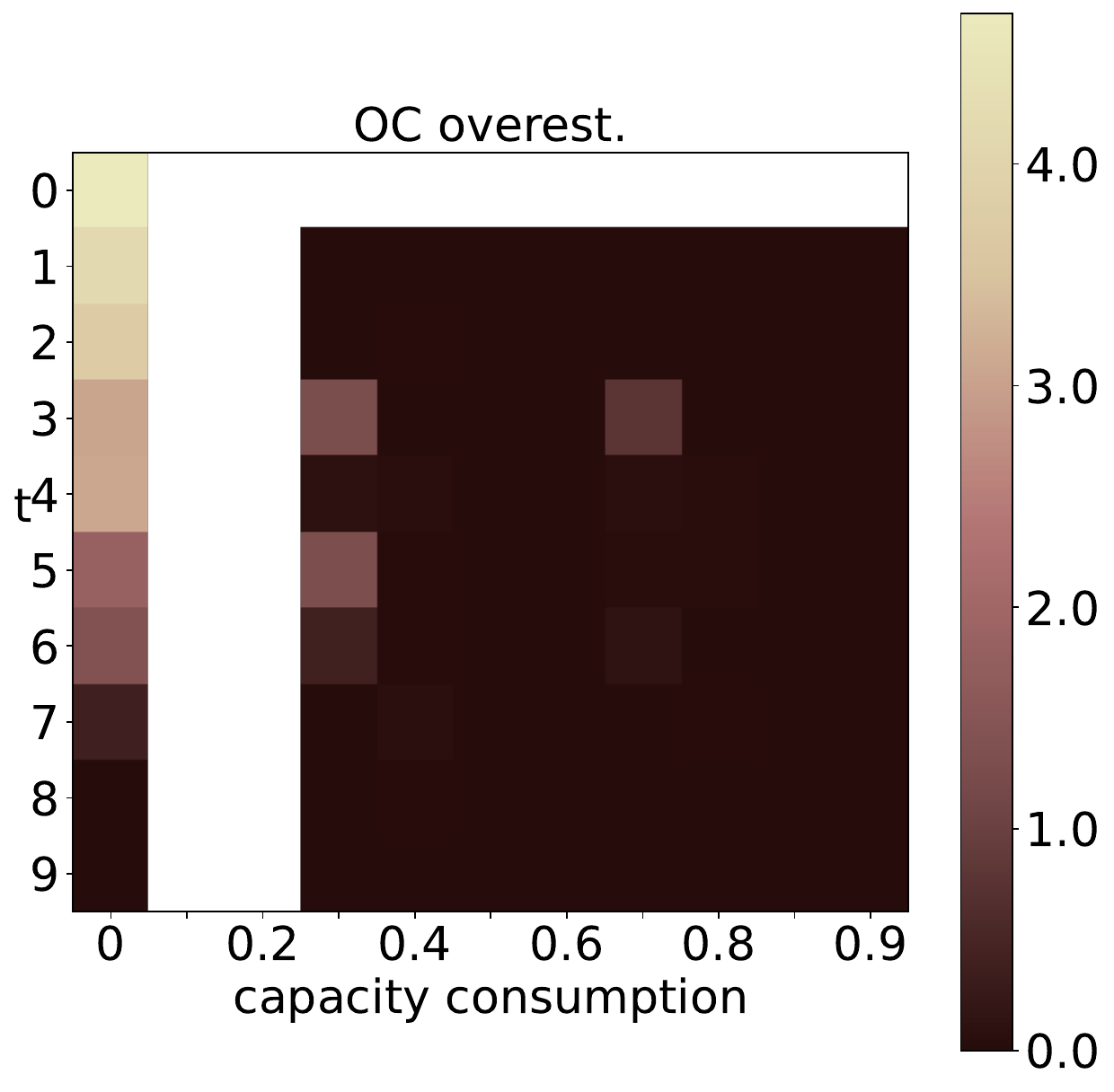}
    		\end{subfigure}
    		\begin{subfigure}[]{0.3\textwidth}
    			\centering
    			\includegraphics[width=\textwidth]{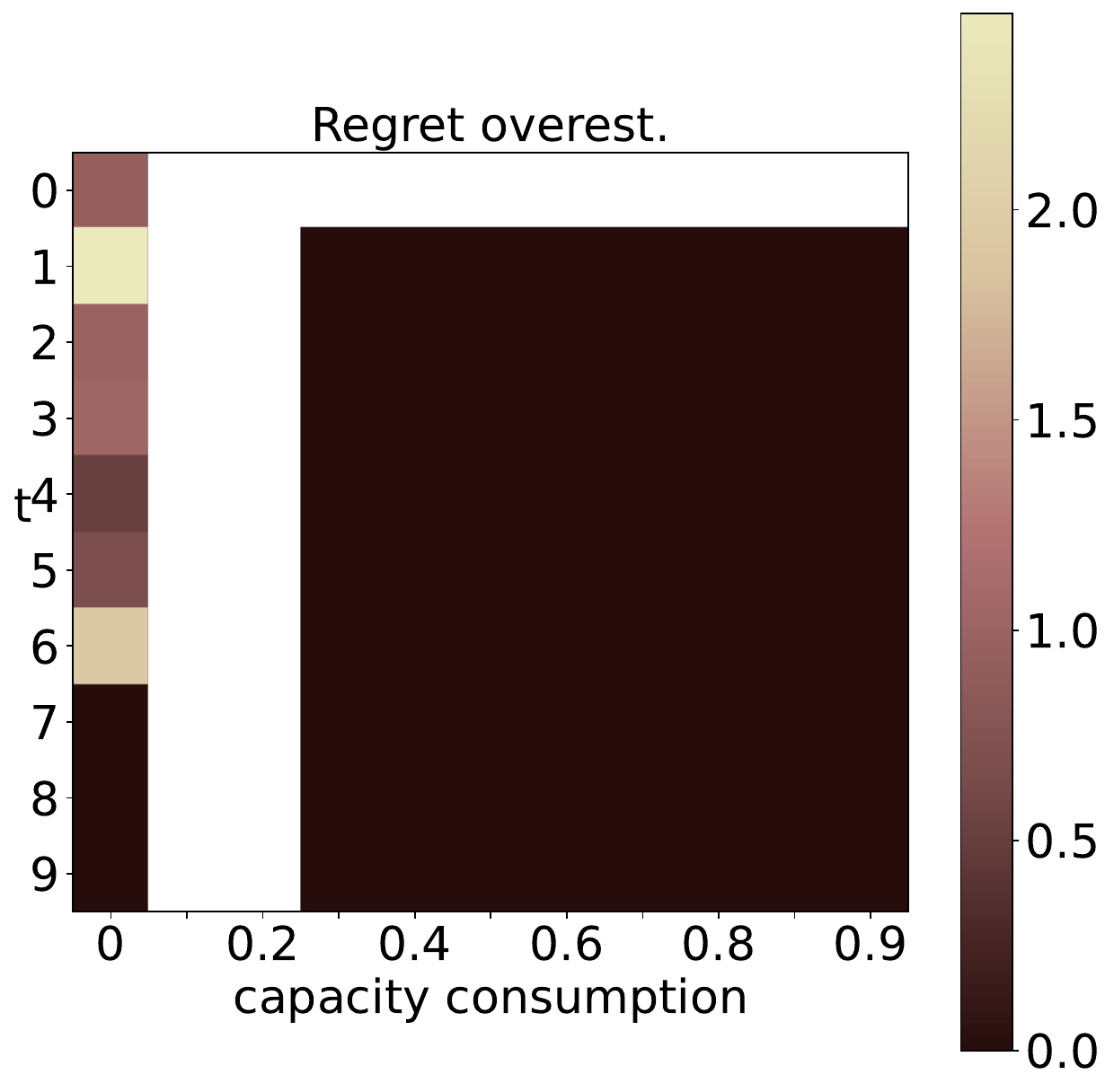}
    		\end{subfigure}
            \begin{subfigure}[]{0.3\textwidth}
    			\centering
    			\includegraphics[width=\textwidth]{Figures/1c0}
    		\end{subfigure}
        \caption{low $\mid$ distance $\mid$ clust $\mid$ l-b-h}
        \label{fig: HM1c}
        \end{subfigure}
    \vskip\baselineskip
        \begin{subfigure}{0.6\textwidth}
        \centering
		\renewcommand\thesubfigure{d}
      	\begin{subfigure}[]{0.3\textwidth}
    			\centering
    			\includegraphics[width=\textwidth]{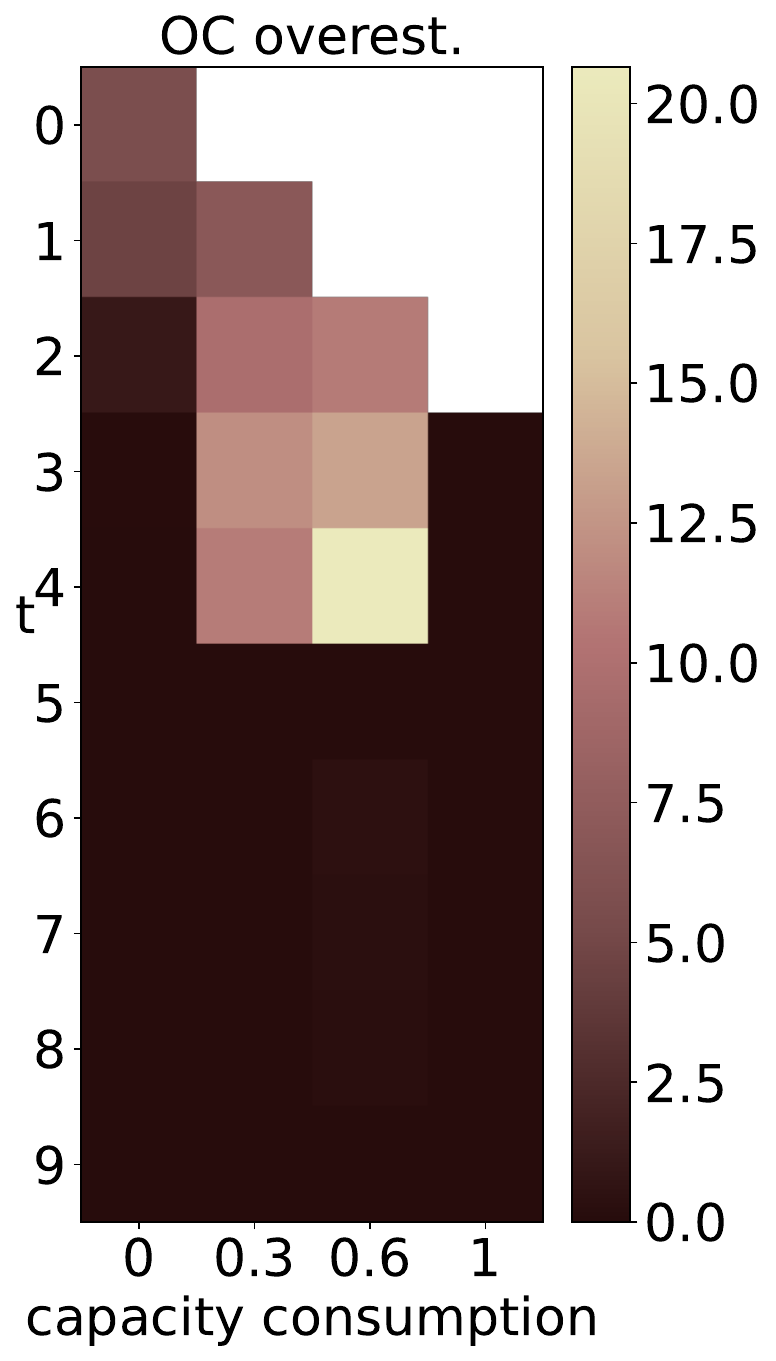}
    		\end{subfigure}
    		\begin{subfigure}[]{0.3\textwidth}
    			\centering
    			\includegraphics[width=\textwidth]{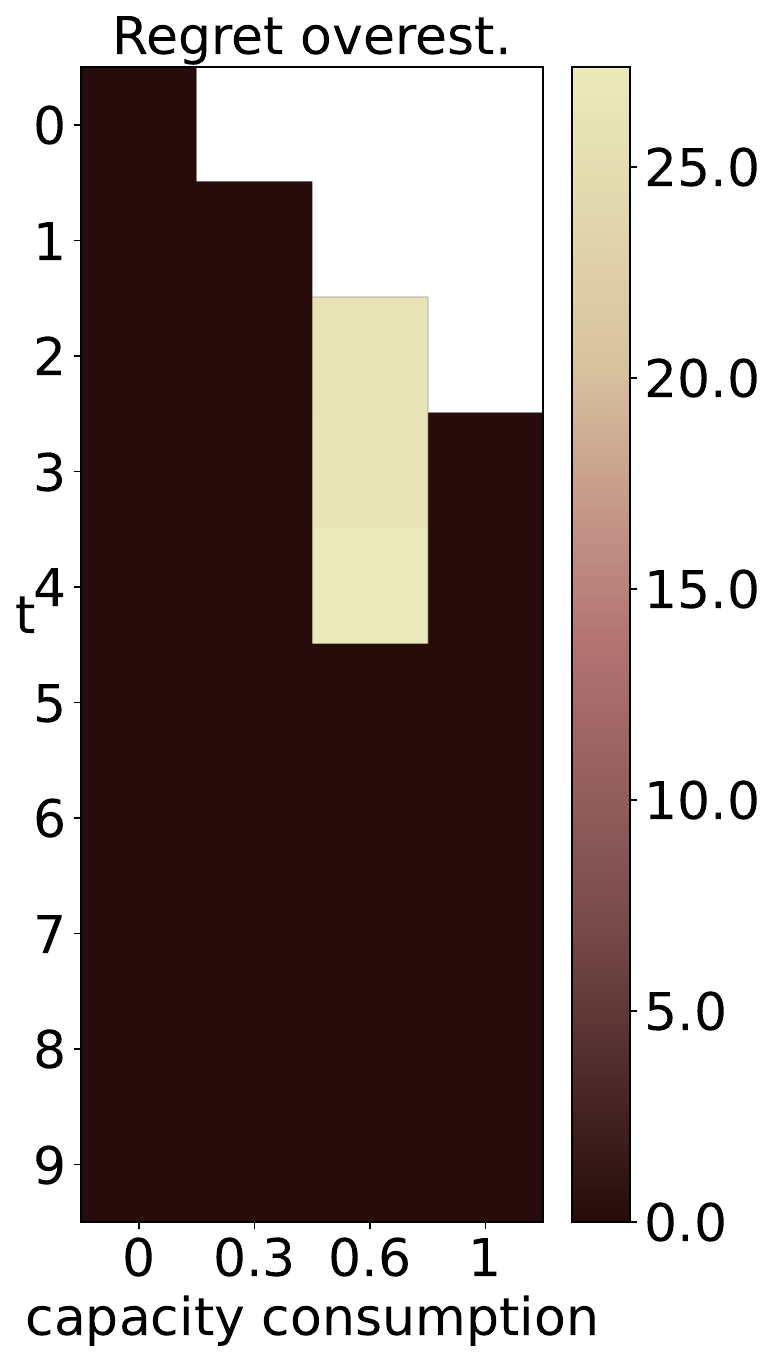}
    		\end{subfigure}
            \begin{subfigure}[]{0.29\textwidth}
    			\centering
    			\includegraphics[width=\textwidth]{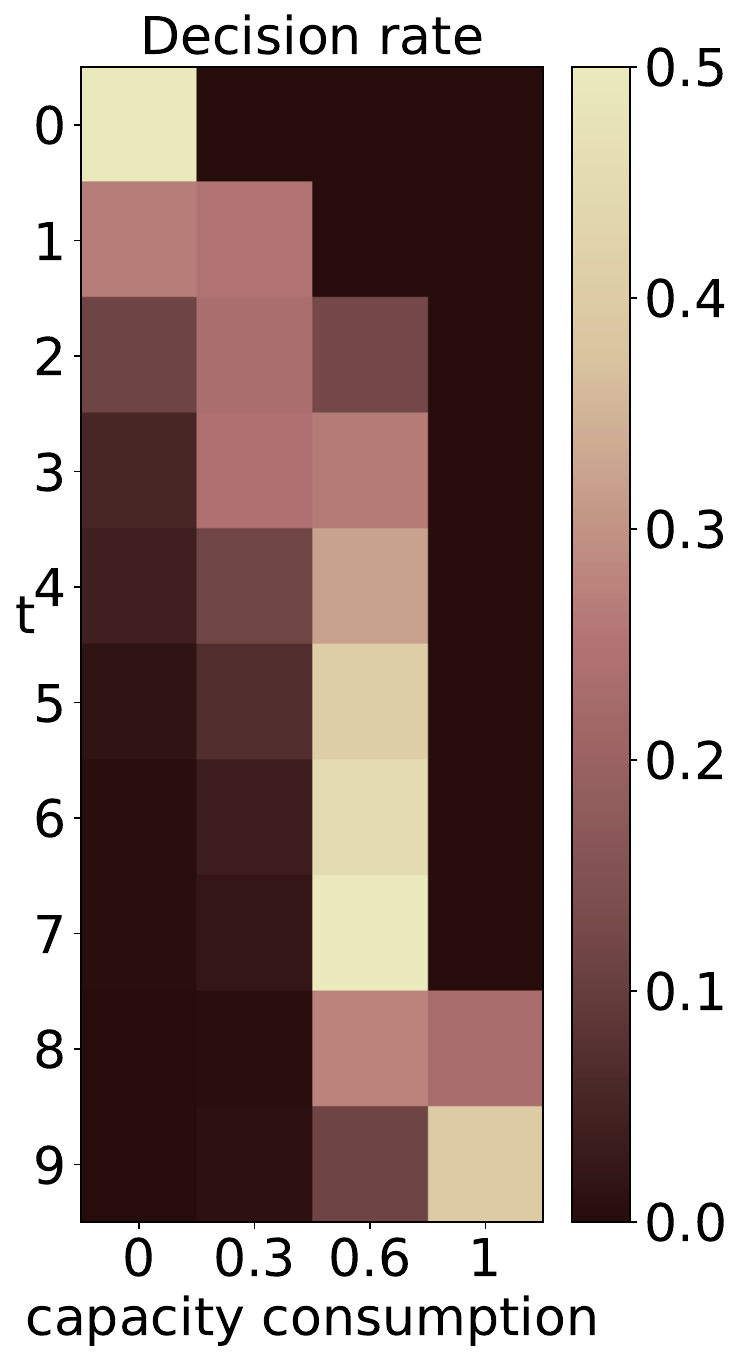}
    		\end{subfigure}
        \caption{low $\mid$ load $\mid$ clust\_sort $\mid$ l-b-h}
        \label{fig: HM1d}
	\end{subfigure}
\end{figure}
\pagebreak

\begin{figure}[H]
\centering
\footnotesize
\caption{MCTS policy - underestimation errors}\label{fig: HM2}
    \vskip\baselineskip
	\centering
	\captionsetup{justification=centering}
        \begin{subfigure}{0.58\textwidth}
    		\renewcommand\thesubfigure{a}
            \begin{subfigure}[]{0.315\textwidth}
    			\centering
    			\includegraphics[width=\textwidth]{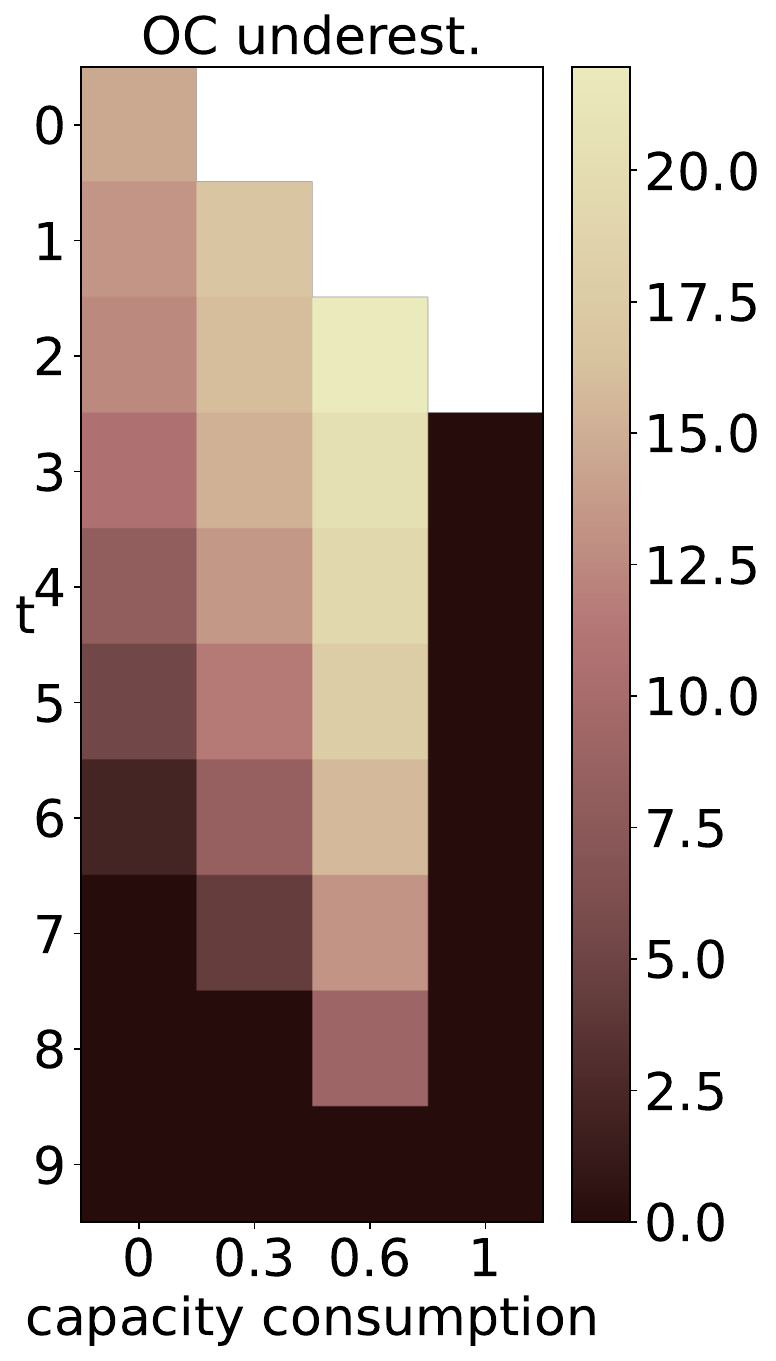}
    		\end{subfigure}
    		\begin{subfigure}[]{0.3\textwidth}
    			\centering
    			\includegraphics[width=\textwidth]{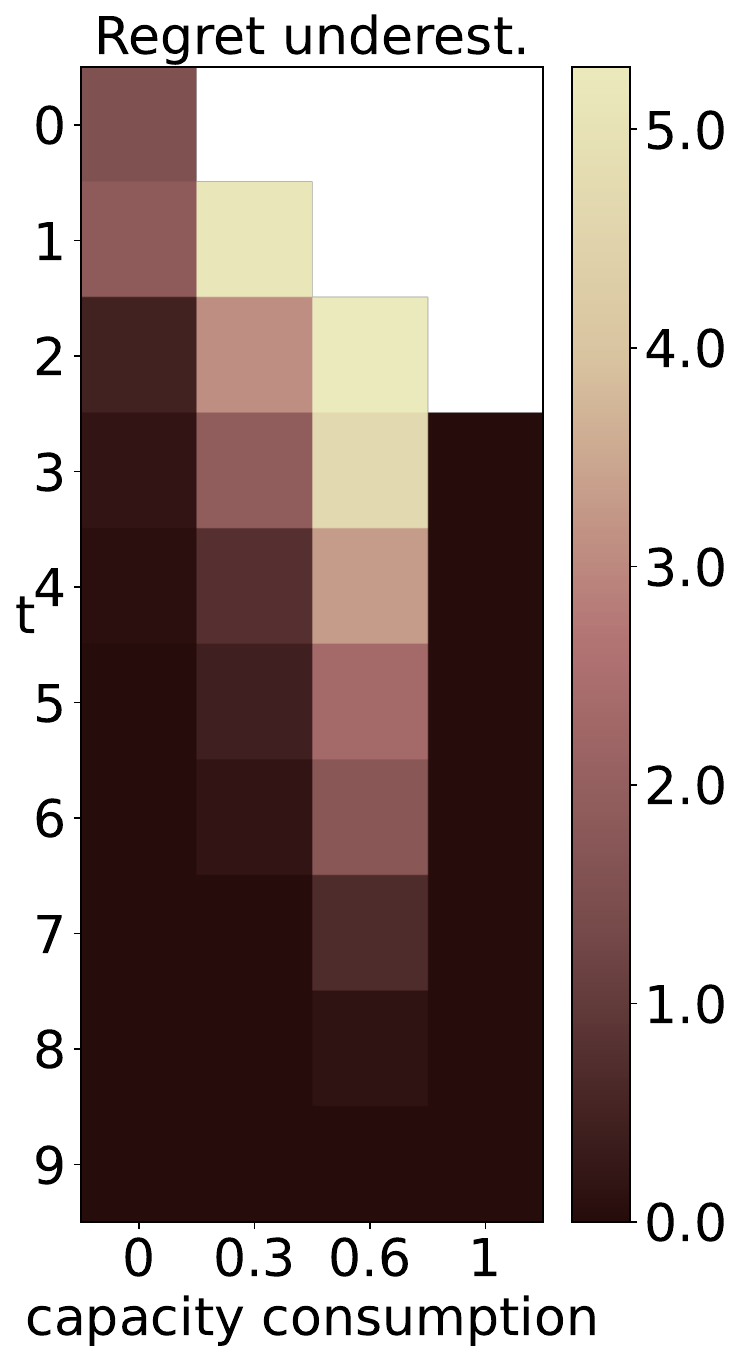}
    		\end{subfigure}
            \begin{subfigure}[]{0.3\textwidth}
    			\centering
    			\includegraphics[width=\textwidth]{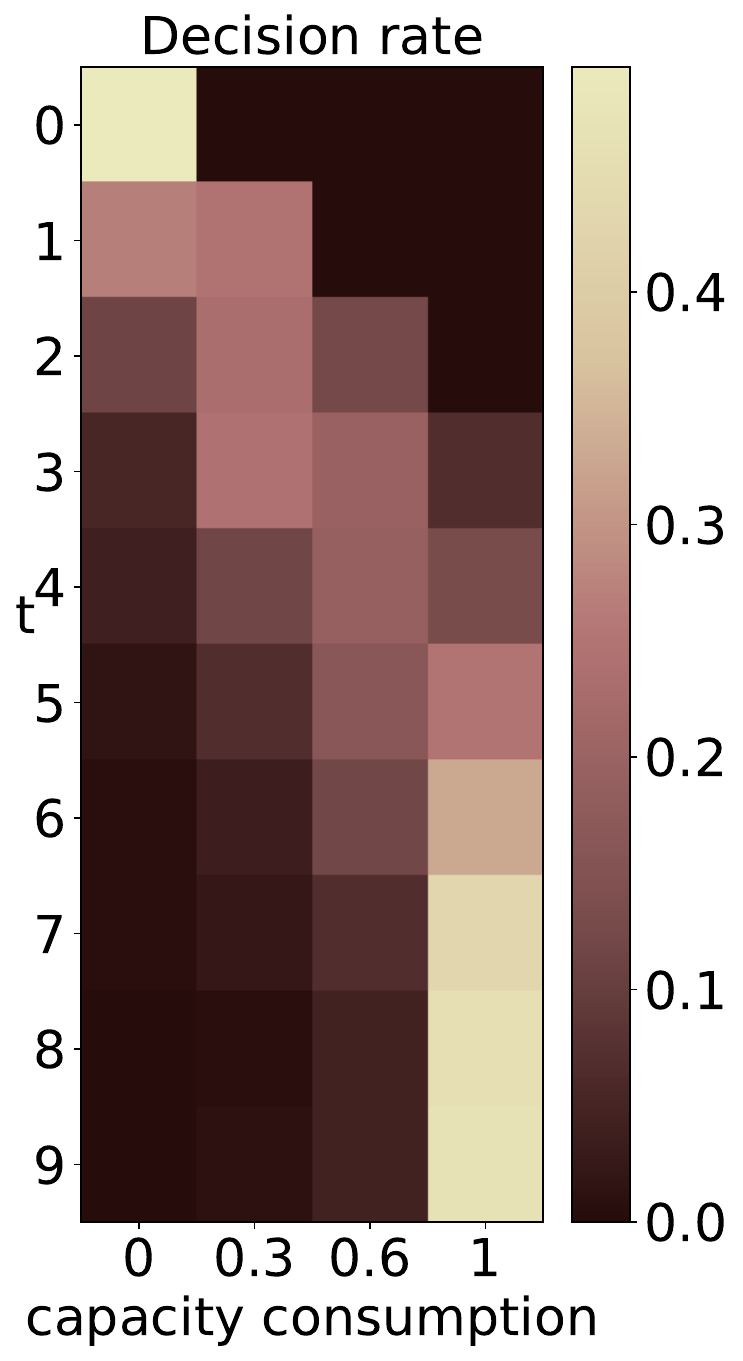}
    		\end{subfigure}
        \caption{high $\mid$ load $\mid$ unif $\mid$ rand}
        \label{fig: HM2a}
        \end{subfigure},
    \vskip\baselineskip
    \begin{subfigure}{0.58\textwidth}
		\renewcommand\thesubfigure{b}
      	\begin{subfigure}[]{0.315\textwidth}
    			\centering
    			\includegraphics[width=\textwidth]{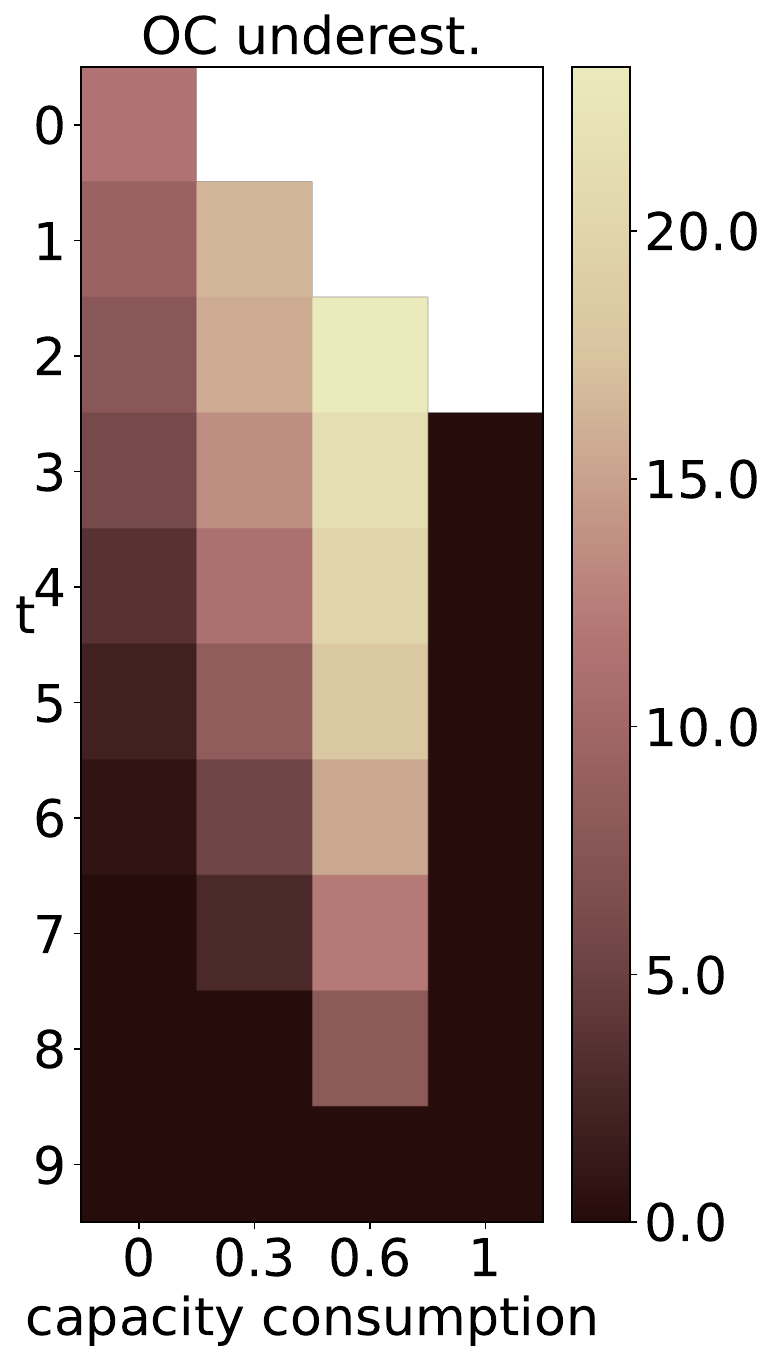}
    		\end{subfigure}
    		\begin{subfigure}[]{0.3\textwidth}
    			\centering
    			\includegraphics[width=\textwidth]{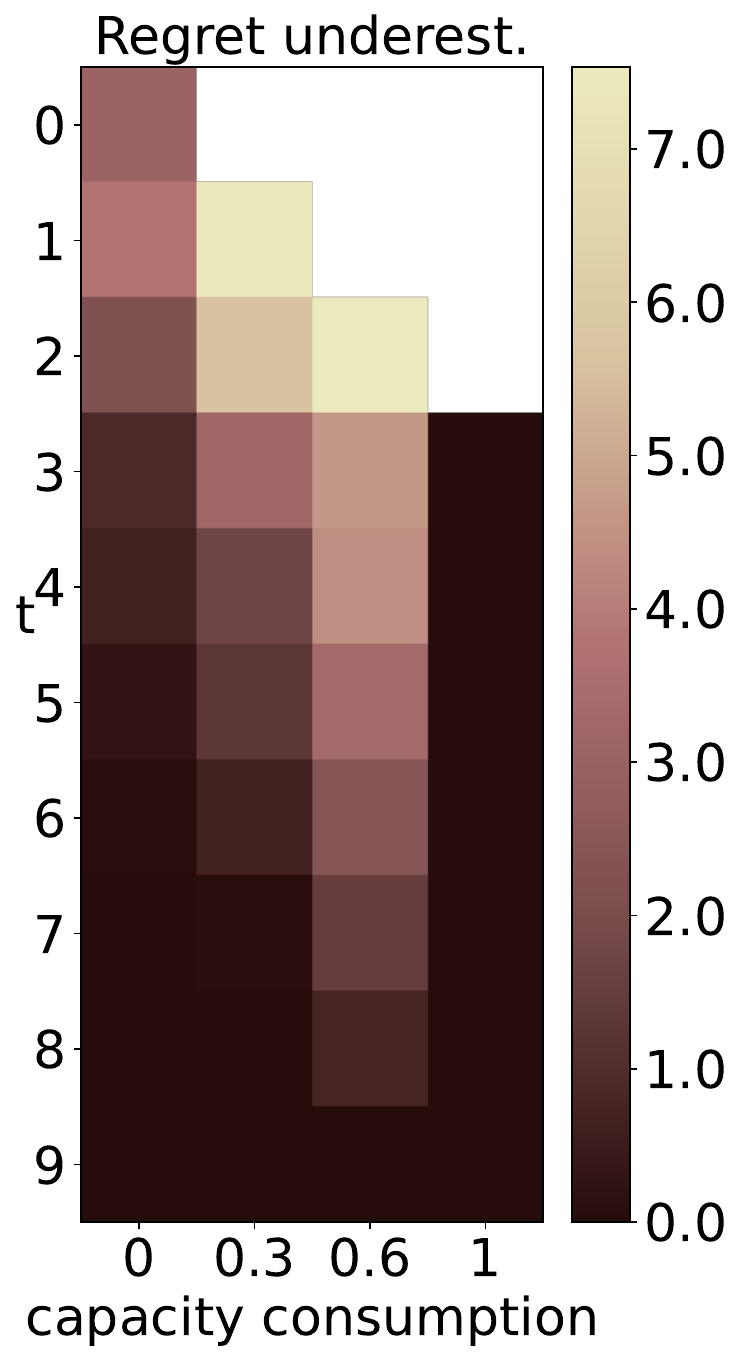}
    		\end{subfigure}
           \begin{subfigure}[]{0.3\textwidth}
    			\centering
    			\includegraphics[width=\textwidth]{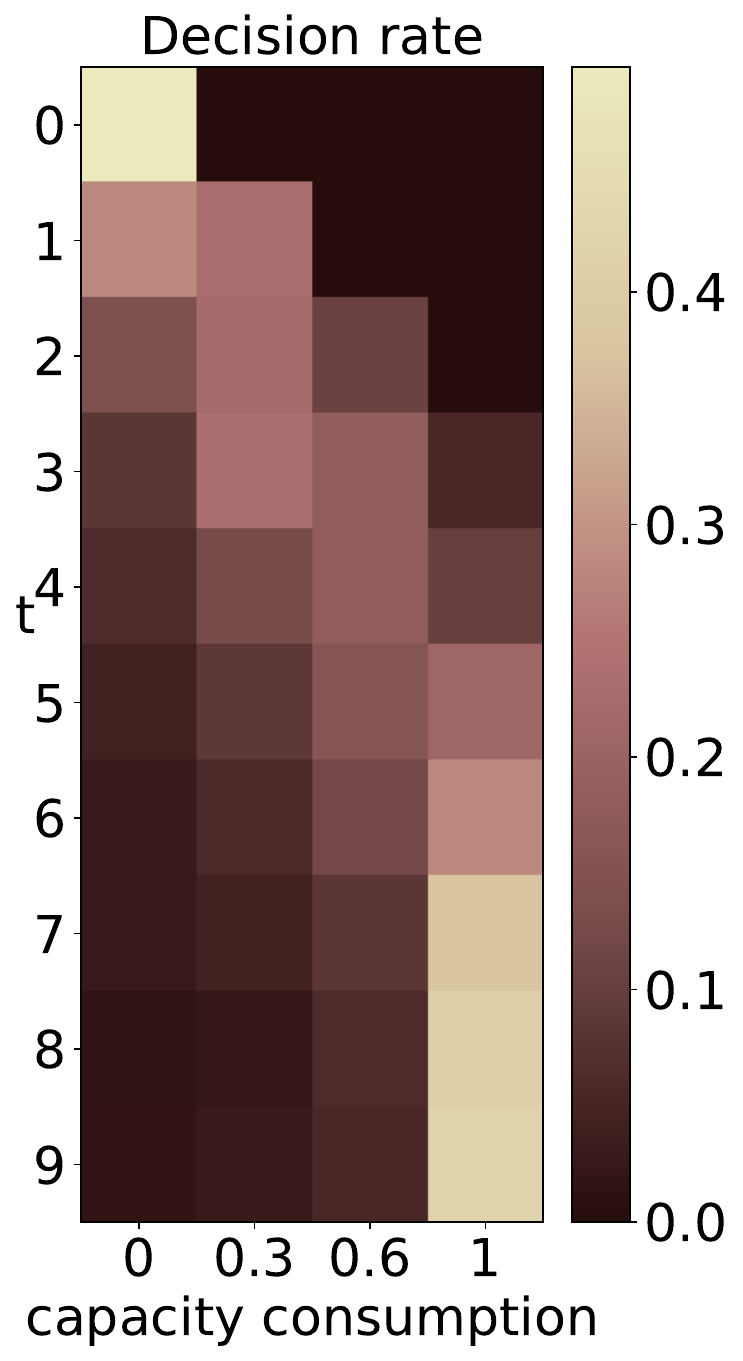}
    		\end{subfigure}
        \caption{med $\mid$ load $\mid$ unif $\mid$ rand }
        \label{fig: HM2b}
        \end{subfigure}
    \vskip\baselineskip
        \begin{subfigure}{0.58\textwidth}
    		\renewcommand\thesubfigure{c}
            \begin{subfigure}[]{0.315\textwidth}
    			\centering
    			\includegraphics[width=\textwidth]{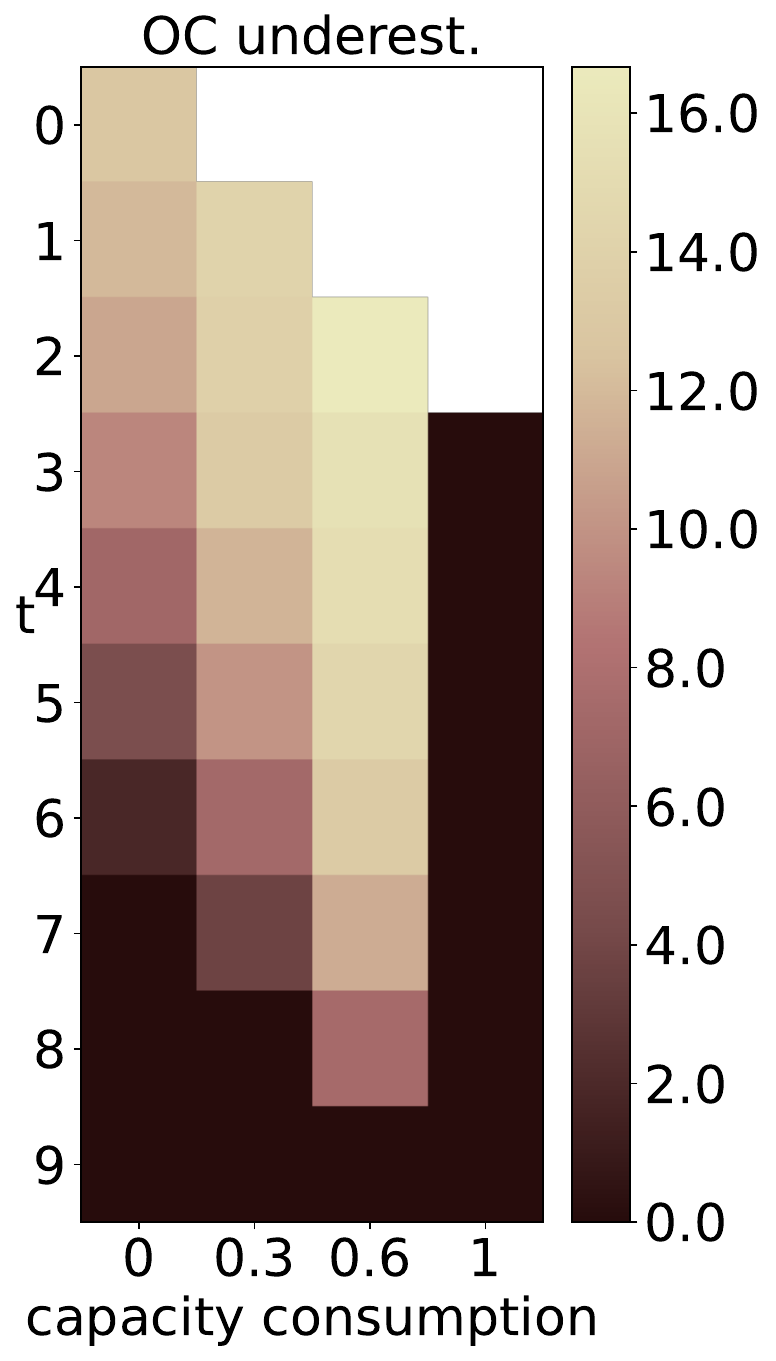}
    		\end{subfigure}
    		\begin{subfigure}[]{0.3\textwidth}
    			\centering
    			\includegraphics[width=\textwidth]{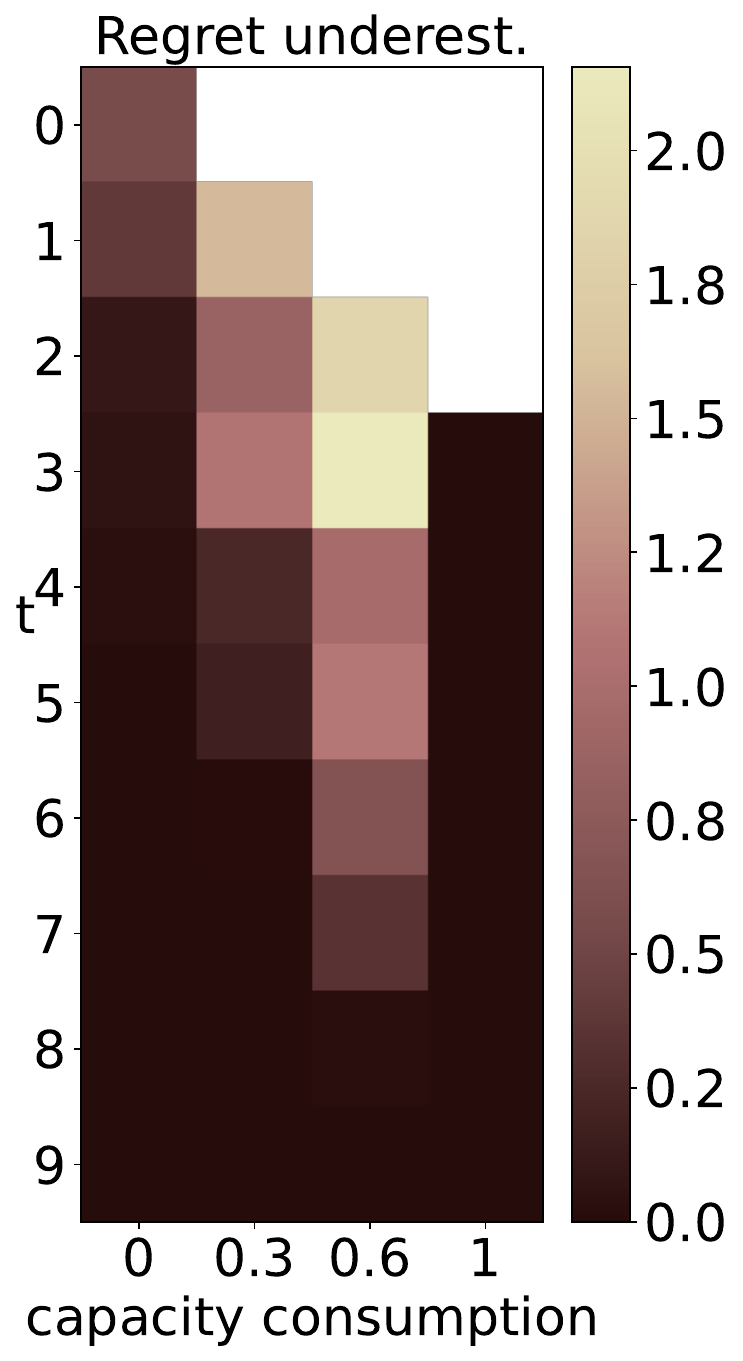}
    		\end{subfigure}
            \begin{subfigure}[]{0.3\textwidth}
    			\centering
    			\includegraphics[width=\textwidth]{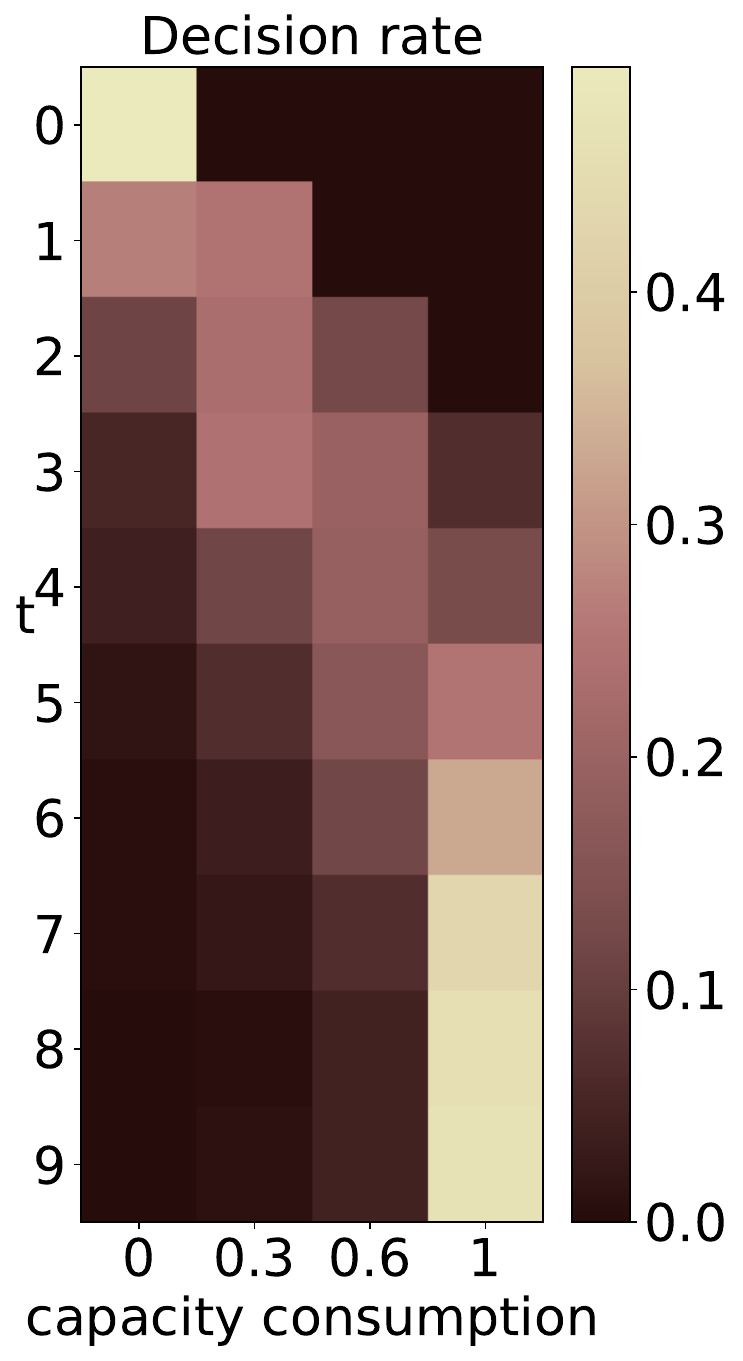}
    		\end{subfigure}
        \caption{high $\mid$ load $\mid$ unif $\mid$ homog }
        \label{fig: HM2c}
        \end{subfigure}
            \vskip\baselineskip
        \begin{subfigure}{0.58\textwidth}
		\renewcommand\thesubfigure{d}
        	\begin{subfigure}[]{0.315\textwidth}
    			\centering
    			\includegraphics[width=\textwidth]{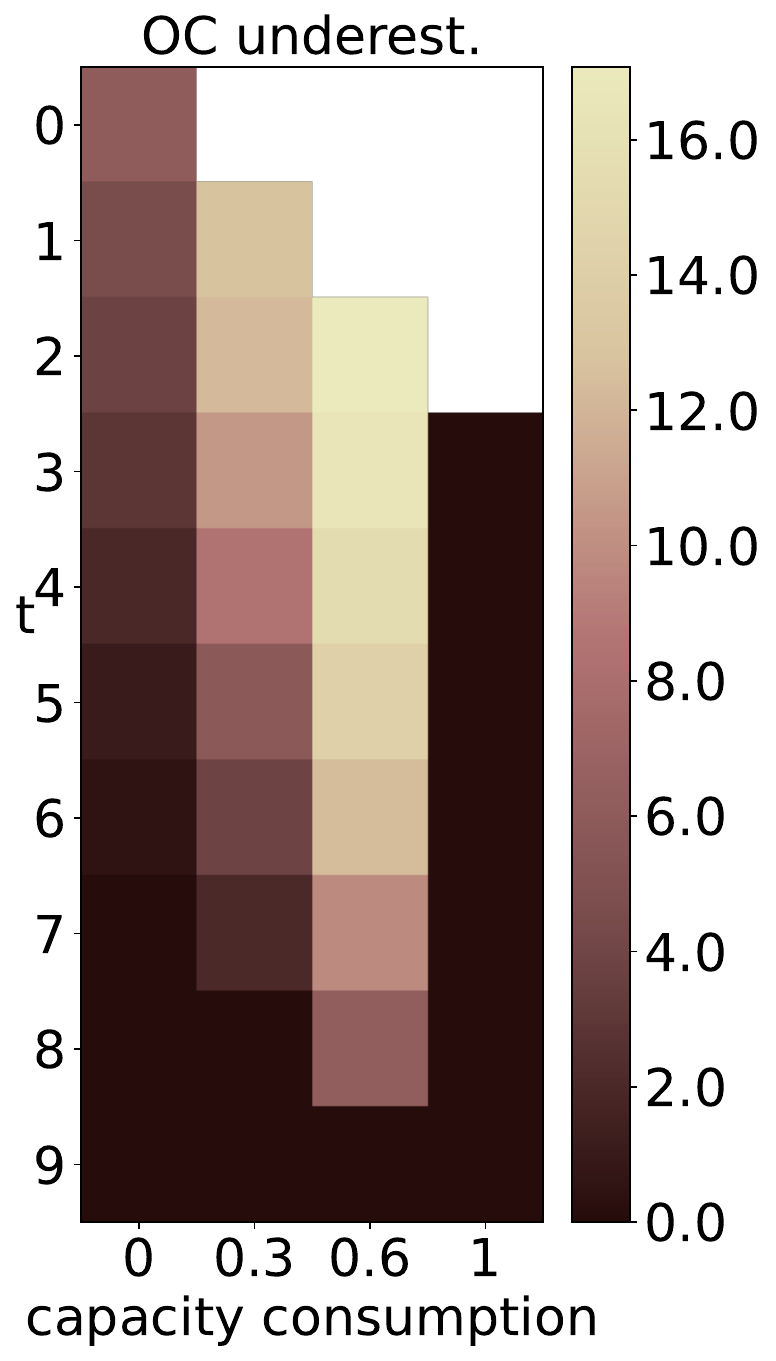}
    		\end{subfigure}
      	\begin{subfigure}[]{0.3\textwidth}
    			\centering
    			\includegraphics[width=\textwidth]{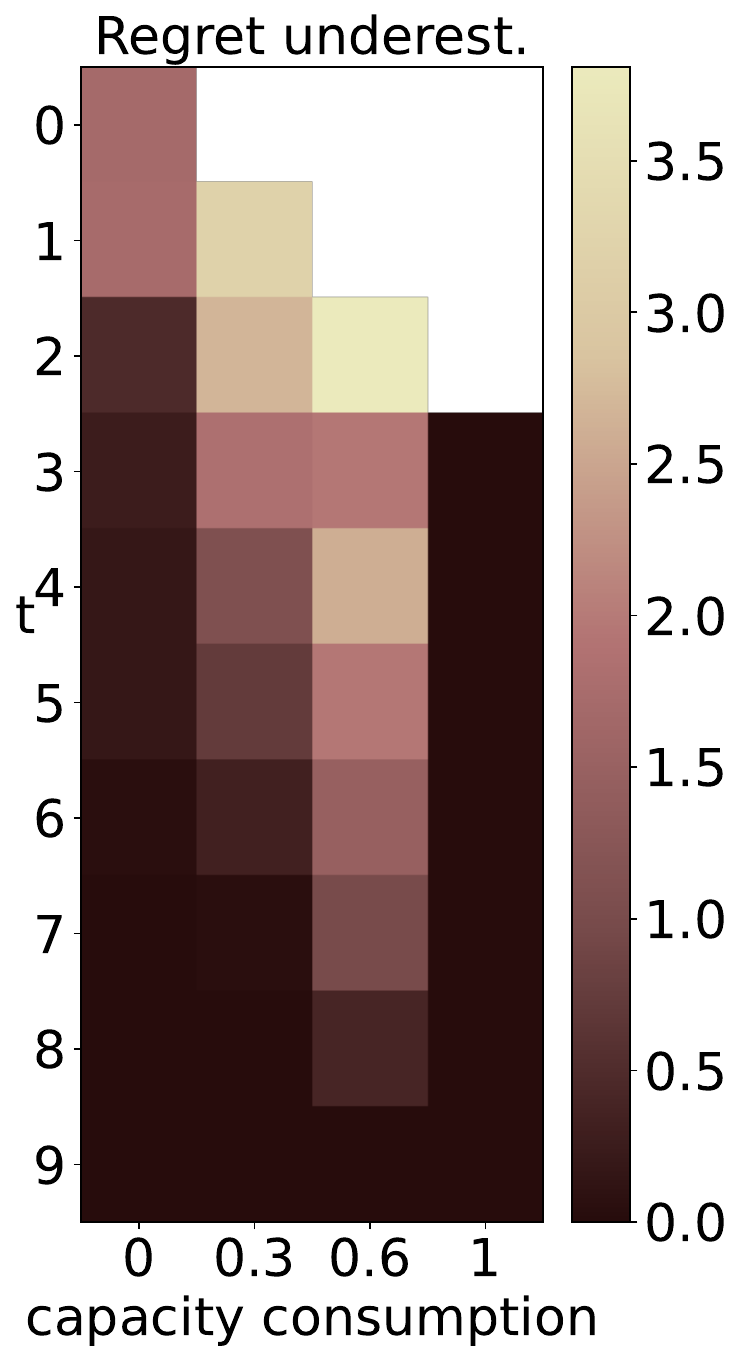}
    		\end{subfigure}
    		\begin{subfigure}[]{0.3\textwidth}
    			\centering
    			\includegraphics[width=\textwidth]{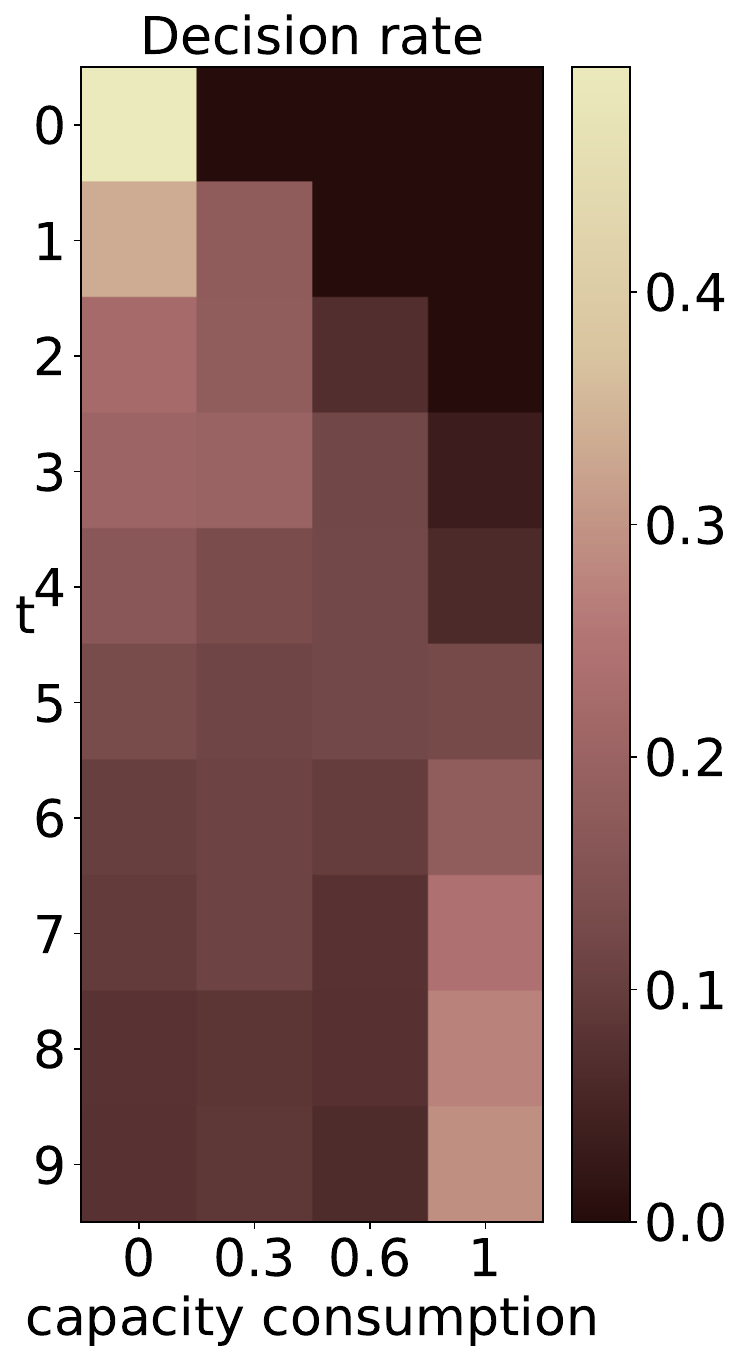}
    		\end{subfigure}
        \caption{med $\mid$ load $\mid$ unif $\mid$ homog }
        \label{fig: HM2d}
        \end{subfigure}
\end{figure}
\vspace{-1cm}
\begin{figure}[H]
\centering
\footnotesize
\caption{MCTS policy - Overestimation error}
\label{fig: HM3}
    \vskip\baselineskip
	\centering
	\captionsetup{justification=centering}
        \begin{subfigure}{0.6\textwidth}
    		\renewcommand\thesubfigure{a}
            \begin{subfigure}[]{0.3\textwidth}
    			\centering
    			\includegraphics[width=\textwidth]{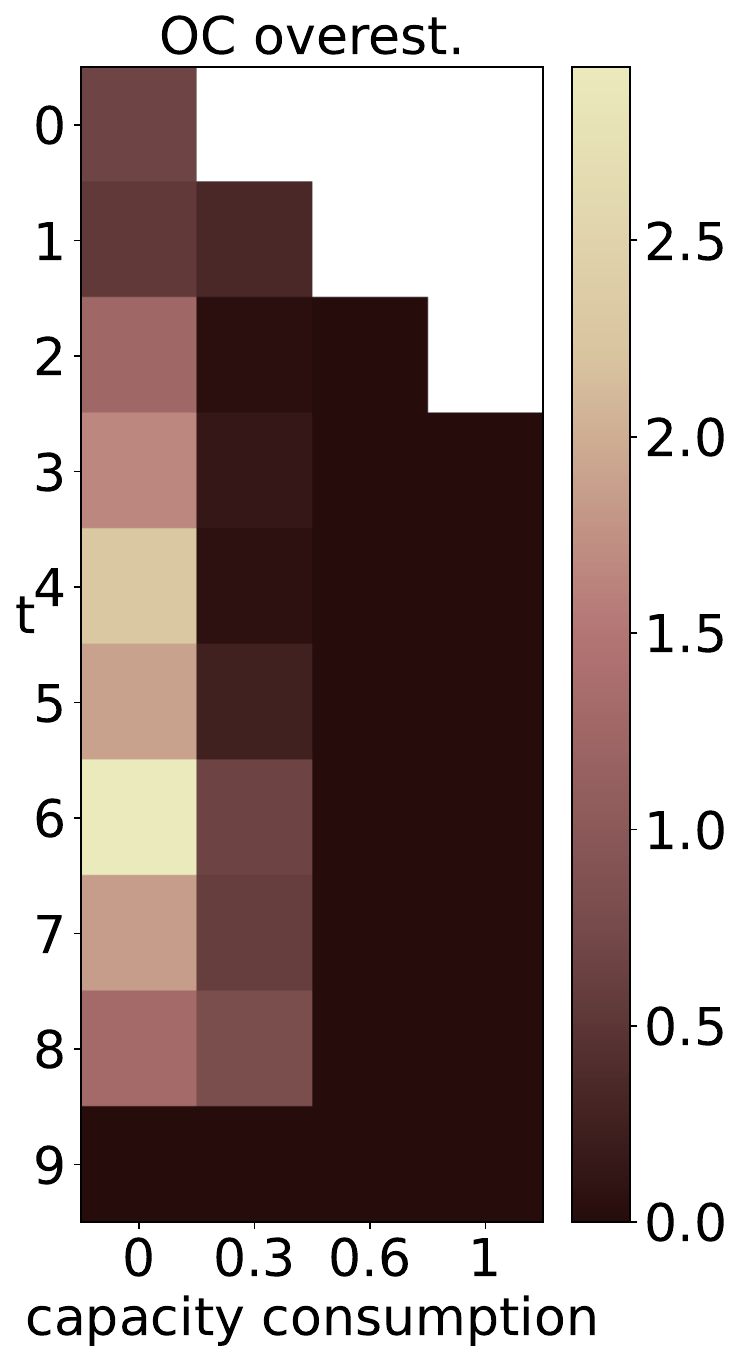}
    		\end{subfigure}
    		\begin{subfigure}[]{0.3\textwidth}
    			\centering
    			\includegraphics[width=\textwidth]{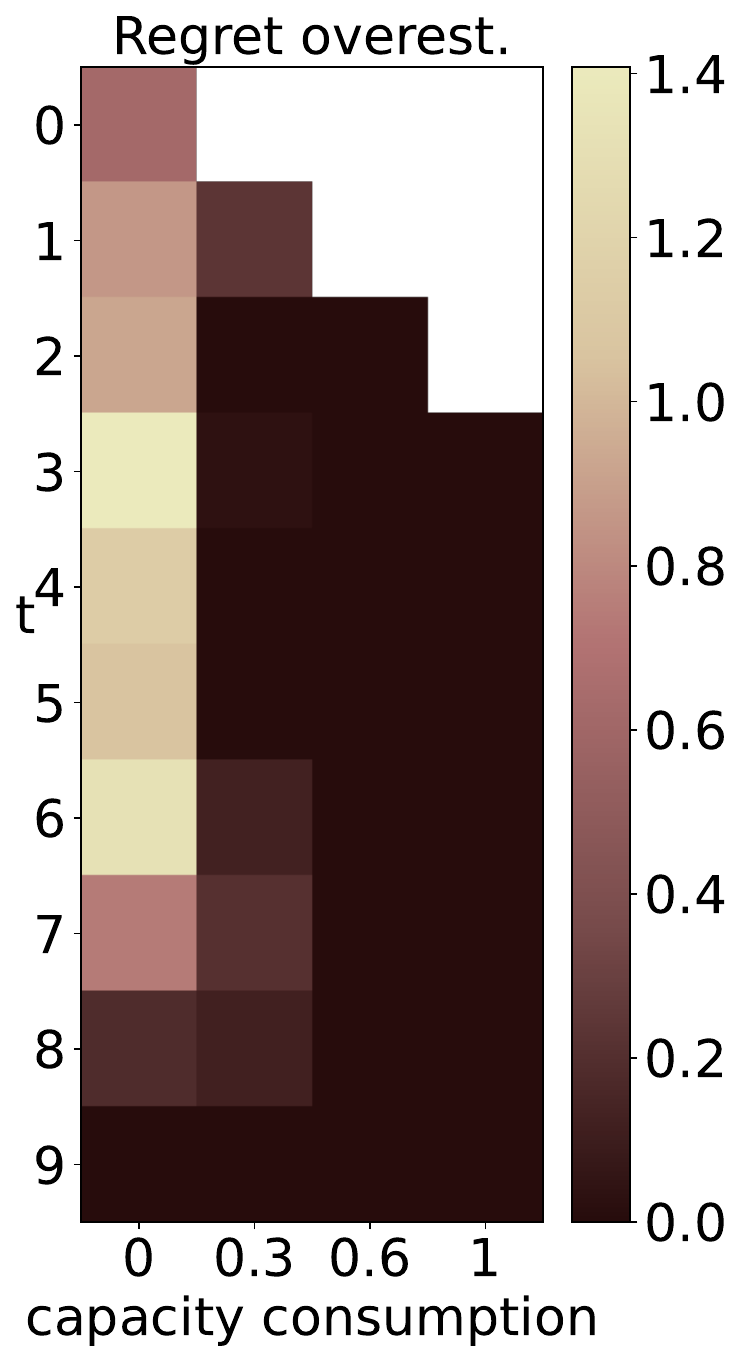}
    		\end{subfigure}
            \begin{subfigure}[]{0.3\textwidth}
    			\centering
    			\includegraphics[width=\textwidth]{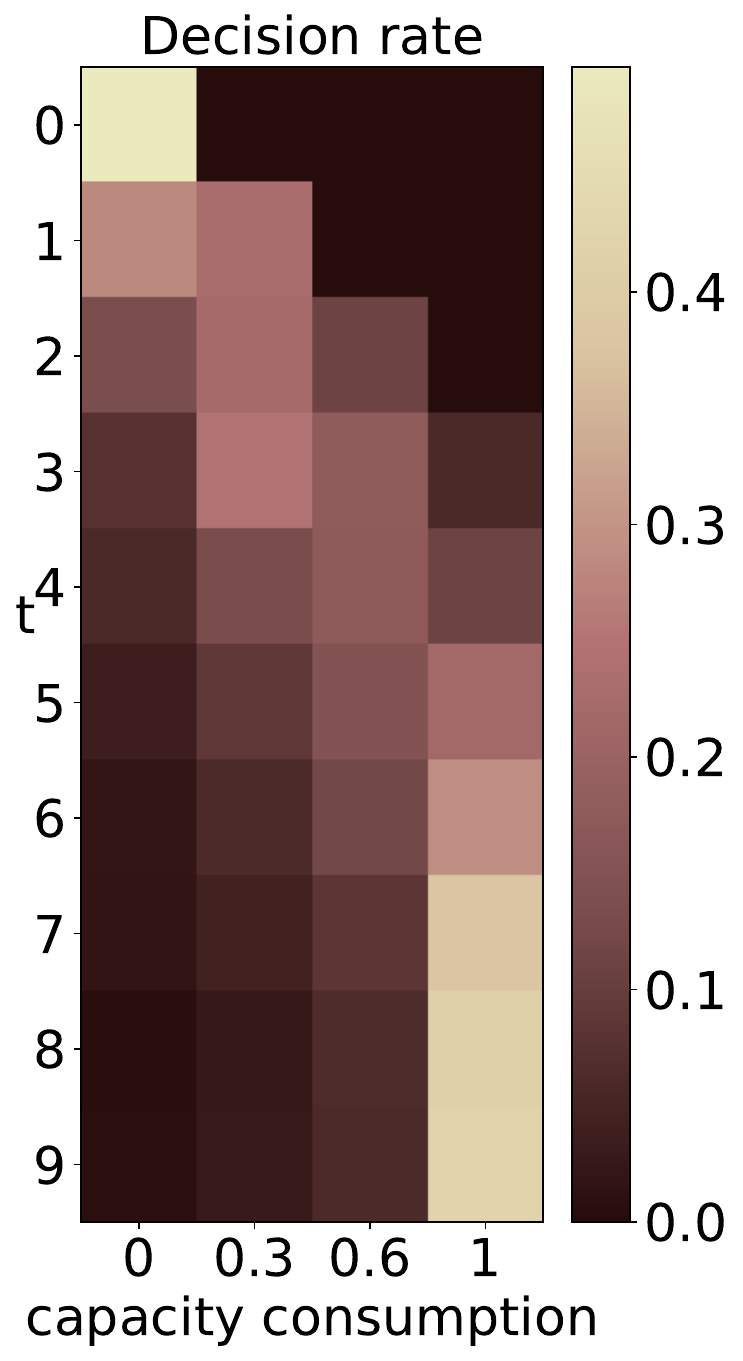}
    		\end{subfigure}
        \caption{med $\mid$ load $\mid$ clust\_sort $\mid$ rand}
        \label{fig: HM3a}
        \end{subfigure}
    \vskip\baselineskip
        \begin{subfigure}{0.6\textwidth}
    		\renewcommand\thesubfigure{b}
            \begin{subfigure}[]{0.3\textwidth}
    			\centering
    			\includegraphics[width=\textwidth]{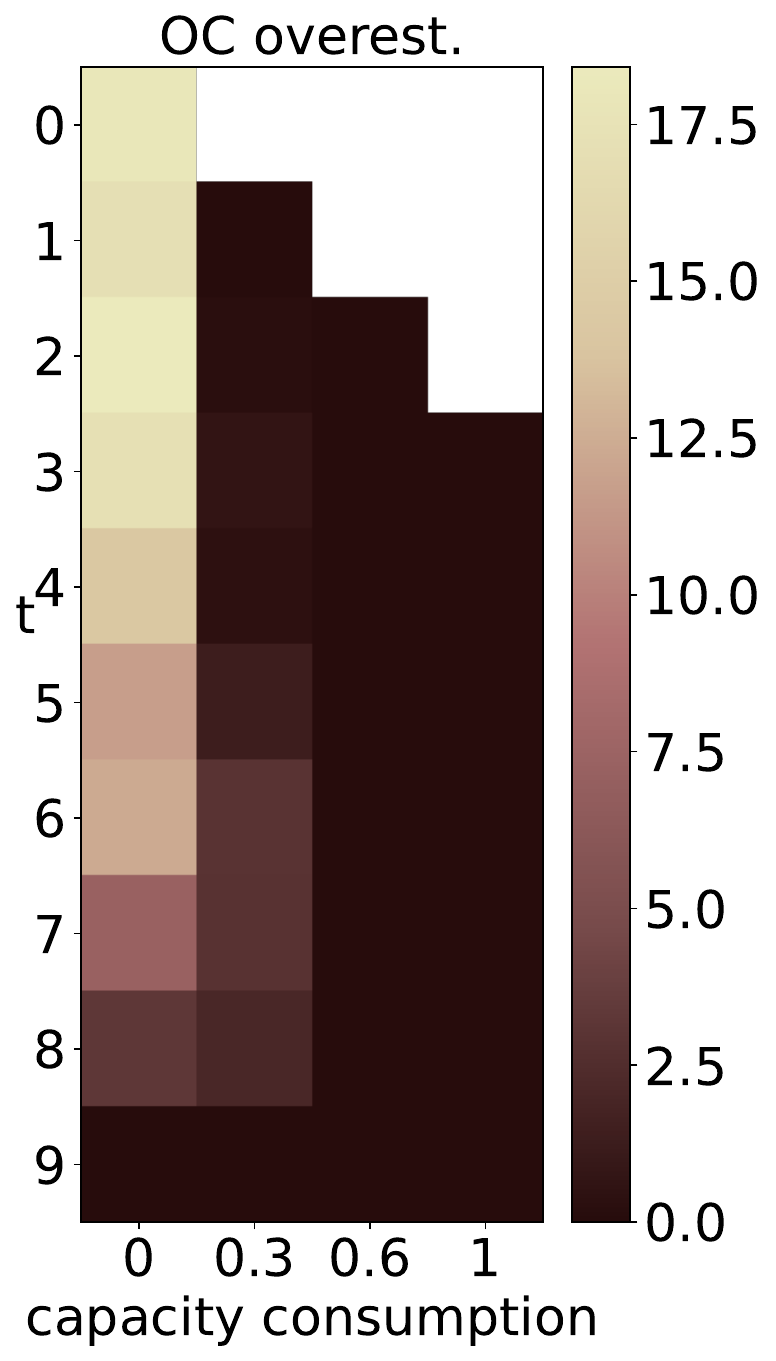}
    		\end{subfigure}
            \begin{subfigure}[]{0.3\textwidth}
    			\centering
    			\includegraphics[width=\textwidth]{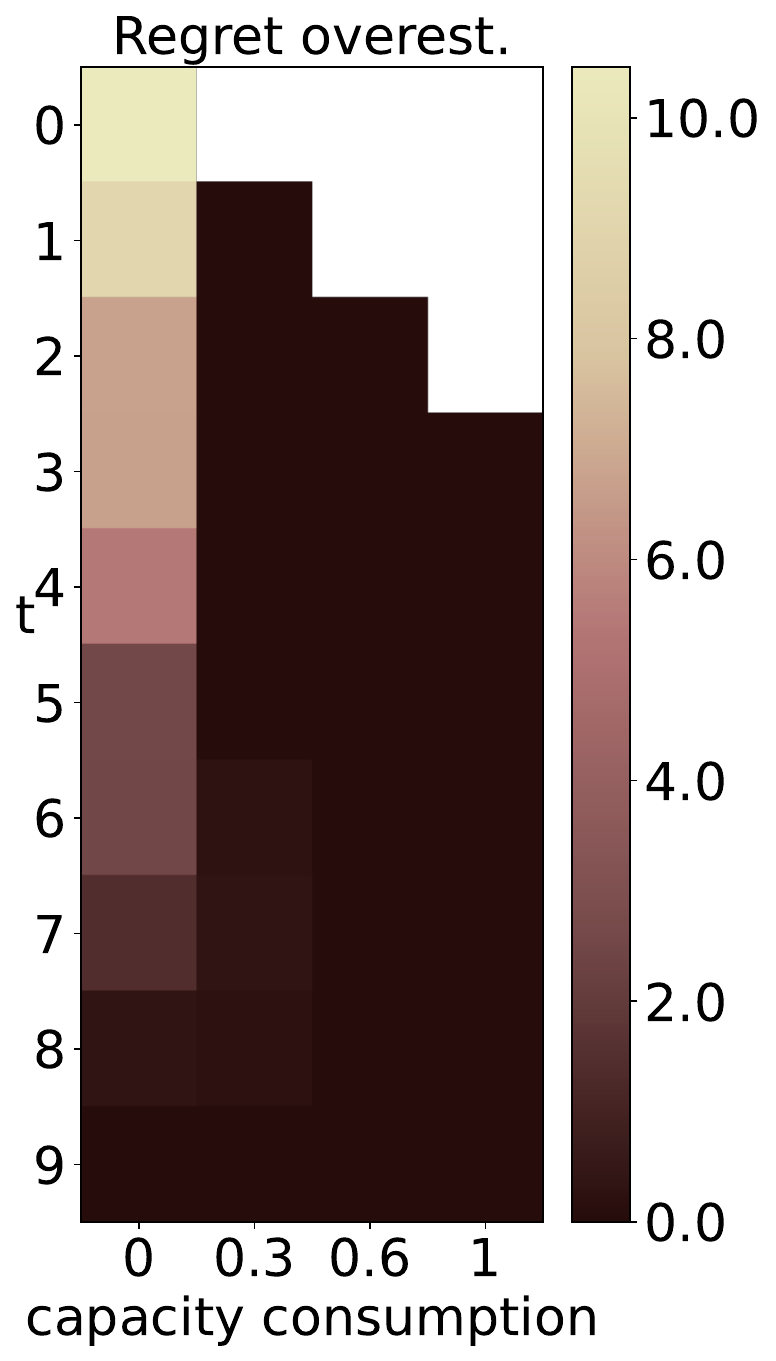}
    		\end{subfigure}
    		\begin{subfigure}[]{0.29\textwidth}
    			\centering
    			\includegraphics[width=\textwidth]{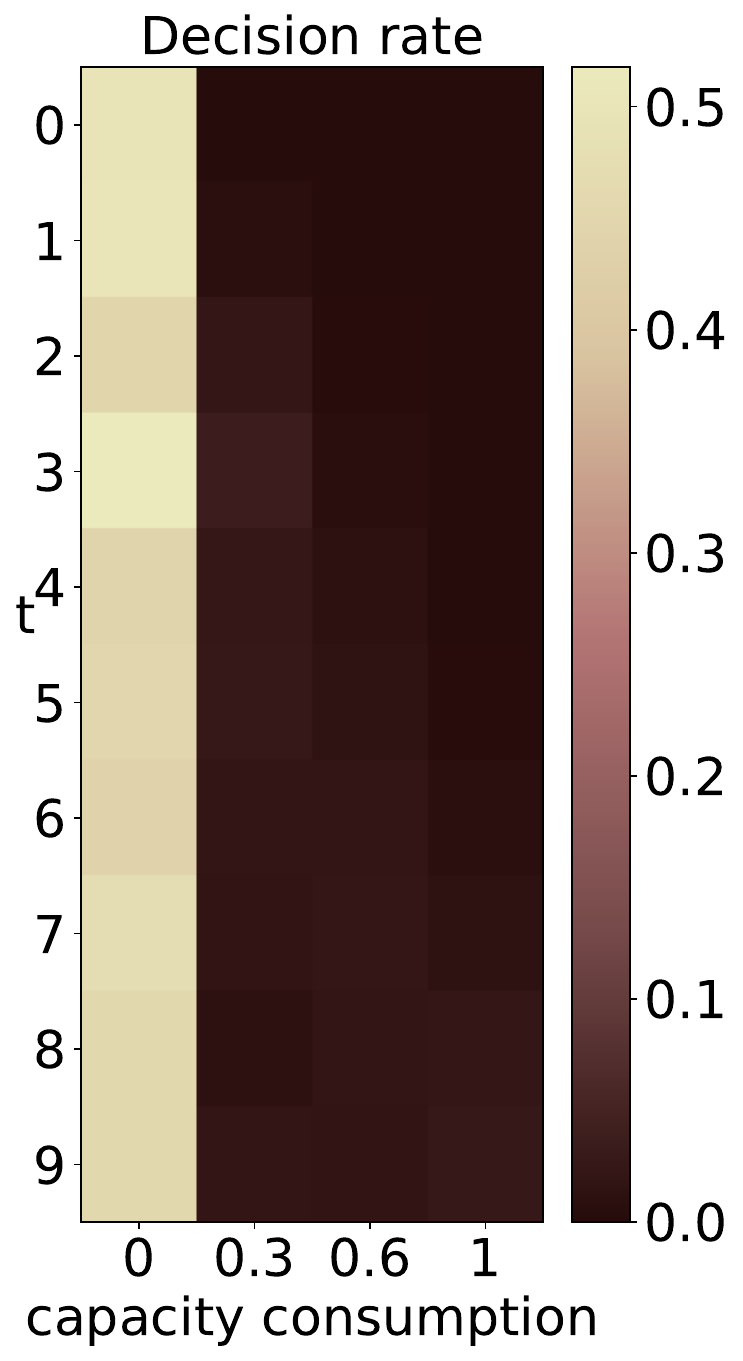}
    		\end{subfigure}
        \caption{low $\mid$ load $\mid$ clust\_sort $\mid$ rand}
        \label{fig: HM3b}
        \end{subfigure}
            \vskip\baselineskip
        \begin{subfigure}{0.8\textwidth}
		\renewcommand\thesubfigure{c}
        	\begin{subfigure}[]{0.3\textwidth}
    			\centering
    			\includegraphics[width=\textwidth]{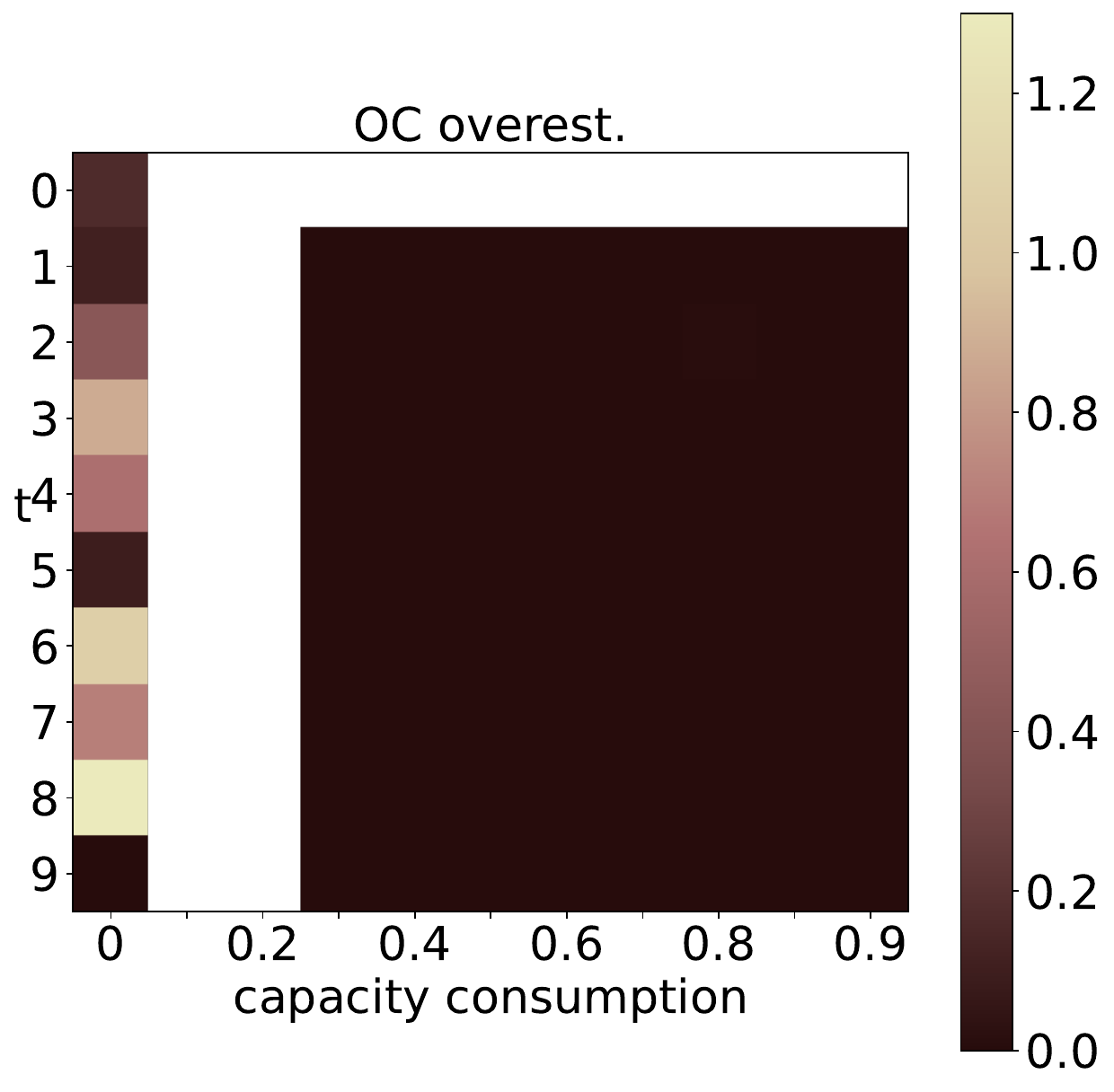}
    		\end{subfigure}
      	\begin{subfigure}[]{0.3\textwidth}
    			\centering
    			\includegraphics[width=\textwidth]{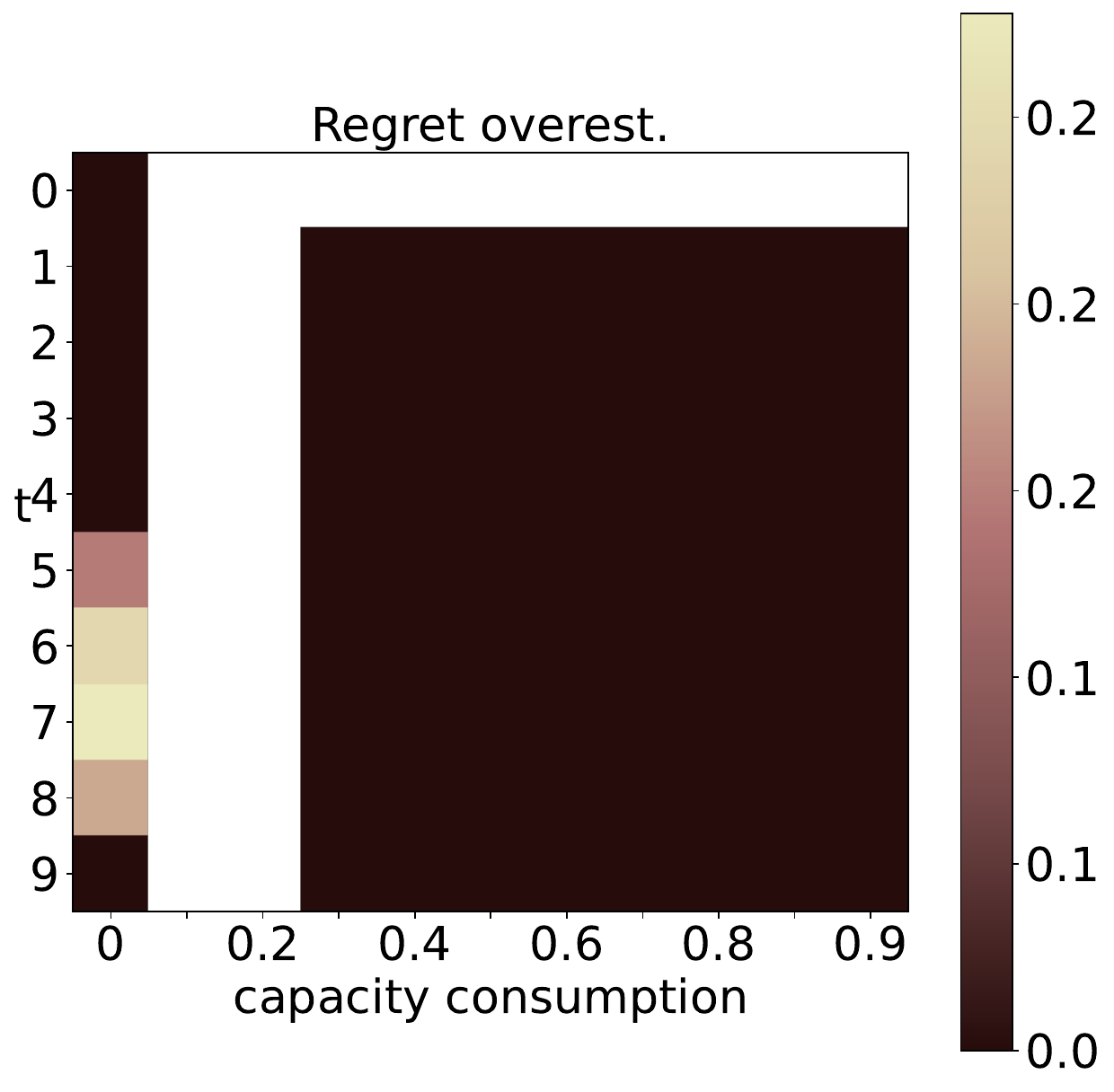}
    		\end{subfigure}
    		\begin{subfigure}[]{0.3\textwidth}
    			\centering
    			\includegraphics[width=\textwidth]{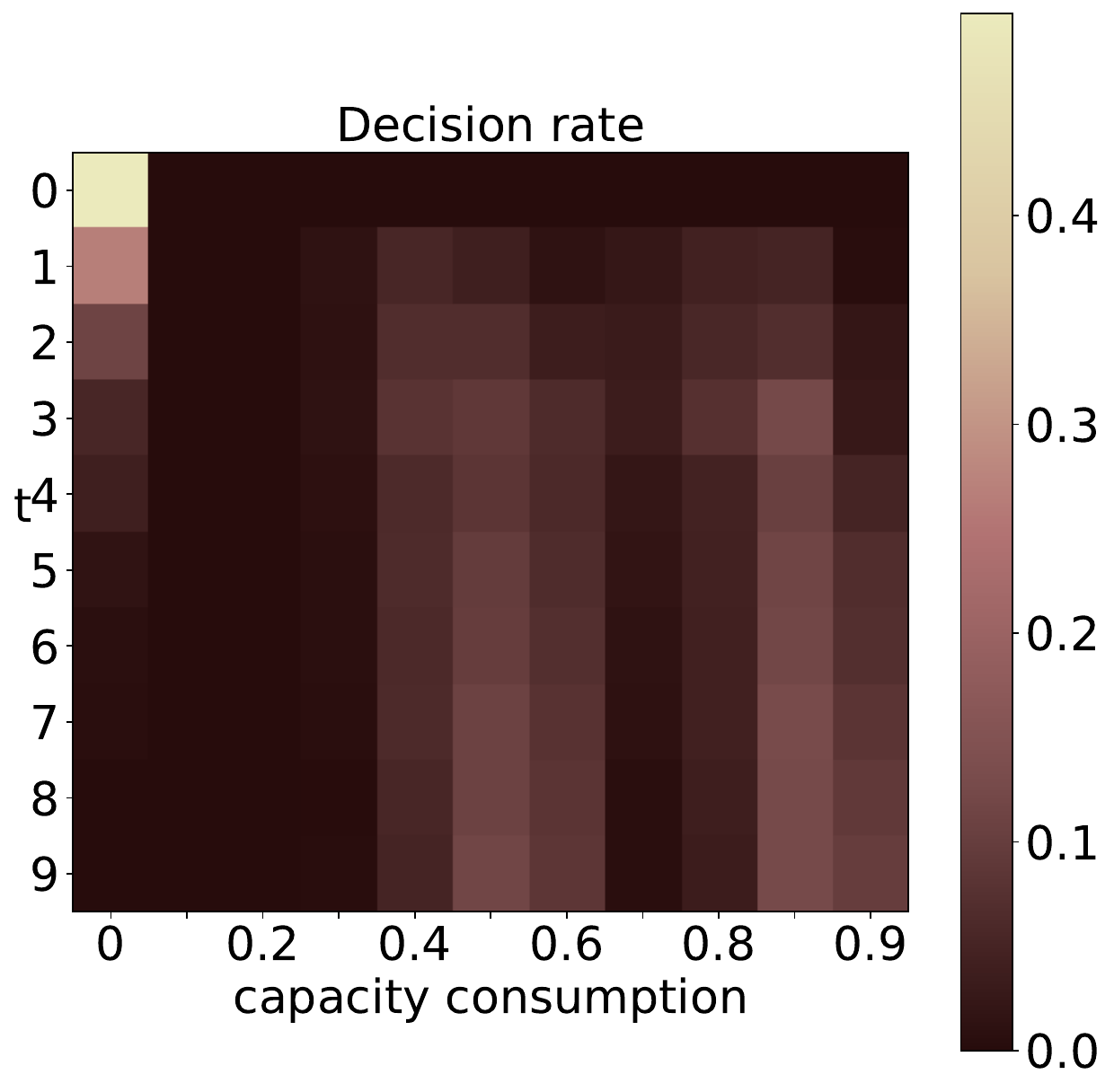}
    		\end{subfigure}
        \caption{med $\mid$ dist $\mid$ clust\_sort $\mid$}%
        \label{fig: HM3c}
        \end{subfigure}
    \vskip\baselineskip
    	\begin{subfigure}{0.8\textwidth}
	\renewcommand\thesubfigure{d}      	
        \begin{subfigure}[]{0.3\textwidth}
    			\centering
    			\includegraphics[width=\textwidth]{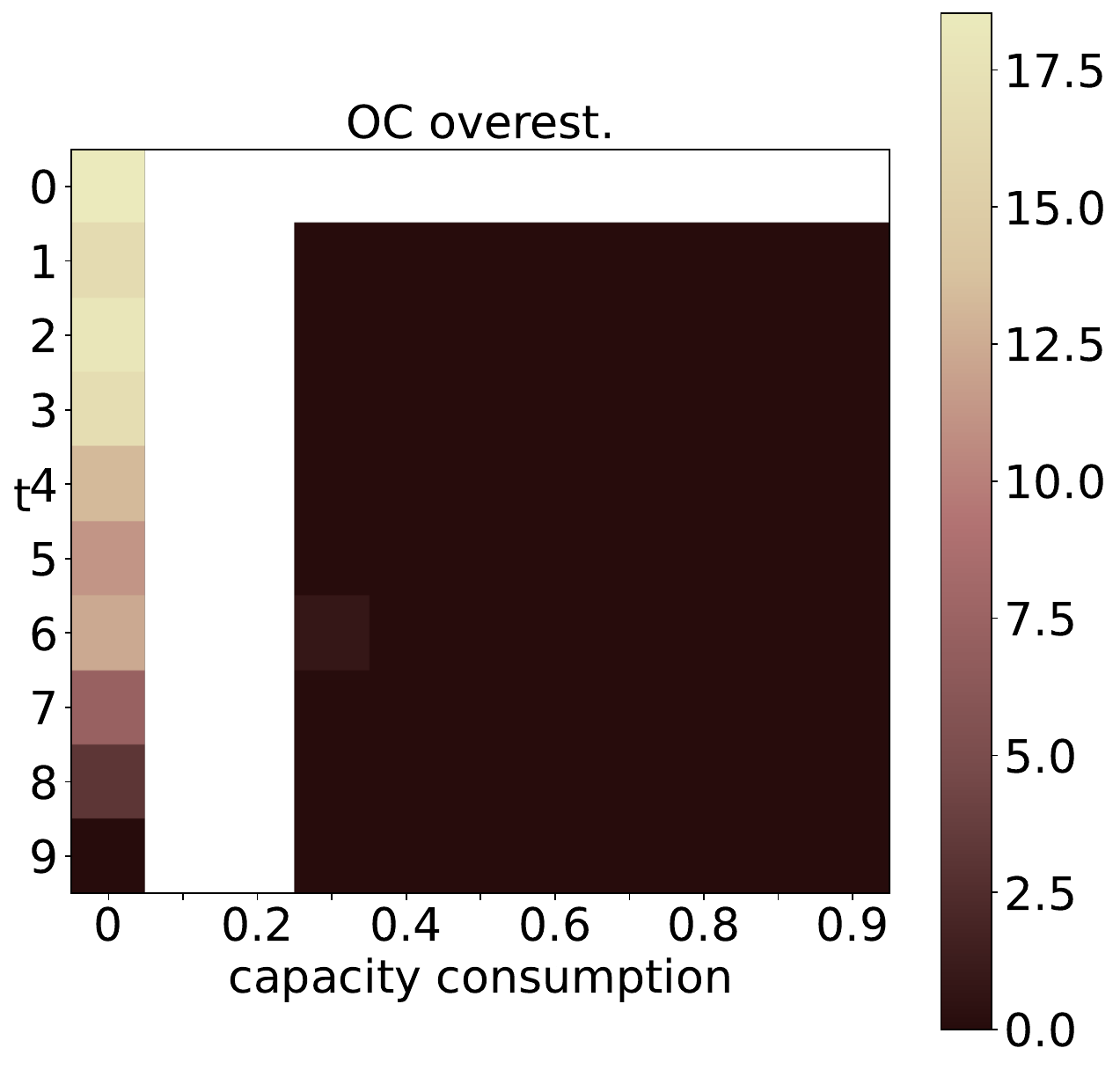}
    		\end{subfigure}
      	\begin{subfigure}[]{0.3\textwidth}
    			\centering
    			\includegraphics[width=\textwidth]{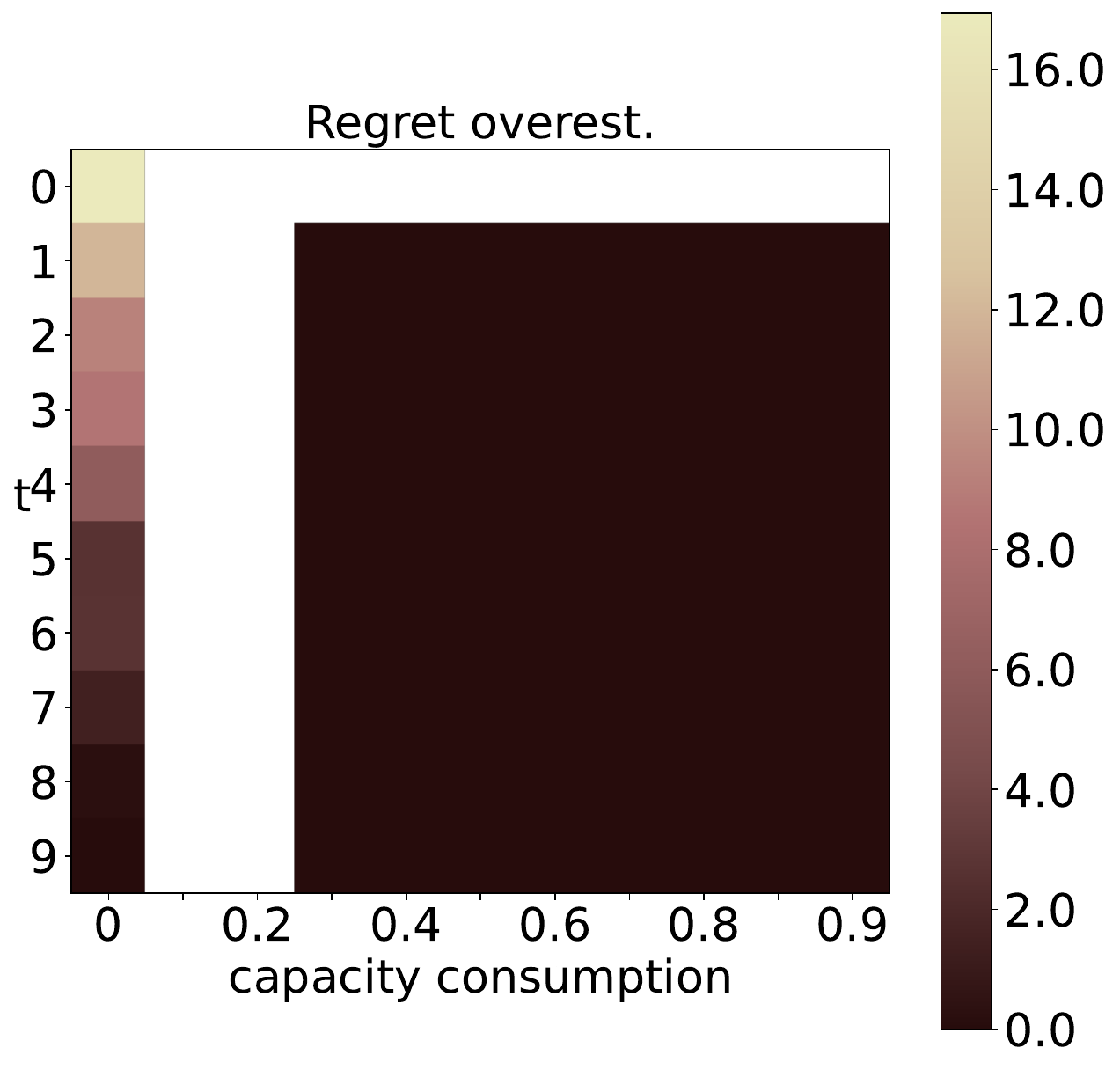}
    		\end{subfigure}
    		\begin{subfigure}[]{0.3\textwidth}
    			\centering
    			\includegraphics[width=\textwidth]{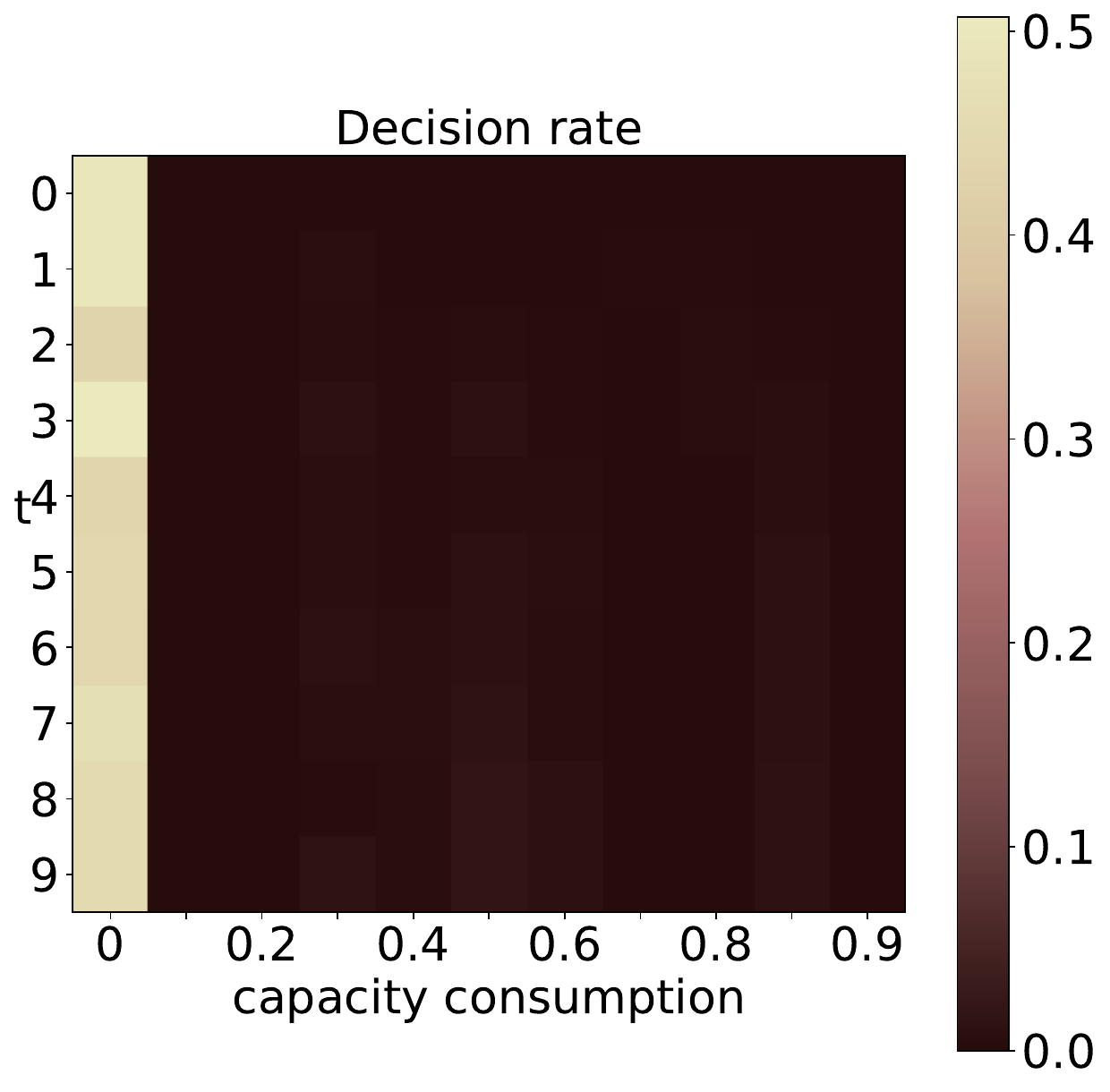}
    		\end{subfigure}
            \caption{low $\mid$ dist $\mid$ clust\_sort $\mid$ rand}
            \label{fig: HM3d}
        \end{subfigure}
    \end{figure}
\vspace{-3cm}
\begin{figure}[H]
\centering
\footnotesize
\caption{MCTS policy - Overestimation error (cont'd)}
\label{fig: HM4}    
    \vskip\baselineskip
	\centering
	\captionsetup{justification=centering}
        \begin{subfigure}{0.6\textwidth}
    		\renewcommand\thesubfigure{a}
                  \begin{subfigure}[]{0.3\textwidth}
    			\centering
    			\includegraphics[width=\textwidth]{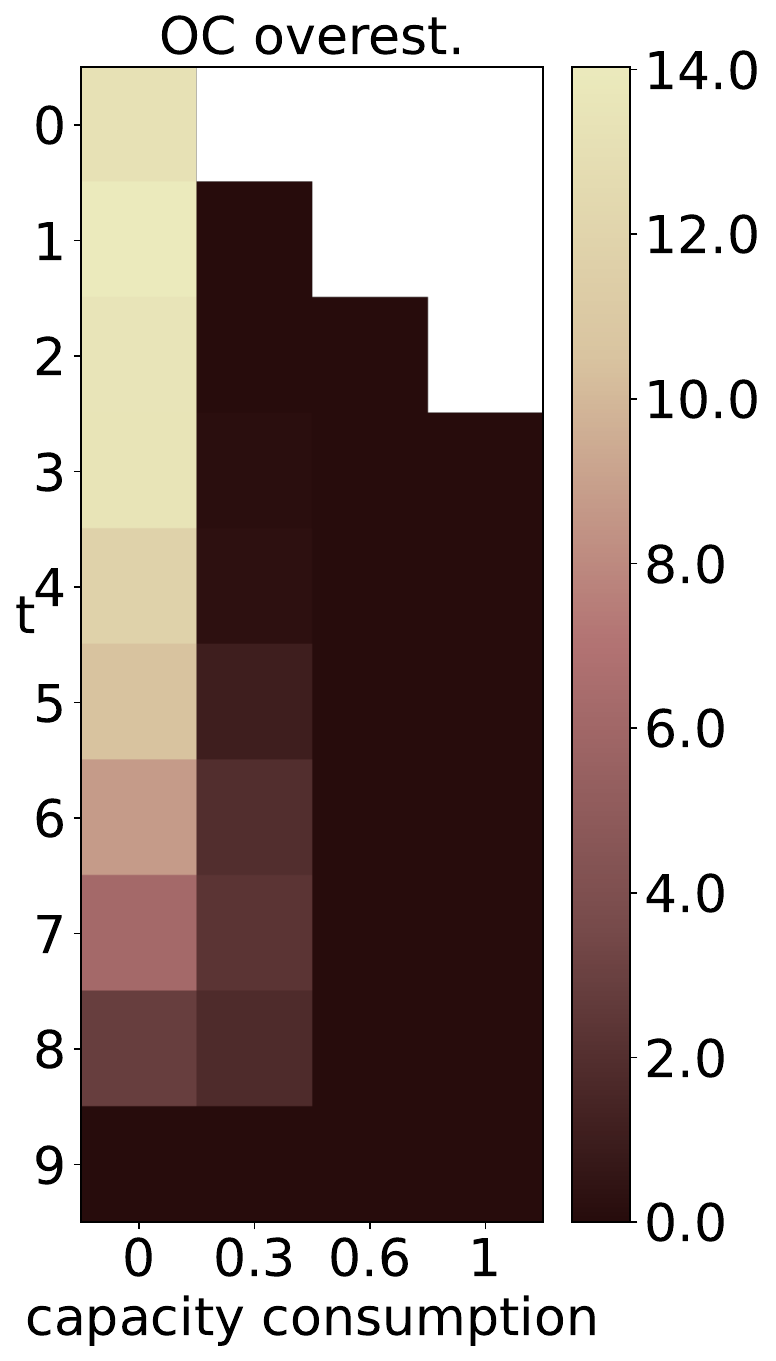}
    		\end{subfigure}
            \begin{subfigure}[]{0.3\textwidth}
    			\centering
    			\includegraphics[width=\textwidth]{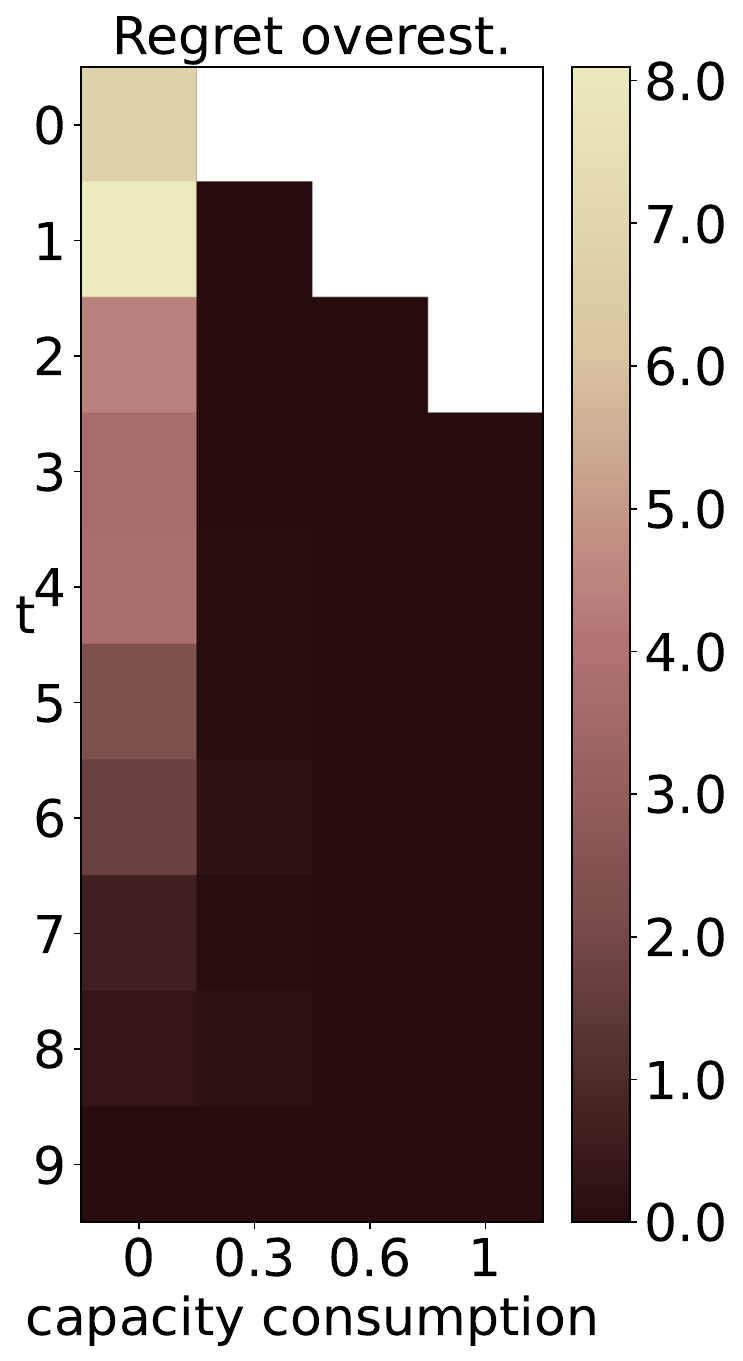}
    		\end{subfigure}
    		\begin{subfigure}[]{0.3\textwidth}
    			\centering
    			\includegraphics[width=\textwidth]{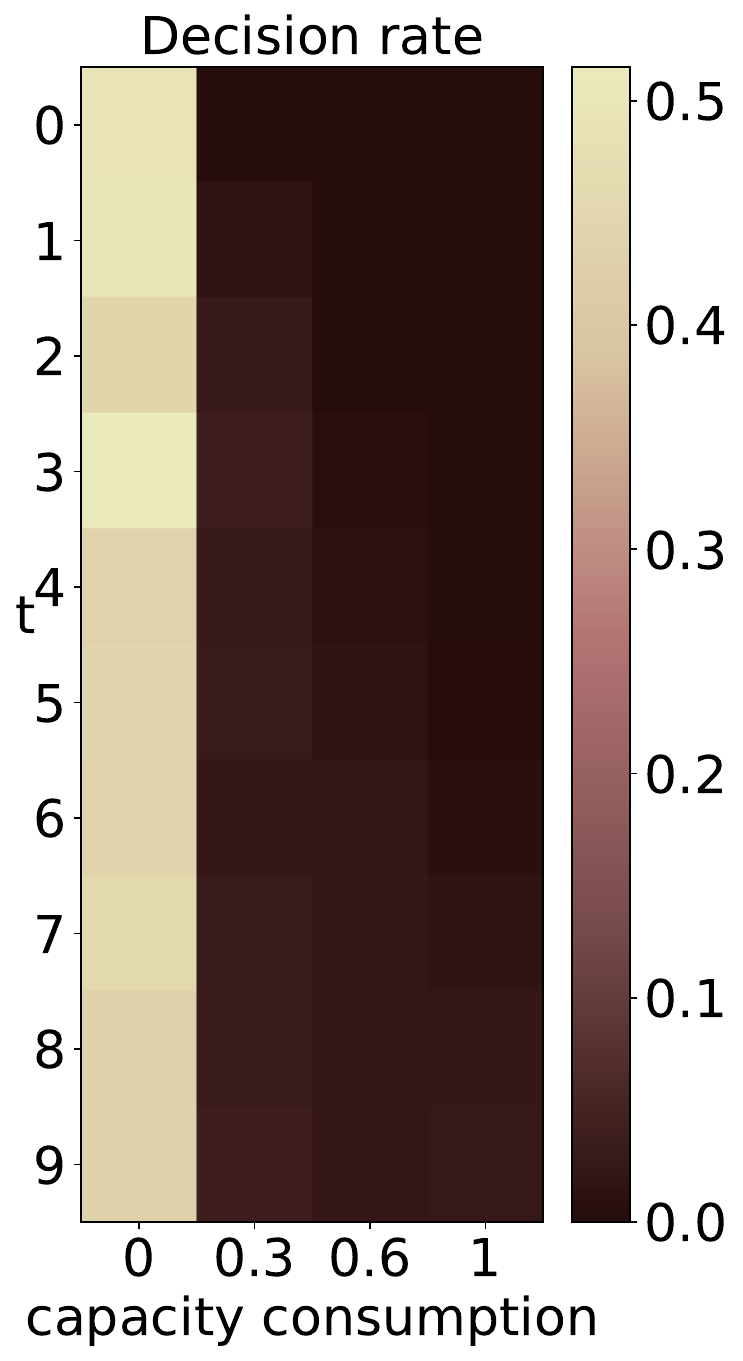}
    		\end{subfigure}
        \caption{low $\mid$ load $\mid$ clust $\mid$ homog }
        \label{fig: HM4a}
        \end{subfigure}
            \vskip\baselineskip
        \begin{subfigure}{0.6\textwidth}
		\renewcommand\thesubfigure{b}
        	\begin{subfigure}[]{0.3\textwidth}
    			\centering
    			\includegraphics[width=\textwidth]{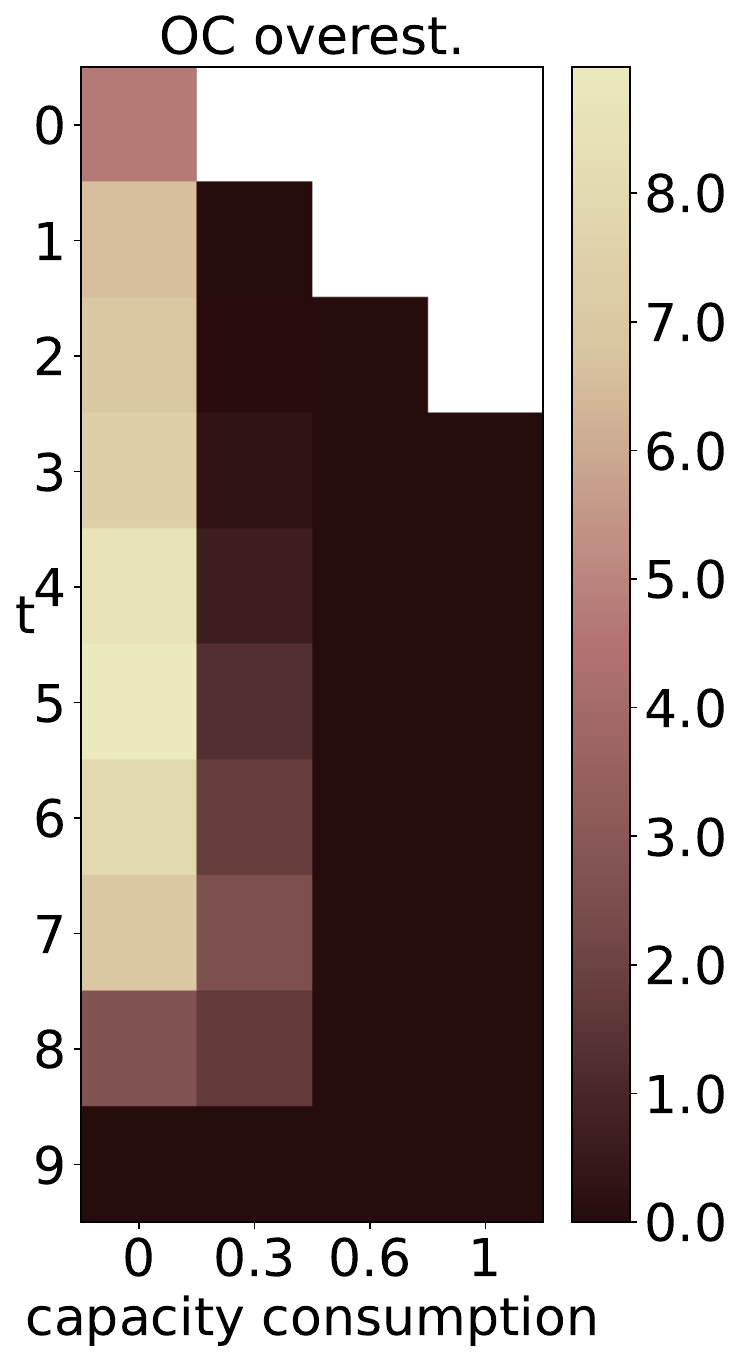}
    		\end{subfigure}
      	\begin{subfigure}[]{0.3\textwidth}
    			\centering
    			\includegraphics[width=\textwidth]{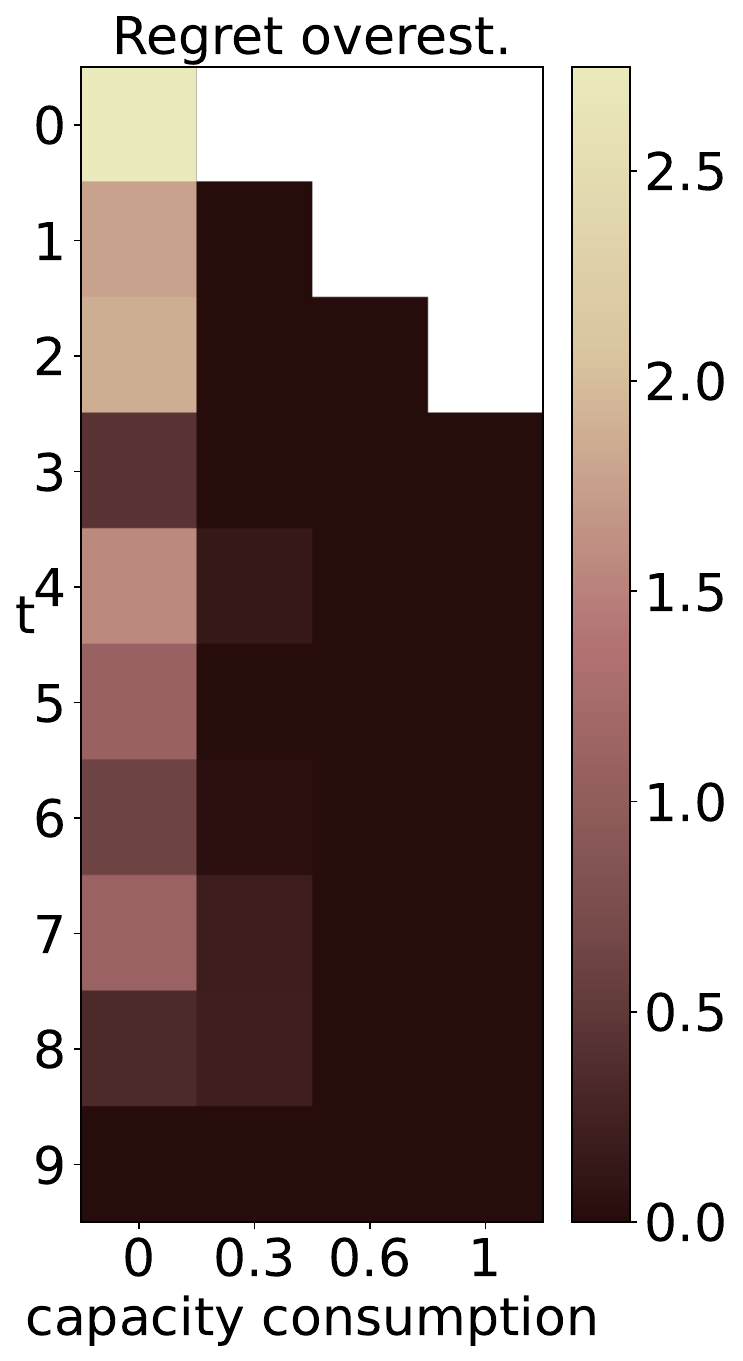}
    		\end{subfigure}
    		\begin{subfigure}[]{0.3\textwidth}
    			\centering
    			\includegraphics[width=\textwidth]{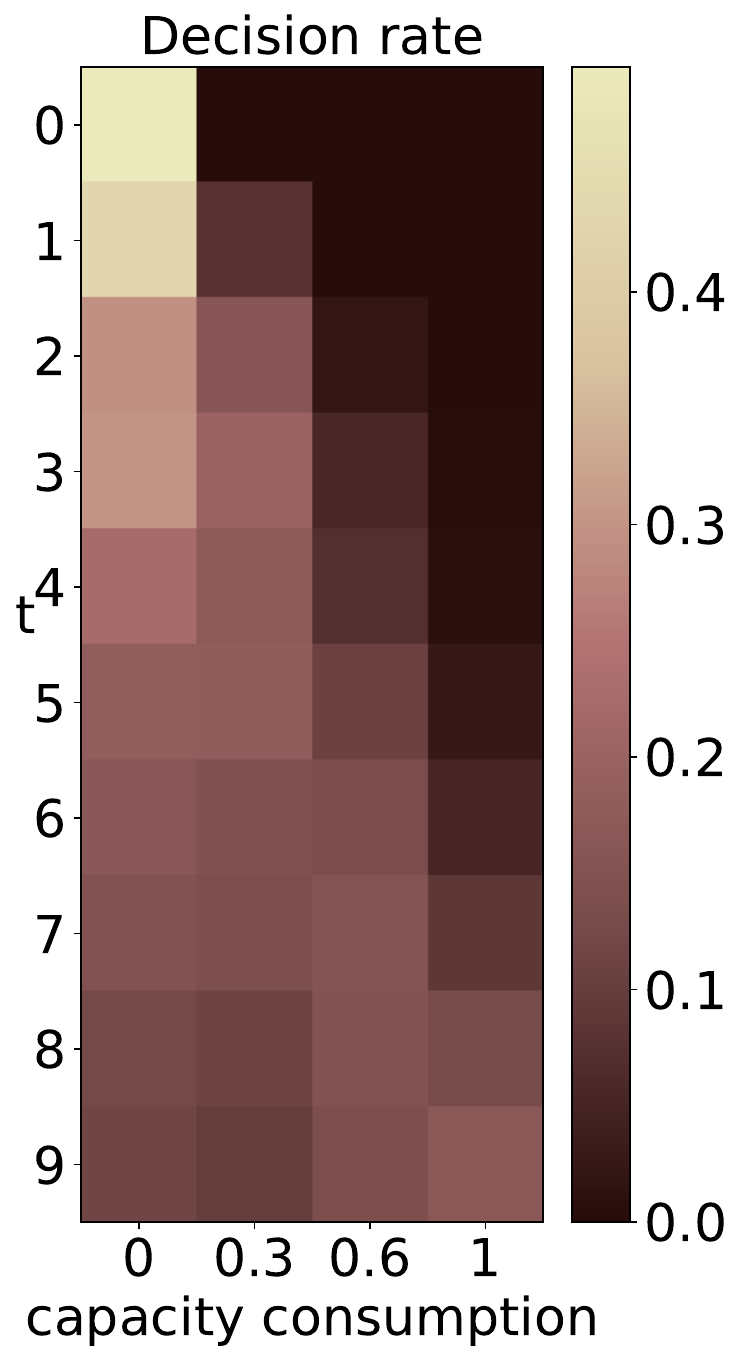}
    		\end{subfigure}
        \caption{low $\mid$ load $\mid$ clust $\mid$ l-b-h }
        \label{fig: HM4b}
        \end{subfigure}
     \vskip\baselineskip
        \begin{subfigure}{0.6\textwidth}
    		\renewcommand\thesubfigure{c}
            \begin{subfigure}[]{0.3\textwidth}
    			\centering
    			\includegraphics[width=\textwidth]{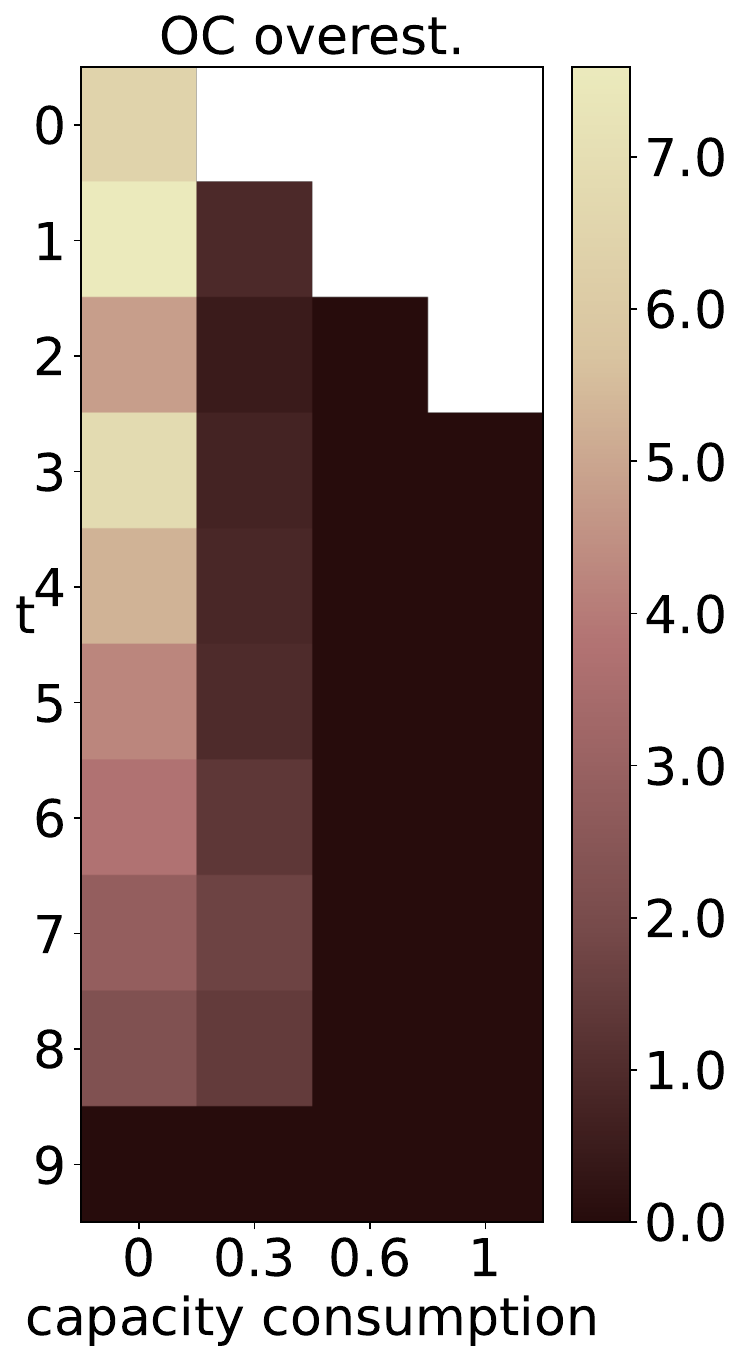}
    		\end{subfigure}
            \begin{subfigure}[]{0.3\textwidth}
    			\centering
    			\includegraphics[width=\textwidth]{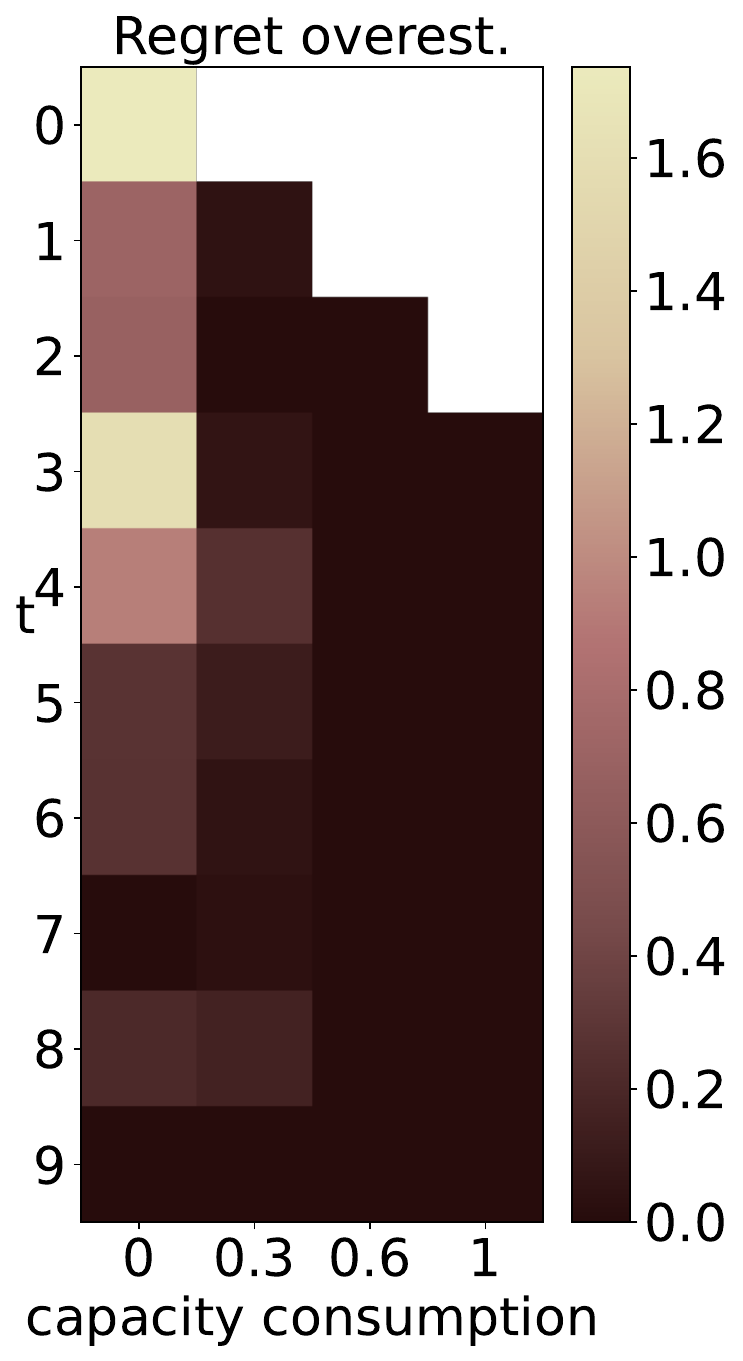}
    		\end{subfigure}
    		\begin{subfigure}[]{0.3\textwidth}
    			\centering
    			\includegraphics[width=\textwidth]{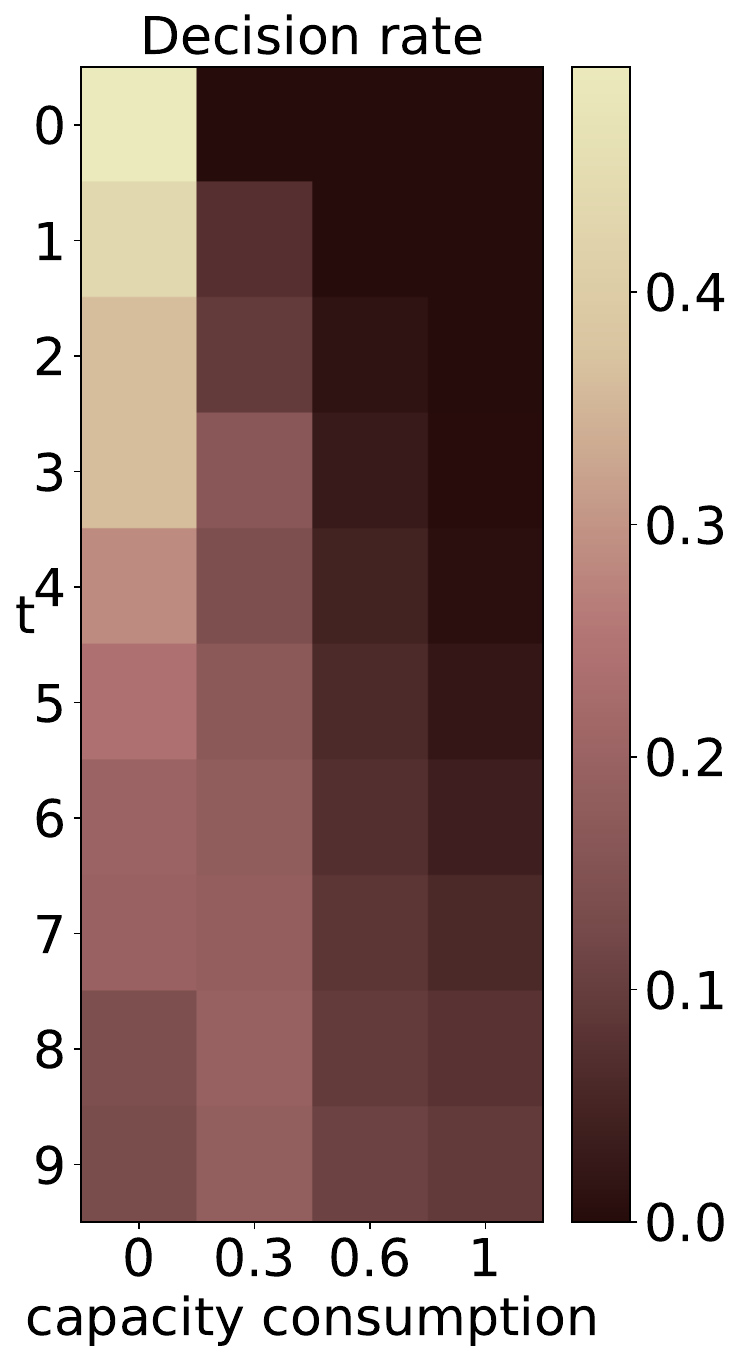}
    		\end{subfigure}
        \caption{low $\mid$ load $\mid$ unif $\mid$ homog }
        \label{fig: HM4c}
        \end{subfigure}
\end{figure}

\vspace{-1cm}
\begin{figure}[H]
\centering
\footnotesize
\caption{Offsetting of underestimation errors and overestimation errors in setting low $\mid$ load $\mid$ unif $\mid$ l-b-h} 
\label{fig: HM5}
    \vskip\baselineskip
	\centering
	\captionsetup{justification=centering}
        \begin{subfigure}{0.8\textwidth}
    		\renewcommand\thesubfigure{a}
            \begin{subfigure}[]{0.251\textwidth}
    			\centering
    			\includegraphics[width=\textwidth]{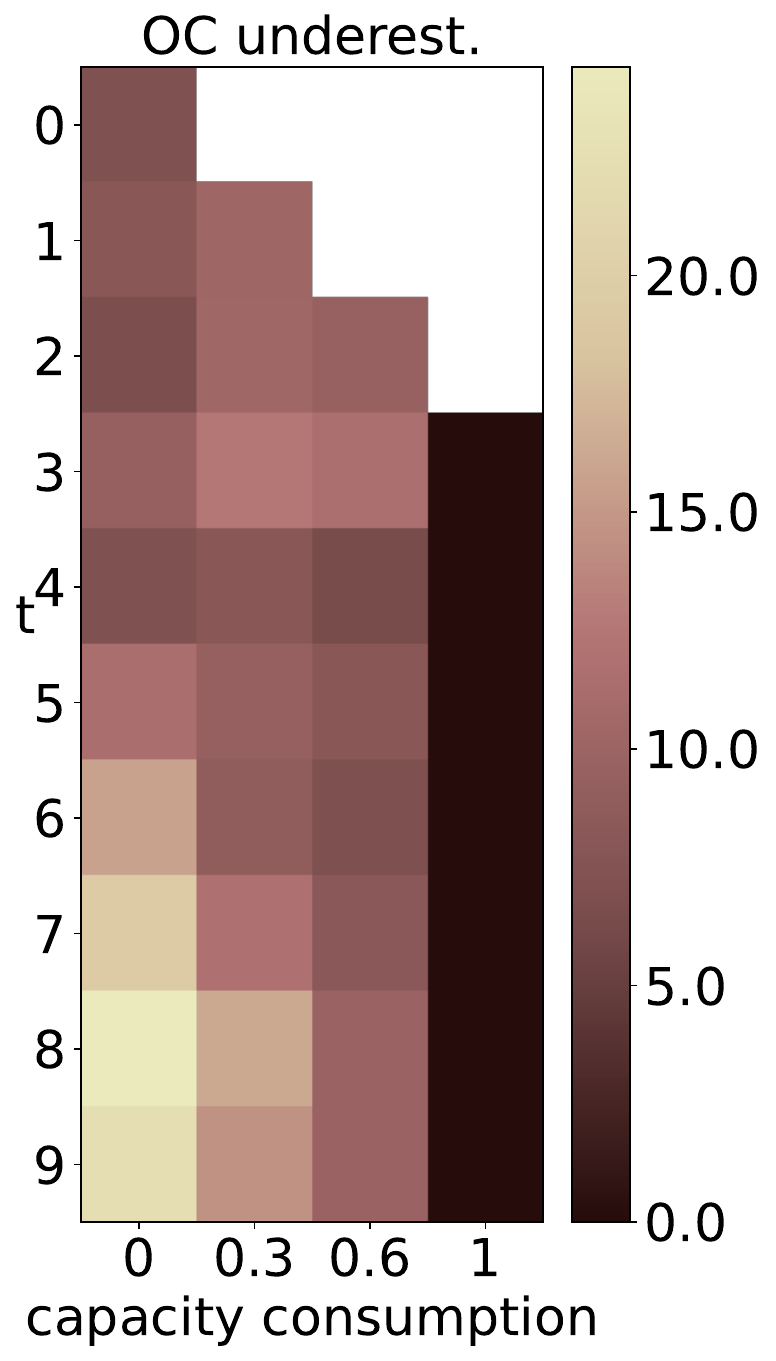}
    		\end{subfigure}
    		\begin{subfigure}[]{0.251\textwidth}
    			\centering
    			\includegraphics[width=\textwidth]{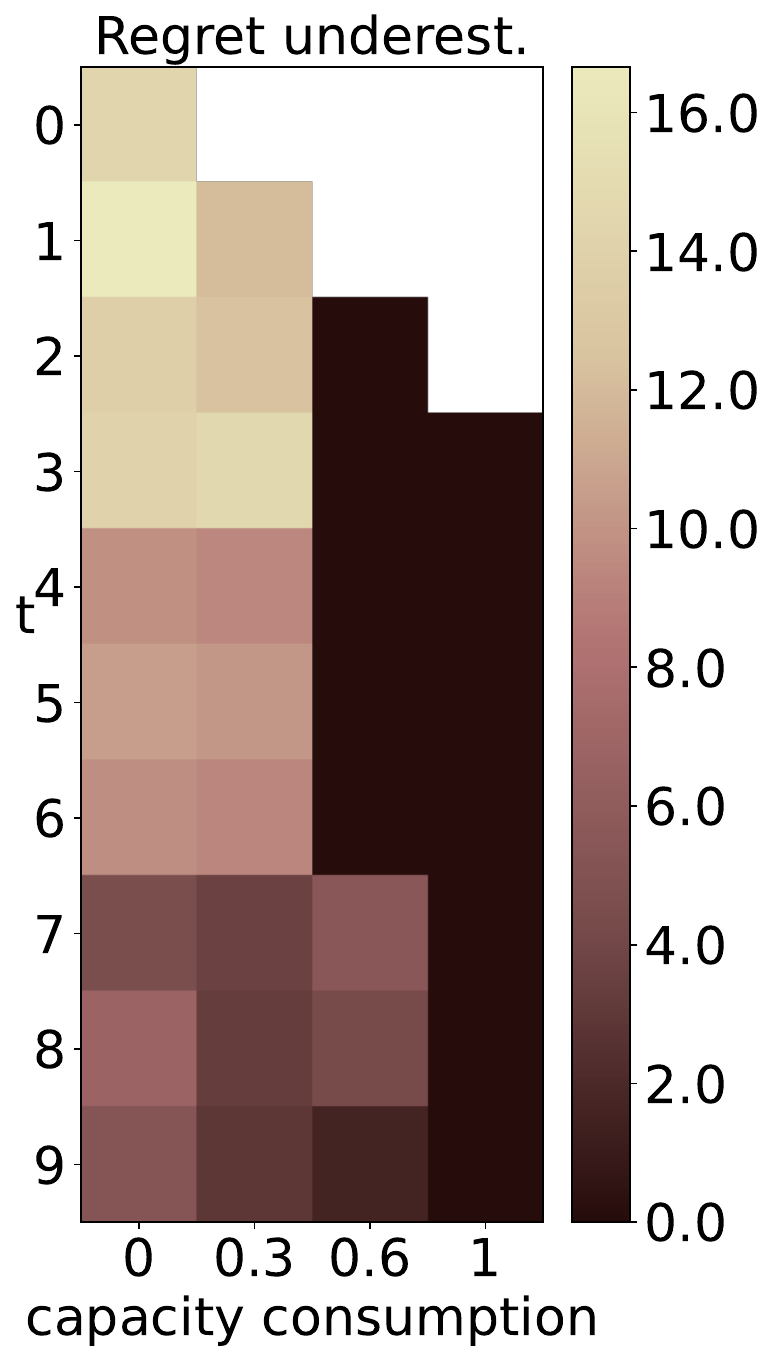}
    		\end{subfigure}
            \begin{subfigure}[]{0.24\textwidth}
    			\centering
    			\includegraphics[width=\textwidth]{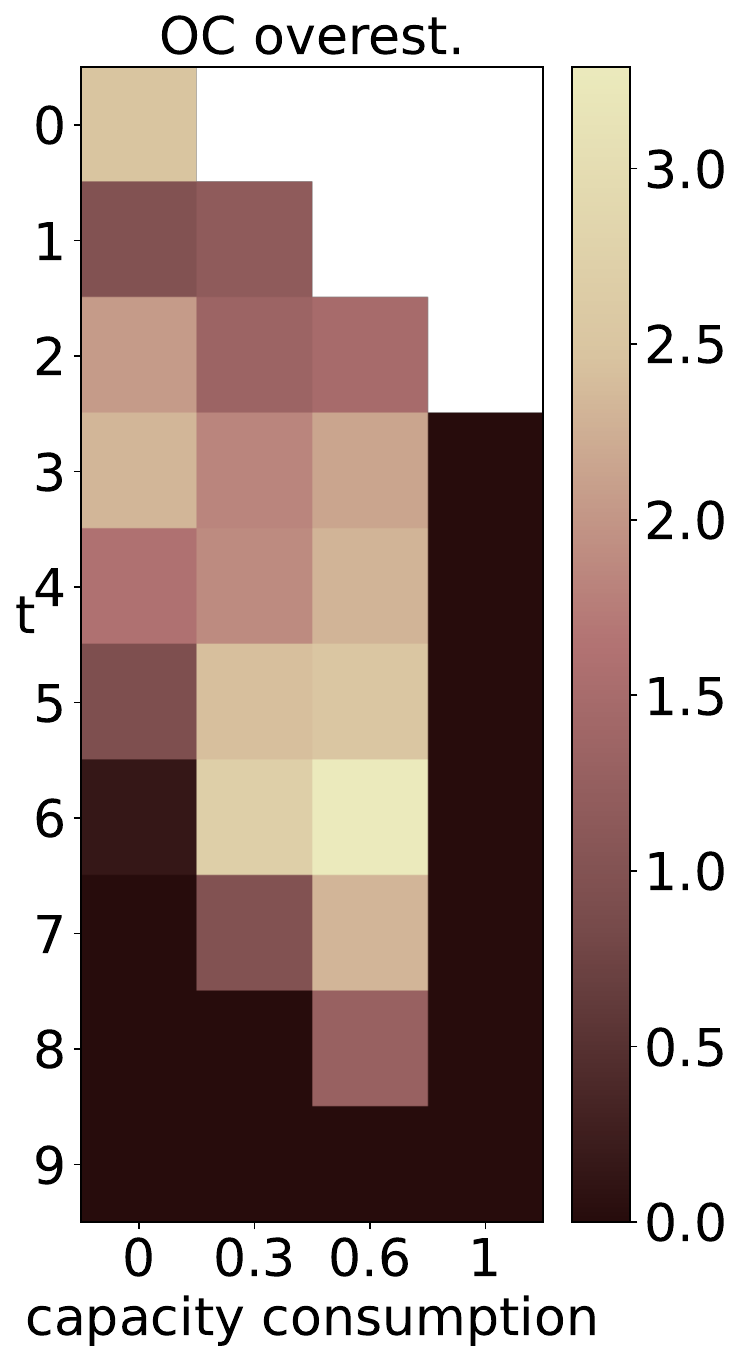}
    		\end{subfigure}
    		\begin{subfigure}[]{0.24\textwidth}
    			\centering
    			\includegraphics[width=\textwidth]{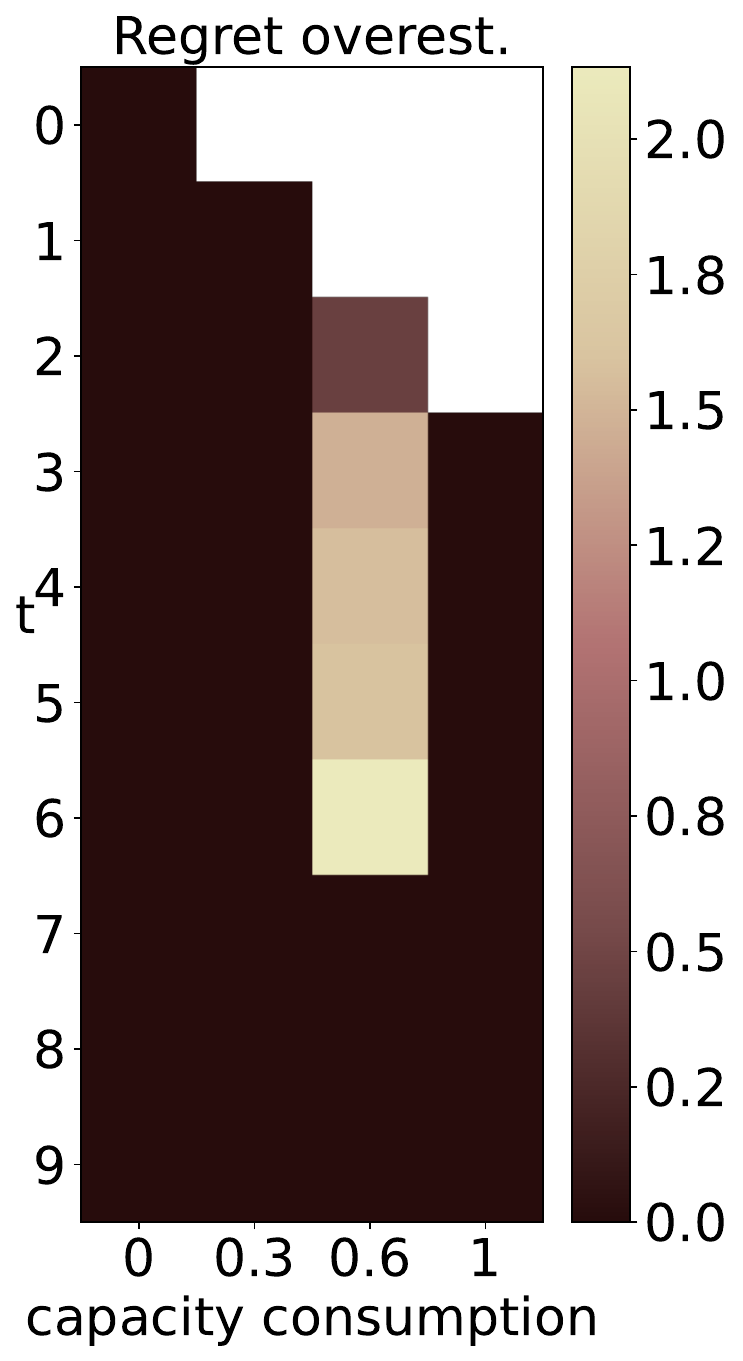}
    		\end{subfigure}
        \caption{DPC}
        \label{fig: HM5a}
        \end{subfigure}
            \vskip\baselineskip
        \begin{subfigure}{0.8\textwidth}
		\renewcommand\thesubfigure{b}
      	\begin{subfigure}[]{0.251\textwidth}
    			\centering
    			\includegraphics[width=\textwidth]{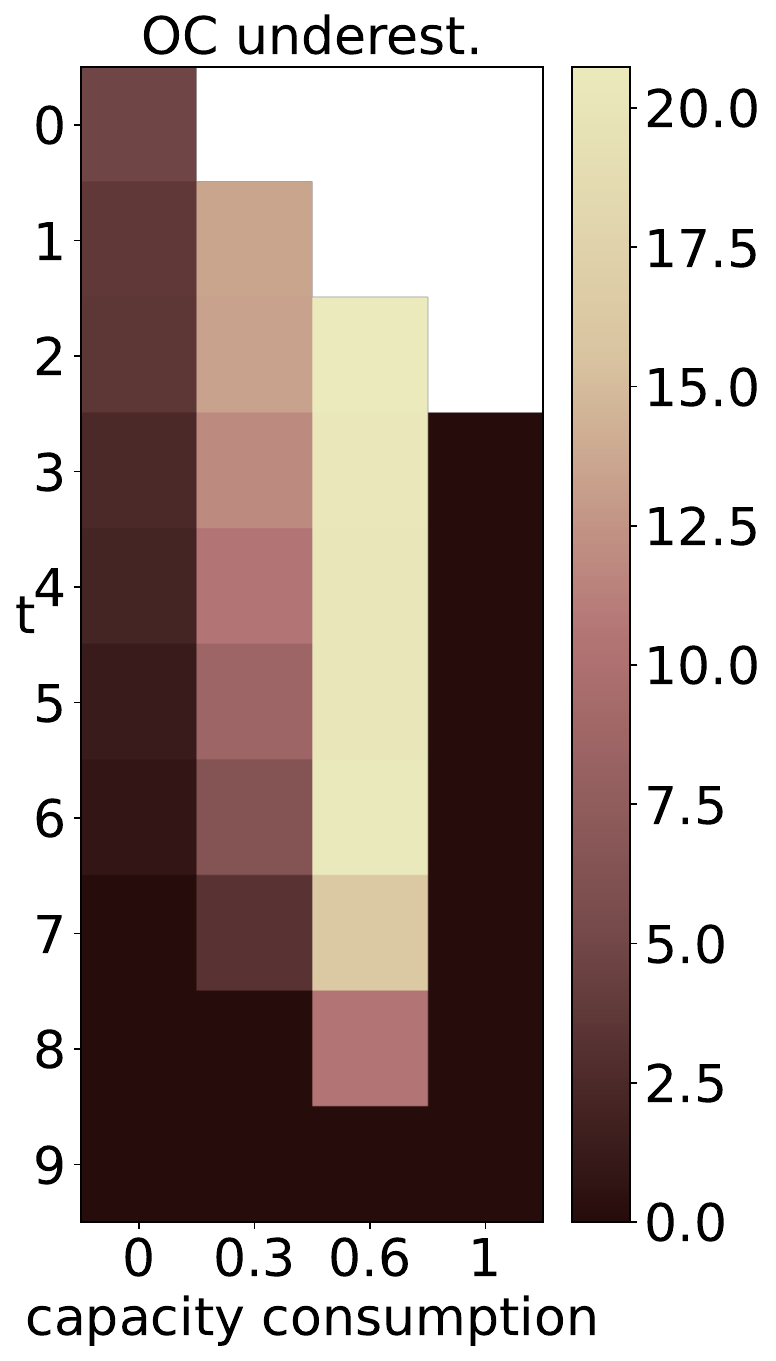}
    		\end{subfigure}
    		\begin{subfigure}[]{0.24\textwidth}
    			\centering
    			\includegraphics[width=\textwidth]{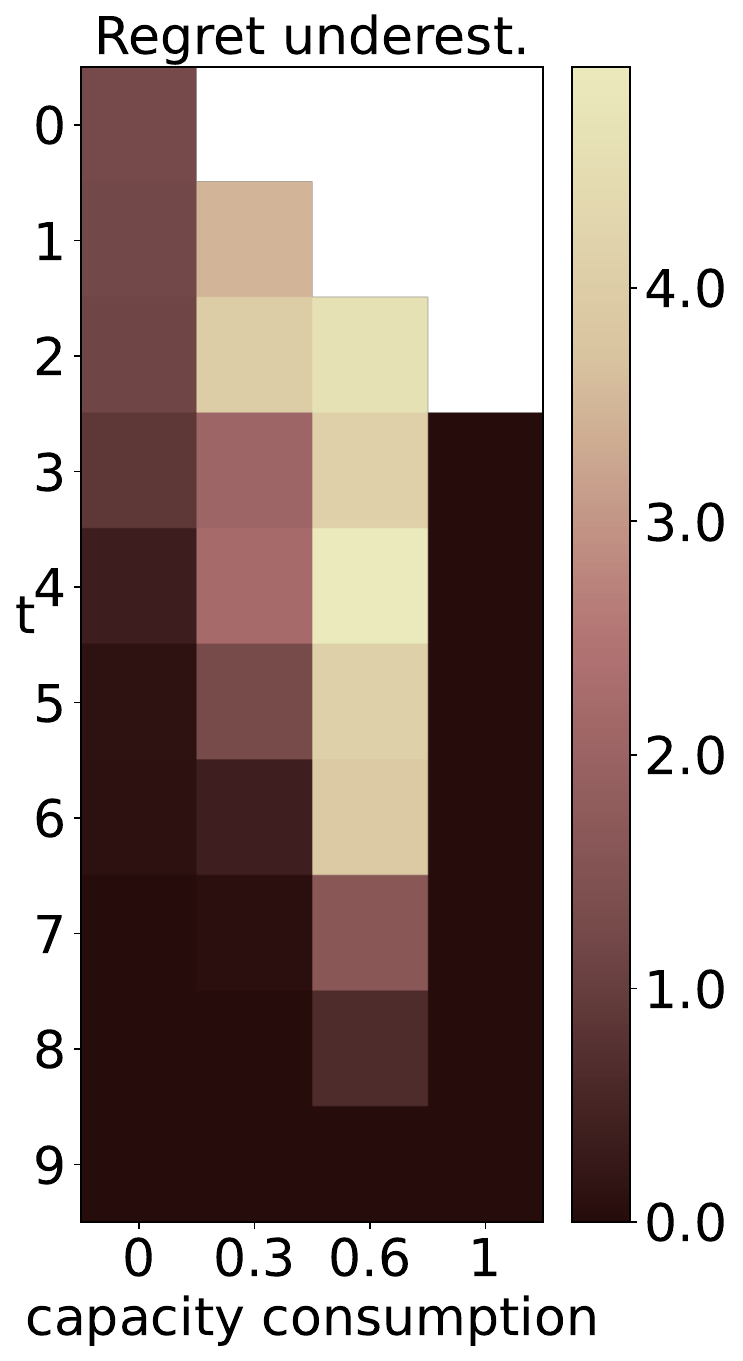}
    		\end{subfigure}
      	\begin{subfigure}[]{0.24\textwidth}
    			\centering
    			\includegraphics[width=\textwidth]{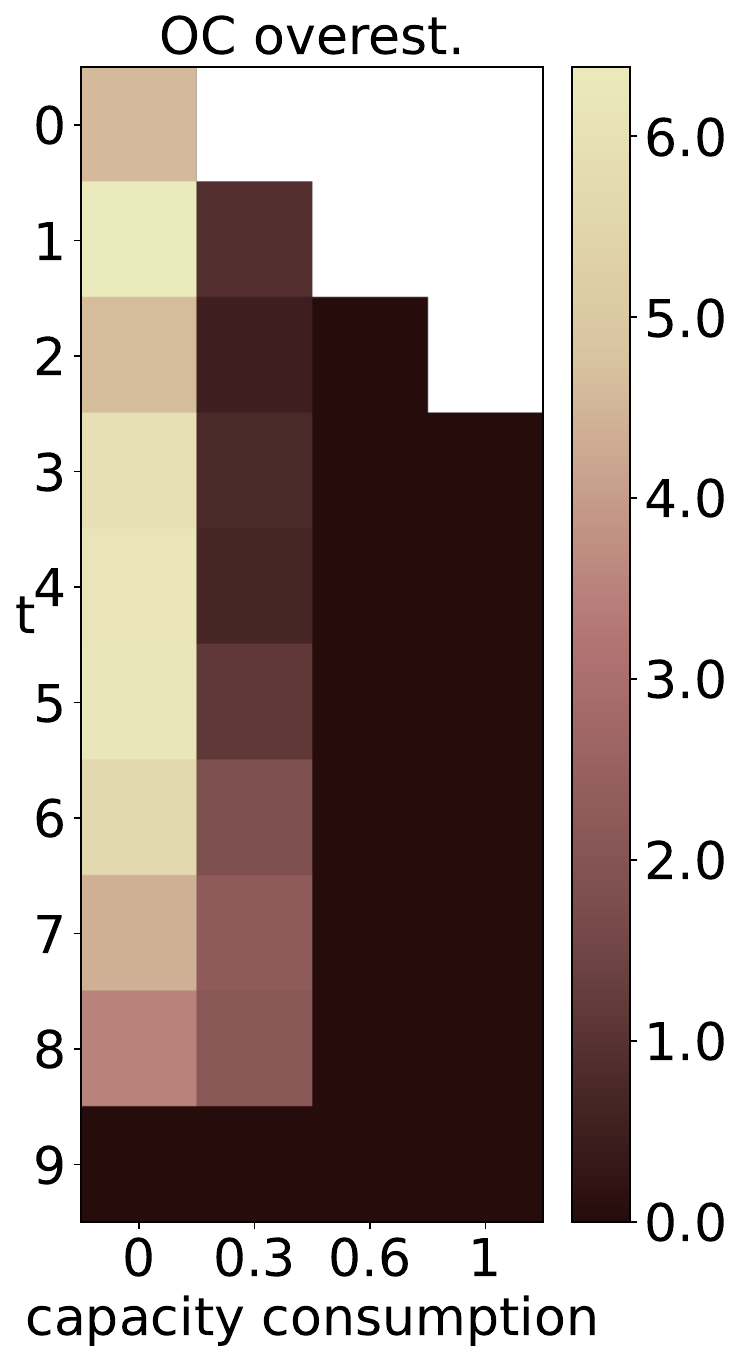}
    		\end{subfigure}
    		\begin{subfigure}[]{0.24\textwidth}
    			\centering
    			\includegraphics[width=\textwidth]{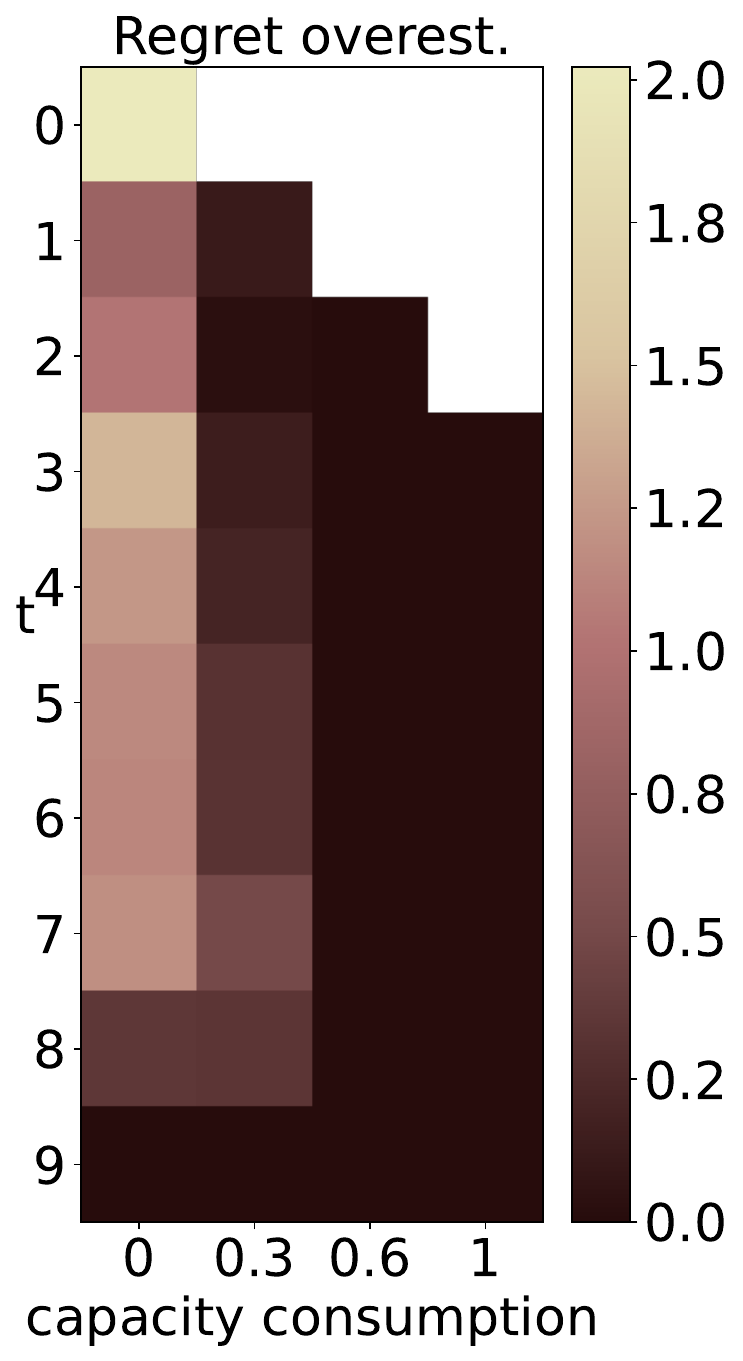}
    		\end{subfigure}
        \caption{MCTS}
        \label{fig: HM5b}
        \end{subfigure}
            \vskip\baselineskip
        \begin{subfigure}{0.8\textwidth}
    		\renewcommand\thesubfigure{c}
            \begin{subfigure}[]{0.251\textwidth}
    			\centering
    			\includegraphics[width=\textwidth]{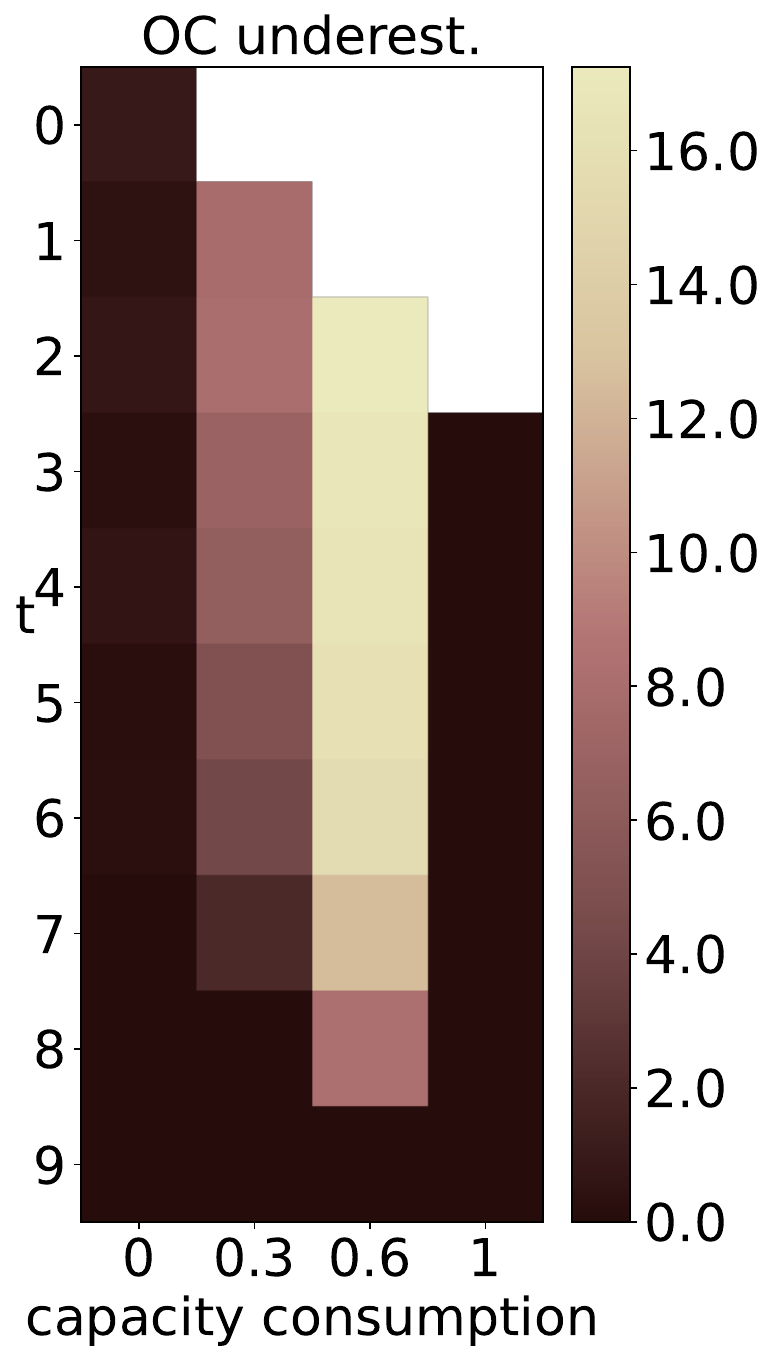}
    		\end{subfigure}
    		\begin{subfigure}[]{0.24\textwidth}
    			\centering
    			\includegraphics[width=\textwidth]{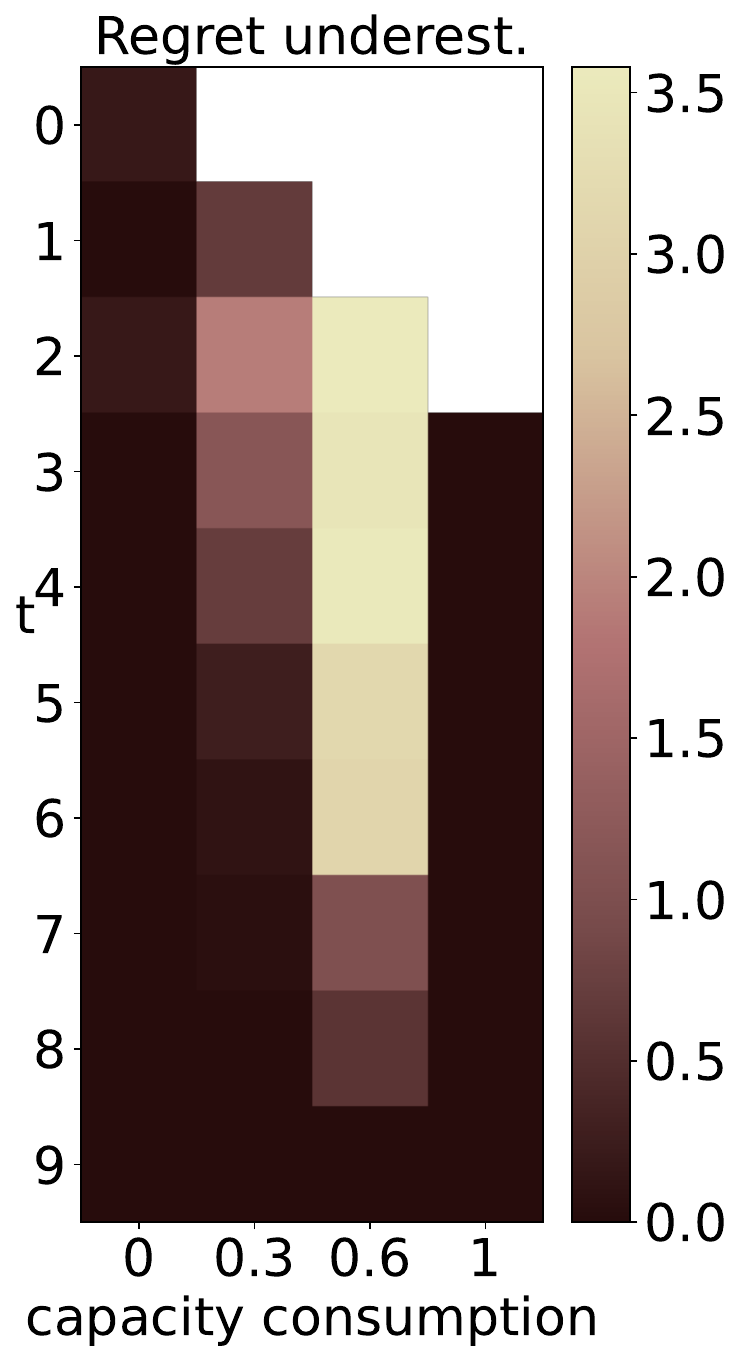}
    		\end{subfigure}
            \begin{subfigure}[]{0.24\textwidth}
    			\centering
    			\includegraphics[width=\textwidth]{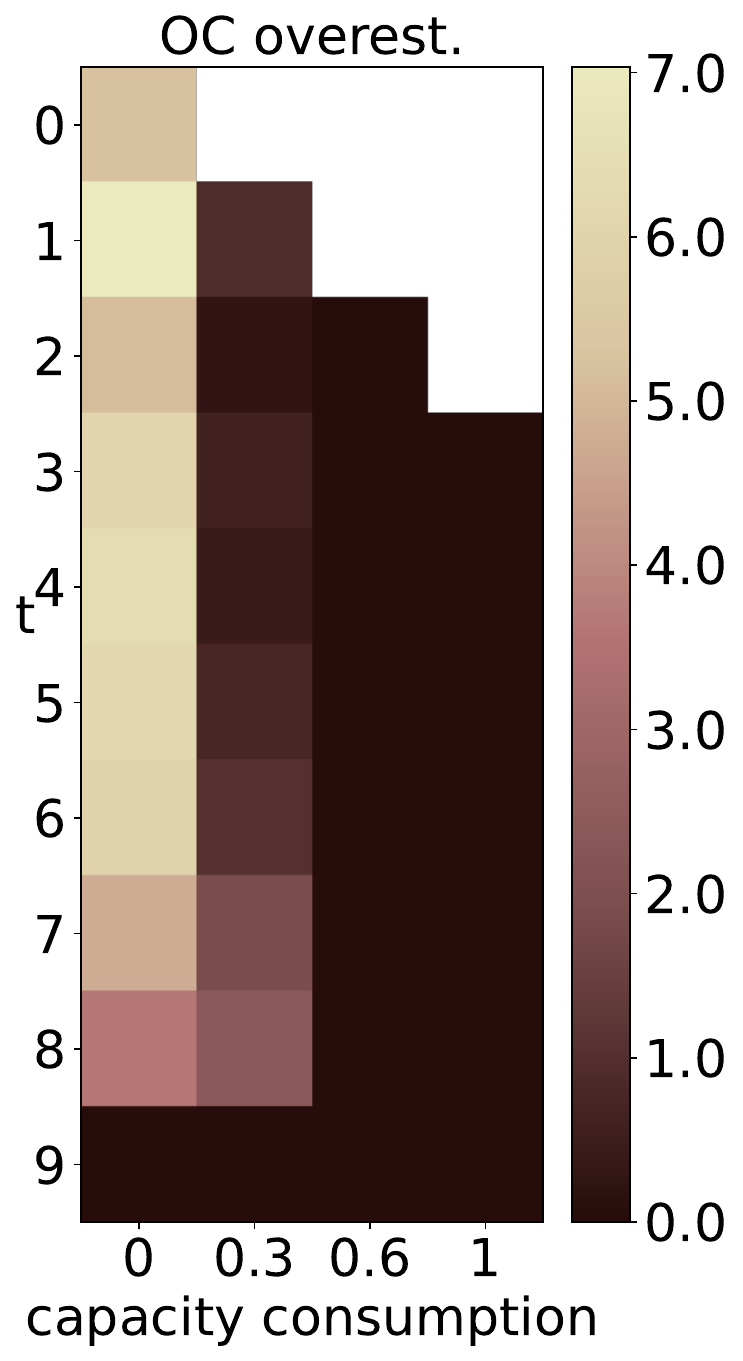}
    		\end{subfigure}
    		\begin{subfigure}[]{0.24\textwidth}
    			\centering
    			\includegraphics[width=\textwidth]{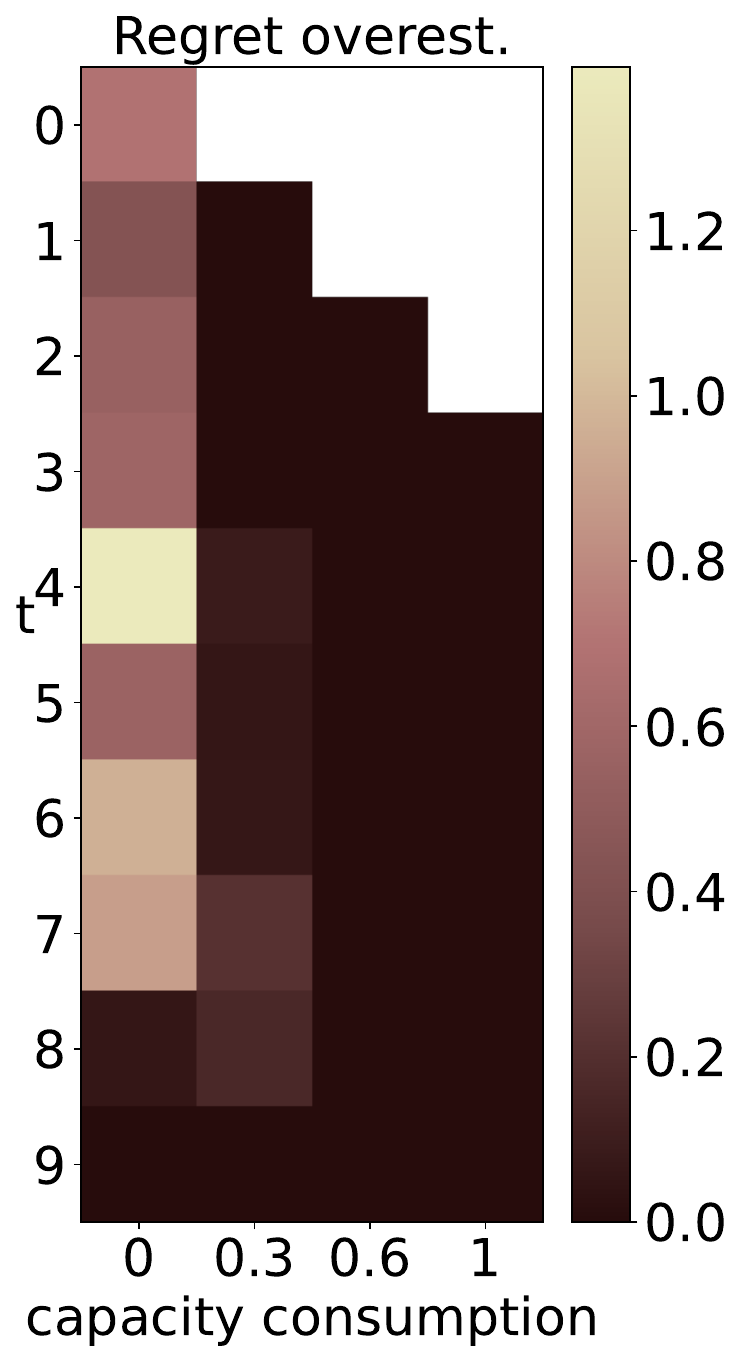}
    		\end{subfigure}
        \caption{mMCTS}
        \end{subfigure}
\end{figure}


\begin{thebibliography}{53}
\providecommand{\natexlab}[1]{#1}
\providecommand{\url}[1]{\texttt{#1}}
\providecommand{\urlprefix}{URL }

\bibitem[{Abdollahi et~al.(2023)Abdollahi, Yang, Nasri, \protect\BIBand{} Fairbank}]{abdollahi2023demand}
Abdollahi M, Yang X, Nasri MI, Fairbank M, 2023 \emph{Demand management in time-slotted last-mile delivery via dynamic routing with forecast orders}. \emph{European Journal of Operational Research} 309(2):704--718.

\bibitem[{Amir, Doshi-Velez, \protect\BIBand{} Sarne(2019)}]{amir2019summarizing}
Amir O, Doshi-Velez F, Sarne D, 2019 \emph{Summarizing agent strategies}. \emph{Autonomous Agents and Multi-Agent Systems} 33:628--644.

\bibitem[{Amorim et~al.(2024)Amorim, DeHoratius, Eng-Larsson, \protect\BIBand{} Martins}]{amorim2024customer}
Amorim P, DeHoratius N, Eng-Larsson F, Martins S, 2024 \emph{Customer preferences for delivery service attributes in attended home delivery}. \emph{Management Science} 70(11):7559--7578.

\bibitem[{Anzenhofer et~al.(2024)Anzenhofer, Fleckenstein, Klein, \protect\BIBand{} Steinhardt}]{anzenhofer2023analyzing}
Anzenhofer F, Fleckenstein D, Klein R, Steinhardt C, 2024 \emph{Analyzing the impact of demand management in rural shared mobility-on-demand systems}. \emph{Working paper, University of Augsburg} .

\bibitem[{Arian, Bai, \protect\BIBand{} Chen(2022)}]{arian2022joint}
Arian E, Bai X, Chen X, 2022 \emph{Joint pricing and routing for a ride-sharing platform in low-density rural areas}. \emph{Working paper, University of Illinois} .

\bibitem[{Arrieta et~al.(2020)Arrieta, D{\'\i}az-Rodr{\'\i}guez, Del~Ser, Bennetot, Tabik, Barbado, Garc{\'\i}a, Gil-L{\'o}pez, Molina, Benjamins et~al.}]{arrieta2020explainable}
Arrieta AB, D{\'\i}az-Rodr{\'\i}guez N, Del~Ser J, Bennetot A, Tabik S, Barbado A, Garc{\'\i}a S, Gil-L{\'o}pez S, Molina D, Benjamins R, et~al., 2020 \emph{Explainable artificial intelligence (XAI): Concepts, taxonomies, opportunities and challenges toward responsible AI}. \emph{Information Fusion} 58:82--115.

\bibitem[{Asdemir, Jacob, \protect\BIBand{} Krishnan(2009)}]{asdemir2009dynamic}
Asdemir K, Jacob VS, Krishnan R, 2009 \emph{Dynamic pricing of multiple home delivery options}. \emph{European Journal of Operational Research} 196(1):246--257.

\bibitem[{Atasoy et~al.(2015)Atasoy, Ikeda, Song, \protect\BIBand{} Ben-Akiva}]{atasoy2015concept}
Atasoy B, Ikeda T, Song X, Ben-Akiva ME, 2015 \emph{The concept and impact analysis of a flexible mobility on demand system}. \emph{Transportation Research Part C: Emerging Technologies} 56:373--392.

\bibitem[{Avraham \protect\BIBand{} Raviv(2021)}]{avraham2021steady}
Avraham E, Raviv T, 2021 \emph{The steady-state mobile personnel booking problem}. \emph{Transportation Research Part B: Methodological} 154:266--288.

\bibitem[{Azi, Gendreau, \protect\BIBand{} Potvin(2012)}]{azi2012dynamic}
Azi N, Gendreau M, Potvin JY, 2012 \emph{A dynamic vehicle routing problem with multiple delivery routes}. \emph{Annals of Operations Research} 199(1):103--112.

\bibitem[{Biggs, Sun, \protect\BIBand{} Ettl(2021)}]{biggs2021model}
Biggs M, Sun W, Ettl M, 2021 \emph{Model distillation for revenue optimization: Interpretable personalized pricing}. \emph{International Conference on Machine Learning}, 946--956 (PMLR).

\bibitem[{Bravo \protect\BIBand{} Shaposhnik(2020)}]{bravo2020mining}
Bravo F, Shaposhnik Y, 2020 \emph{Mining optimal policies: A pattern recognition approach to model analysis}. \emph{INFORMS Journal on Optimization} 2(3):145--166.

\bibitem[{Campbell \protect\BIBand{} Savelsbergh(2005)}]{campbell2005decision}
Campbell AM, Savelsbergh M, 2005 \emph{Decision support for consumer direct grocery initiatives}. \emph{Transportation Science} 39(3):313--327.

\bibitem[{Campbell \protect\BIBand{} Savelsbergh(2006)}]{campbell2006incentive}
Campbell AM, Savelsbergh M, 2006 \emph{Incentive schemes for attended home delivery services}. \emph{Transportation Science} 40(3):327--341.

\bibitem[{Castillo, Knoepfle, \protect\BIBand{} Weyl(2024)}]{castillo2022matching}
Castillo JC, Knoepfle DT, Weyl EG, 2024 \emph{Matching in ride hailing: Wild goose chases and how to solve them}. \emph{Working paper, University of Pennsylvania} .

\bibitem[{De~Bock et~al.(2024)De~Bock, Coussement, De~Caigny, Slowi{\'n}ski, Baesens, Boute, Choi, Delen, Kraus, Lessmann et~al.}]{debock2023explainable}
De~Bock KW, Coussement K, De~Caigny A, Slowi{\'n}ski R, Baesens B, Boute RN, Choi TM, Delen D, Kraus M, Lessmann S, et~al., 2024 \emph{Explainable AI for operational research: A defining framework, methods, applications, and a research agenda}. \emph{European Journal of Operational Research} 317(2):249--272.

\bibitem[{Fanelli(2012)}]{fanelli2012negative}
Fanelli D, 2012 \emph{Negative results are disappearing from most disciplines and countries}. \emph{Scientometrics} 90(3):891--904.

\bibitem[{Fleckenstein et~al.(2024)Fleckenstein, Klein, Klein, \protect\BIBand{} Steinhardt}]{fleckenstein2023concept}
Fleckenstein D, Klein R, Klein V, Steinhardt C, 2024 \emph{On the concept of opportunity cost in integrated demand management and vehicle routing}. \emph{Working paper, University of Augsburg} .

\bibitem[{Fleckenstein, Klein, \protect\BIBand{} Steinhardt(2023)}]{fleckenstein2021recent}
Fleckenstein D, Klein R, Steinhardt C, 2023 \emph{Recent advances in integrating demand management and vehicle routing: A methodological review}. \emph{European Journal of Operational Research} 306(2):499--518.

\bibitem[{Gallego \protect\BIBand{} Topaloglu(2019)}]{gallego2019revenue}
Gallego G, Topaloglu H, 2019 \emph{Revenue management and pricing analytics} (Springer).

\bibitem[{Goerigk \protect\BIBand{} Hartisch(2023)}]{goerigk2023framework}
Goerigk M, Hartisch M, 2023 \emph{A framework for inherently interpretable optimization models}. \emph{European Journal of Operational Research} 310(3):1312--1324.

\bibitem[{Heitmann et~al.(2023)Heitmann, Soeffker, Ulmer, \protect\BIBand{} Mattfeld}]{heitmann2023combining}
Heitmann RJO, Soeffker N, Ulmer MW, Mattfeld DC, 2023 \emph{Combining value function approximation and multiple scenario approach for the effective management of ride-hailing services}. \emph{EURO Journal on Transportation and Logistics} 12:100104.

\bibitem[{Hildebrandt, Thomas, \protect\BIBand{} Ulmer(2023)}]{hildebrandt2023opportunities}
Hildebrandt FD, Thomas BW, Ulmer MW, 2023 \emph{Opportunities for reinforcement learning in stochastic dynamic vehicle routing}. \emph{Computers \& Operations Research} 150:106071.

\bibitem[{Juozapaitis et~al.(2019)Juozapaitis, Koul, Fern, Erwig, \protect\BIBand{} Doshi-Velez}]{juozapaitis2019explainable}
Juozapaitis Z, Koul A, Fern A, Erwig M, Doshi-Velez F, 2019 \emph{Explainable reinforcement learning via reward decomposition}. \emph{IJCAI/ECAI Workshop on Explainable Artificial Intelligence}.

\bibitem[{Klapp, Erera, \protect\BIBand{} Toriello(2020)}]{klapp2020request}
Klapp MA, Erera AL, Toriello A, 2020 \emph{Request acceptance in same-day delivery}. \emph{Transportation Research Part E: Logistics and Transportation Review} 143:102083.

\bibitem[{Klein et~al.(2020)Klein, Koch, Steinhardt, \protect\BIBand{} Strauss}]{klein2020review}
Klein R, Koch S, Steinhardt C, Strauss AK, 2020 \emph{A review of revenue management: Recent generalizations and advances in industry applications}. \emph{European Journal of Operational Research} 284(2):397--412.

\bibitem[{Klein et~al.(2018)Klein, Mackert, Neugebauer, \protect\BIBand{} Steinhardt}]{klein2018model}
Klein R, Mackert J, Neugebauer M, Steinhardt C, 2018 \emph{A model-based approximation of opportunity cost for dynamic pricing in attended home delivery}. \emph{OR Spectrum} 40(4):969--996.

\bibitem[{Klein \protect\BIBand{} Steinhardt(2023)}]{klein2021dynamic}
Klein V, Steinhardt C, 2023 \emph{Dynamic demand management and online tour planning for same-day delivery}. \emph{European Journal of Operational Research} 307(2):860--886.

\bibitem[{Koch \protect\BIBand{} Klein(2020)}]{koch2020route}
Koch S, Klein R, 2020 \emph{Route-based approximate dynamic programming for dynamic pricing in attended home delivery}. \emph{European Journal of Operational Research} 287(2):633--652.

\bibitem[{Lang, Cleophas, \protect\BIBand{} Ehmke(2021)}]{lang2021anticipative}
Lang MA, Cleophas C, Ehmke JF, 2021 \emph{Anticipative dynamic slotting for attended home deliveries}. \emph{Operations Research Forum} 2:70.

\bibitem[{Lebedev, Goulart, \protect\BIBand{} Margellos(2021)}]{lebedev2021dynamic}
Lebedev D, Goulart P, Margellos K, 2021 \emph{A dynamic programming framework for optimal delivery time slot pricing}. \emph{European Journal of Operational Research} 292(2):456--468.

\bibitem[{Li, Archetti, \protect\BIBand{} Ljubic(2024)}]{liemerging}
Li Y, Archetti C, Ljubic I, 2024 \emph{Emerging optimization problems for distribution in same-day delivery}. \emph{Working paper, ESSEC Business School} .

\bibitem[{Mackert(2019)}]{mackert2019choice}
Mackert J, 2019 \emph{Choice-based dynamic time slot management in attended home delivery}. \emph{Computers \& Industrial Engineering} 129:333--345.

\bibitem[{Mannor et~al.(2007)Mannor, Simester, Sun, \protect\BIBand{} Tsitsiklis}]{mannor2007bias}
Mannor S, Simester D, Sun P, Tsitsiklis JN, 2007 \emph{Bias and variance approximation in value function estimates}. \emph{Management Science} 53(2):308--322.

\bibitem[{Milani et~al.(2024)Milani, Topin, Veloso, \protect\BIBand{} Fang}]{milani2023explainable}
Milani S, Topin N, Veloso M, Fang F, 2024 \emph{Explainable reinforcement learning: A survey and comparative review}. \emph{ACM Computing Surveys} 56(7):1--36.

\bibitem[{Powell(2011)}]{powell2007approximate}
Powell WB, 2011 \emph{Approximate Dynamic Programming: Solving the curses of dimensionality} (John Wiley \& Sons).

\bibitem[{Powell(2022)}]{powell2022reinforcement}
Powell WB, 2022 \emph{Reinforcement Learning and Stochastic Optimization: A Unified Framework for Sequential Decisions} (John Wiley \& Sons).

\bibitem[{Psaraftis, Wen, \protect\BIBand{} Kontovas(2016)}]{psaraftis2016dynamic}
Psaraftis HN, Wen M, Kontovas CA, 2016 \emph{Dynamic vehicle routing problems: Three decades and counting}. \emph{Networks} 67(1):3--31.

\bibitem[{Puterman(2014)}]{puterman2014markov}
Puterman ML, 2014 \emph{Markov decision processes: discrete stochastic dynamic programming} (John Wiley \& Sons).

\bibitem[{Russell \protect\BIBand{} Zimdars(2003)}]{russell2003q}
Russell SJ, Zimdars A, 2003 \emph{Q-decomposition for reinforcement learning agents}. \emph{Proceedings of the 20th International Conference on Machine Learning (ICML-03)}, 656--663.

\bibitem[{Snoeck, Merch{\'a}n, \protect\BIBand{} Winkenbach(2020)}]{snoeck2020revenue}
Snoeck A, Merch{\'a}n D, Winkenbach M, 2020 \emph{Revenue management in last-mile delivery: state-of-the-art and future research directions}. \emph{Transportation Research Procedia} 46:109--116.

\bibitem[{Soeffker, Ulmer, \protect\BIBand{} Mattfeld(2022)}]{soeffker2022stochastic}
Soeffker N, Ulmer MW, Mattfeld DC, 2022 \emph{Stochastic dynamic vehicle routing in the light of prescriptive analytics: A review}. \emph{European Journal of Operational Research} 298(3):801--820.

\bibitem[{Strauss, Klein, \protect\BIBand{} Steinhardt(2018)}]{strauss2018review}
Strauss AK, Klein R, Steinhardt C, 2018 \emph{A review of choice-based revenue management: Theory and methods}. \emph{European Journal of Operational Research} 271(2):375--387.

\bibitem[{Talluri \protect\BIBand{} Van~Ryzin(2004)}]{talluri2004theory}
Talluri K, Van~Ryzin G, 2004 \emph{The theory and practice of revenue management} (Springer).

\bibitem[{Torrey \protect\BIBand{} Taylor(2013)}]{torrey2013teaching}
Torrey L, Taylor M, 2013 \emph{Teaching on a budget: Agents advising agents in reinforcement learning}. \emph{Proceedings of the 2013 international conference on autonomous agents and multi-agent systems}, 1053--1060.

\bibitem[{Toth \protect\BIBand{} Vigo(2014)}]{toth2014vehicle}
Toth P, Vigo D, 2014 \emph{Vehicle routing: problems, methods, and applications} (SIAM).

\bibitem[{Ulmer(2020)}]{ulmer2020dynamic}
Ulmer MW, 2020 \emph{Dynamic pricing and routing for same-day delivery}. \emph{Transportation Science} 54(4):1016--1033.

\bibitem[{Ulmer \protect\BIBand{} Thomas(2020)}]{ulmer2020meso}
Ulmer MW, Thomas BW, 2020 \emph{Meso-parametric value function approximation for dynamic customer acceptances in delivery routing}. \emph{European Journal of Operational Research} 285(1):183--195.

\bibitem[{Van~Seijen et~al.(2017)Van~Seijen, Fatemi, Romoff, Laroche, Barnes, \protect\BIBand{} Tsang}]{van2017hybrid}
Van~Seijen H, Fatemi M, Romoff J, Laroche R, Barnes T, Tsang J, 2017 \emph{Hybrid reward architecture for reinforcement learning}. \emph{Advances in Neural Information Processing Systems} 30.

\bibitem[{Vansteenwegen et~al.(2022)Vansteenwegen, Melis, Akta{\c{s}}, Montenegro, Vieira, \protect\BIBand{} S{\"o}rensen}]{vansteenwegen2022survey}
Vansteenwegen P, Melis L, Akta{\c{s}} D, Montenegro BDG, Vieira FS, S{\"o}rensen K, 2022 \emph{A survey on demand-responsive public bus systems}. \emph{Transportation Research Part C: Emerging Technologies} 137:103573.

\bibitem[{Wa{\ss}muth et~al.(2023)Wa{\ss}muth, K{\"o}hler, Agatz, \protect\BIBand{} Fleischmann}]{wassmuth2022demand}
Wa{\ss}muth K, K{\"o}hler C, Agatz N, Fleischmann M, 2023 \emph{Demand management for attended home delivery--a literature review}. \emph{European Journal of Operational Research} 311(3):801--815.

\bibitem[{Yang \protect\BIBand{} Strauss(2017)}]{yang2017approximate}
Yang X, Strauss AK, 2017 \emph{An approximate dynamic programming approach to attended home delivery management}. \emph{European Journal of Operational Research} 263(3):935--945.

\bibitem[{Yang et~al.(2016)Yang, Strauss, Currie, \protect\BIBand{} Eglese}]{yang2016choice}
Yang X, Strauss AK, Currie CS, Eglese R, 2016 \emph{Choice-based demand management and vehicle routing in e-fulfillment}. \emph{Transportation Science} 50(2):473--488.

\end{thebibliography}
\end{document}